\documentclass[acmsmall]{acmart}

\usepackage{threeparttable}
\usepackage{multirow}
\usepackage{subfigure}
\usepackage{tablefootnote}
\usepackage{colortbl}
\definecolor{Ocean}{RGB}{200, 230, 250}
\definecolor{lightpink}{RGB}{255, 220, 230} % 更浅的淡粉色

\AtBeginDocument{%
  }

\setcopyright{acmlicensed}
\copyrightyear{2018}
\acmYear{2018}
\acmDOI{XXXXXXX.XXXXXXX}

\acmJournal{JACM}
\acmVolume{37}
\acmNumber{4}
\acmArticle{111}
\acmMonth{8}

\begin{document}

\title{Towards Reliable Detection of LLM-Generated Texts: A Comprehensive Evaluation Framework with CUDRT}

\author{Zhen Tao}
\affiliation{%
  \institution{Renmin University of China}
  \city{Beijing}
  \country{China}
}
\email{taozhen@ruc.edu.cn}

\author{Yanfang Chen}
\affiliation{%
  \institution{Renmin University of China}
  \city{Beijing}
  \country{China}
}
\email{cyf@ruc.edu.cn}

\author{Dinghao Xi}
\authornote{Corresponding author.}
\affiliation{%
  \institution{Shanghai University of Finance and Economics}
  \city{Shanghai}
  % \state{Texas}
  \country{China}}
\email{xidinghao@mail.shufe.edu.cn}

\author{Zhiyu Li}
\affiliation{%
  \institution{Institute for Advanced Algorithms Research}
  \city{Shanghai}
  \country{China}
}
\email{lizy@iaar.ac.cn}

\author{Wei Xu}
\affiliation{%
  \institution{Renmin University of China}
  \city{Beijing}
  \country{China}}
\email{weixu@ruc.edu.cn}

% \authornote{\textsuperscript{\dag} Corresponding author.}

\renewcommand{\shortauthors}{Tao et al.}

\begin{abstract}
The increasing prevalence of large language models (LLMs) has significantly advanced text generation, but the human-like quality of LLM outputs presents major challenges in reliably distinguishing between human-authored and LLM-generated texts. Existing detection benchmarks are constrained by their reliance on static datasets, scenario-specific tasks (e.g., question answering and text refinement), and a primary focus on English, overlooking the diverse linguistic and operational subtleties of LLMs. To address these gaps, we propose \textbf{CUDRT}, a comprehensive evaluation framework and bilingual benchmark in Chinese and English, categorizing LLM activities into five key operations: \textbf{C}reate, \textbf{U}pdate, \textbf{D}elete, \textbf{R}ewrite, and \textbf{T}ranslate. \textbf{CUDRT} provides extensive datasets tailored to each operation, featuring outputs from state-of-the-art LLMs to assess the reliability of LLM-generated text detectors. This framework supports scalable, reproducible experiments and enables in-depth analysis of how operational diversity, multilingual training sets, and LLM architectures influence detection performance. Our extensive experiments demonstrate the framework’s capacity to optimize detection systems, providing critical insights to enhance reliability, cross-linguistic adaptability, and detection accuracy. By advancing robust methodologies for identifying LLM-generated texts, this work contributes to the development of intelligent systems capable of meeting real-world multilingual detection challenges. Source code and dataset are available at GitHub\footnote{https://github.com/TaoZhen1110/CUDRT}.

\end{abstract}




\received{20 February 2007}
\received[revised]{12 March 2009}
\received[accepted]{5 June 2009}

\maketitle

%%%%%%%%%%%%%%%%%%%%%%%%%%%%%%%%%%%%%%%%%%%%%%%%%%%%%%%%%%%%%
\section{Introduction}\label{sec1}

The emergence and rapid growth of large language models (LLMs) such as ChatGPT \cite{achiam2023gpt}, Llama \cite{touvron2023llama}, and Qwen \cite{bai2023qwen} have revolutionized natural language processing (NLP), enabling unprecedented advances in text understanding and generation \cite{zhou2024vision}. LLMs are now widely applied in diverse domains, ranging from content creation and automated customer service to scientific research and legal document drafting. However, the increasing human-like quality of LLM outputs presents critical challenges in reliably distinguishing between human-authored and LLM-generated content, raising concerns related to information security, copyright, and ethical misuse \cite{chen2023can}.

Currently, there are numerous methods for detecting and identifying LLM-generated texts. Given that human performance in distinguishing between texts generated by LLMs and those written by humans barely surpasses random guessing, research has significantly shifted towards automated detection methods. These methods aim to identify subtle cues that are typically imperceptible to humans. The existing LLM-generated text detection methods can be divided into two categories based on their working principles \cite{2023-coco}: \textbf{metric-based methods} and \textbf{model-based methods}. Metric-based methods, such as GLTR \cite{gltr}, primarily rely on quantitative statistical indicators to evaluate text characteristics, while model-based methods, like RoBERTa \cite{liu2019roberta}, depend on deep learning models to identify generated texts. Each type of method possesses its own distinct advantages and limitations. Metric-based methods offer transparency and are computationally efficient but may lack the ability to capture complex language patterns, leading to lower accuracy \cite{he2023mgtbench}. In contrast, model-based methods like RoBERTa provide high accuracy and adaptability through the use of deep learning, but they act as black boxes for detection, lack some interpretability, and require a large amount of labeled data. Metric-based methods operate based on pre-defined rules, eliminating the need for training and allowing their test results to correlate directly with model performance. Conversely, model-based methods require supervised training on labeled datasets, and their final detection performance is influenced not only by the model architecture but also by the quality and quantity of training data. Given that most commercial detectors are model-based \cite{prohl2024benchmarking}, establishing a comprehensive benchmark to evaluate detector performance is invaluable for guiding model-based detector development.

Despite offering useful insights, current benchmark frameworks often fail to capture the complexities of real-world text generation, making it difficult to fully assess the true performance of existing detection models in practical applications. Firstly, existing benchmarks primarily focus on the direct testing of detectors, overlooking the significant impact that the training data of model-based detectors can have on their performance. This limits their usefulness for guiding the iterative development of detectors, particularly when data and computational resources are constrained. Secondly, many existing benchmarks focus primarily on the impact of different LLMs or data sources on detector performance, while overlooking the influence of varying text generation tasks. Current evaluations tend to emphasize question-answering tasks, with limited consideration for operations such as text refinement or expansion. These tasks introduce varying levels of LLM involvement, which significantly affects the generated text. Consequently, the varying degrees of LLM participation across tasks pose a critical challenge for detectors, as they must account for these differences to accurately assess performance across a broader range of use cases. Furthermore, current benchmarks are primarily designed for English texts \cite{xu2023generalization, gao2024llm}, failing to address the linguistic diversity and unique challenges presented by other languages, such as Chinese. Given that the leading LLMs are predominantly developed in the United States and China, this limitation in language coverage is particularly problematic. Consequently, there is a critical need for a more comprehensive benchmarking framework that not only addresses the current limitations but also reflects the diverse, real-world contexts in which LLMs are deployed. Such a benchmark must simulate a variety of text generation tasks and account for linguistic diversity across multiple languages. This will facilitate a more accurate and nuanced evaluation of detection models.

To address these limitations, we propose \textbf{CUDRT}, a comprehensive evaluation framework and bilingual benchmark for LLM-generated text detection. CUDRT categorizes LLM-generated text into five key operations: Create, Update, Delete, Rewrite, and Translate. These operations encompass the full range of text generation tasks, from content creation and refinement to translation. To construct CUDRT, we collected pre-LLM human-authored texts from Chinese and English sources and generated corresponding LLM texts using state-of-the-art models \cite{yang2024harnessing}. This benchmark supports a robust evaluation framework for scalable, reproducible experiments, enabling in-depth analysis of how multilingual training, task diversity, and model architectures impact detection performance. By systematically assessing detector performance across these dimensions, CUDRT provides actionable insights for optimizing detection systems and ensuring their adaptability to real-world applications. The primary contributions of this paper are summarized as follows:

\begin{itemize}
\item We propose \textbf{CUDRT}, a bilingual evaluation framework covering five essential LLM operations—\textbf{Create}, \textbf{Update}, \textbf{Delete}, \textbf{Rewrite}, and \textbf{Translate}—to systematically evaluate the reliability of LLM-generated text detectors.

\item We construct a large-scale, high-quality Chinese-English dataset from diverse sources, including news articles and academic papers, to support cross-dataset, cross-operation, and cross-LLM evaluations.

\item We conduct extensive experiments using state-of-the-art detection methods, offering critical insights and guidance for improving detection accuracy, adaptability, and reliability in diverse scenarios.
\end{itemize}

The remainder of this paper is structured as follows. Section \ref{sec2} reviews related work, identifying limitations in existing benchmarks and detection methods. Section \ref{sec3} describes the CUDRT evaluation framework, including its design and data collection process. Section \ref{sec4} presents the experimental setup and evaluates detector performance under various conditions. Finally, Section \ref{sec5} concludes with key findings and future directions.

\section{Related Work}\label{sec2}

\subsection{Overview of Large Language Models}

The evolutionary path of language models began with initial rule-based models, developed into statistical models, followed by early neural language models, then advanced models incorporating pre-training strategies, and has now culminated in today's complex large language models (LLMs) \cite{chang2024survey}. Each generation has significantly pushed the boundaries of natural language processing technology, enhancing both the depth and breadth of language handling capabilities. Currently, LLMs represent the cutting edge in processing and generating natural language text. LLMs can be categorized into open-source and closed-source models based on the availability of their parameters \cite{liang2023uhgeval}. Open-source models, such as Alibaba Cloud's Qwen \cite{bai2023qwen}, Meta's Llama \cite{touvron2023llama}, and Tsinghua University's ChatGLM \cite{glm}, make their code and training methods public, facilitating further development by researchers. Conversely, closed-source models like OpenAI's GPT series \cite{achiam2023gpt} do not disclose their full code or data, instead offering services via APIs.

Given the broad applications of LLMs, both open-source and closed-source models have demonstrated their capabilities in a variety of real-world scenarios, including creative writing, automated customer service, legal document drafting, and scientific research \cite{chang2024survey}. They demonstrate versatility and efficacy in producing coherent, contextually relevant, and high-quality text. For example, Li et al. \cite{li2024flexkbqa} developed the FlexKBQA framework to improve Knowledge Base Question Answering (KBQA) performance despite the lack of high-quality annotated data. Shu et al. \cite{shu2024rewritelm} proposed novel instruction tuning and reinforcement learning strategies to optimize LLM performance on cross-sentence rewriting tasks. Wang et al. \cite{wang2023large} introduced EvLP, a reference-free machine translation evaluation method inspired by post-editing, to assess translations polished by LLMs. Clusmann et al. \cite{clusmann2023future} found that LLMs enhanced nursing outcomes by translating medical terminology into layman's terms and summarizing patient information. Xie et al. \cite{xienext} demonstrated that a few example stories as prompts can enable LLMs to generate new stories rivaling those by human authors. Furthermore, LLMs show significant potential in accessing scientific literature and aiding in programming tasks \cite{nam2024using}. This paper explores how LLMs are applied across various text generation scenarios, examining the differences between open-source and closed-source LLMs.

\subsection{LLM-generated Text Detection Methods}

Due to the increasingly human-like text generated by LLMs, manual differentiation of LLM-generated text is costly and often inaccurate. Researchers are developing targeted LLM-generated text detection algorithms to prevent misuse of these models \cite{sarzaeim2023framework}. These detectors are categorized into metric-based and model-based methods \cite{2023-coco, rouf2024instantops}.

Metric-based methods use quantitative metrics such as textual consistency, complexity, word rank, and entropy to evaluate whether text is LLM-generated \cite{rouf2024instantops}. For example, Gehrmann et al. \cite{gltr} proposed GLTR, which uses statistical approaches to examine the generation probability and entropy of each word. Mitchell et al. \cite{2023detectgpt} introduced DetectGPT, which analyzes the curvature in the model's log probability function to enhance detection accuracy. Su et al. \cite{su2023detectllm} developed zero-shot detection methods DetectLLM-LRR and DetectLLM-NPR to identify LLM-generated text without requiring perturbations. Nevertheless, the robustness of metric-based methods diminishes when applied to texts generated by increasingly sophisticated models, as their rules are predefined and static. This limitation has driven a shift in the development of mainstream commercial detectors, where model-based methods, capable of learning from data and adapting to new patterns, have become the preferred methodology\cite{he-etal-2023-blind}.

Model-based methods train deep learning models to classify texts by learning from both human-generated and LLM-generated data. Guo et al. \cite{guo2023close} proposed the ChatGPT Detector using the RoBERTa architecture, fine-tuned on various scenarios. Wang et al. \cite{seqxgpt} introduced SeqXGPT for sentence-level detection using log probabilities processed through convolutional and self-attention networks. Liu et al. \cite{2023-coco} proposed COCO, which enhances text coherence under low-resource conditions using contrastive learning and an improved loss function for better detection performance. While model-based methods have shown promising results, their performance is not only contingent on the underlying architecture but also heavily influenced by the quality and diversity of the training data, especially under resource-limited conditions. As LLMs continue to evolve at a rapid pace, ensuring the robustness of detection methods in adapting to these advancements poses an even greater challenge. Therefore, a comprehensive evaluation of both the detection performance and the long-term robustness of these models, particularly in the face of ever-improving generative models, remains a critical issue that demands further investigation.

In this study, we conduct a comprehensive evaluation of model robustness across diverse datasets, tasks, and outputs generated by various LLMs. By analyzing these factors, we aim to provide actionable insights that can guide the improvement of model-based detection methods, ensuring their adaptability and performance enhancement in the face of evolving generative technologies.

\subsection{Evaluation Benchmarks for LLM Text Detectors}

In the development and optimization of LLM text detectors, effective performance assessment is critical. Following initial discussions on performance assessment, significant advancements have been made in refining algorithms and enhancing benchmark quality to improve detection accuracy. This section reviews benchmark literature, focusing on the development of benchmarks, tasks involving LLM-generated text, and data sources, with key details summarized in Table \ref{tab1}.

% Table generated by Excel2LaTeX from sheet 'Related work'
\begin{table*}[t]
  \centering
  \caption{Summary of existing LLM-generated text detection benchmarks.}
  \renewcommand{\arraystretch}{1.5}
  \resizebox{\textwidth}{!}{
    \begin{tabular}{p{7.415em}p{11.79em}lp{10.25em}p{13.71em}p{7.79em}p{11.875em}}
    \toprule
    \textbf{Benchmark} & \textbf{Data Composition} & \multicolumn{1}{p{5.79em}}{\textbf{Scale}} & \textbf{Evaluation Metrics} & \textbf{Detectors} & \textbf{LLM Operations} & \textbf{Languages} \\
    \midrule
    M4GTBench \cite{wang2024m4gt} & Student Essays, Peer Reviews & 65,177 & Accuracy, Precision, Recall, F1 & RoBERTa, XLM-R, GLTR, SVM & QA, Create & Arabic, Bulgarian, Chinese, Indonesian, Russian, Urdu, Italian, German \\
    \midrule
    MGTBench \cite{he2023mgtbench} & Essay, Story, Reuters & 3,000 & Accuracy, Precision, Recall, F1, AUC & Log-Likelihood, Rank, DetectGPT, GLTR, Entropy & Create & English \\
    \midrule
    MULTITuDE \cite{macko2023multitude} & News  & 44,786 & Accuracy, Precision, Recall, F1, FPR,FNR, AUC & Roberta, GPT-2, ELECTRA, GLTR, Rank, GPTZero & Create & English, Spanish, Russian \\
    \midrule
    HC3 \cite{guo2023close} & Baike, Reddit, Wiki, Community & 44,425 & F1    & RoBERTa, GLTR & QA    & Chinese, English \\
    \midrule
    HPPT \cite{yang2023chatgpt} & Abstract & 6,050 & Accuracy, AUC & GPTZero, DetectGPT, GPT-2, Roberta & Polish & English \\
    \midrule
    MixSet \cite{gao2024llm} & News, Abstract, Email & 3,600 & Accuracy & Log-Likelihood, GLTR, DetectGPT, Radar, Entropy & Polish, Complete, Rewrite & English \\
    \midrule
    M4 \cite{wang2023m4}    & Wiki, Reddit, Abstract & 122,481 & Accuracy, Precision, Recall, F1 & RoBERTa, GLTR, NELA & Create & English, Chinese, Russian \\
    \midrule
    AIG-ASAP \cite{peng2024hidding} & Students Essays & 1,000 & Accuracy, AUC & RoBERTa, CheckGPT, ArguGPT & Rewrite & English \\
    \midrule
    \textbf{CUDRT(Ours)} & \textbf{News, Paper, Baike, Wiki, Community, Reddit} & \textbf{480,000} & \textbf{Accuracy, Precision, Recall, F1} & \textbf{MPU, Roberta, XLNet} & \textbf{Create, Update, Delete, Rewrite, Translate} & \textbf{Chinese, English} \\
    \bottomrule
    \end{tabular}}
  \label{tab1}%
\end{table*}%

In the domain of LLM-generated text detection, question answering (QA) datasets were initially widely adopted for benchmarks \cite{guo2023close, wang2024m4gt}. By utilizing LLMs to act as question answering experts, researchers created pairs of LLM-generated question and answer datasets. Guo et al. \cite{guo2023close} developed the Human-ChatGPT Comparison Corpus (HC3), which includes about 40,000 questions and responses from human experts and ChatGPT across various domains such as computer science, finance, medicine, law, and psychology. Human responses were collected from publicly available QA datasets with highly rated responses and wiki-based texts. Given the powerful capabilities and widespread application of LLMs in various text generation tasks, researchers are exploring other scenarios to test and improve detector performance. MGTBench \cite{he2023mgtbench} generated essays and news articles using GPT-3.5-Turbo and Claude. Su et al. \cite{su2023hc3} expanded HC3 to HC3 Plus, introducing tasks like translation, summarization, and paraphrasing. Wang et al. \cite{wang2023m4} introduced M4, a multilingual corpus covering various text generation tasks, including news writing, question answering, and academic paper abstracts, reflecting LLM capabilities in different languages. Despite ongoing improvements, current benchmarks for LLM-generated text detectors still fall short in covering the full range of text generation operations, failing to comprehensively encompass the extensive applications of LLMs. Additionally, these benchmarks predominantly rely on direct testing of detectors, a practice that is particularly disadvantageous for model-based methods. Given that these models are trained under varying conditions and with diverse datasets, direct benchmarking often fails to accurately capture their true performance across different scenarios. Such testing provides limited insight into the iterative development and fine-tuning of detectors, making it challenging to offer actionable guidance for enhancing detector robustness and adaptability.

In response to these limitations, this study proposes a novel evaluation framework, CUDRT, which not only encompasses a comprehensive range of LLM operation tasks but also adopts a train-then-test approach for detectors under a unified environment. This framework is designed to evaluate the robustness of detectors across datasets, LLM operations, and LLMs, providing a more nuanced understanding of their adaptability and performance in real-world scenarios.

\begin{figure*}[t]
  \centering
  \includegraphics[width=\textwidth]{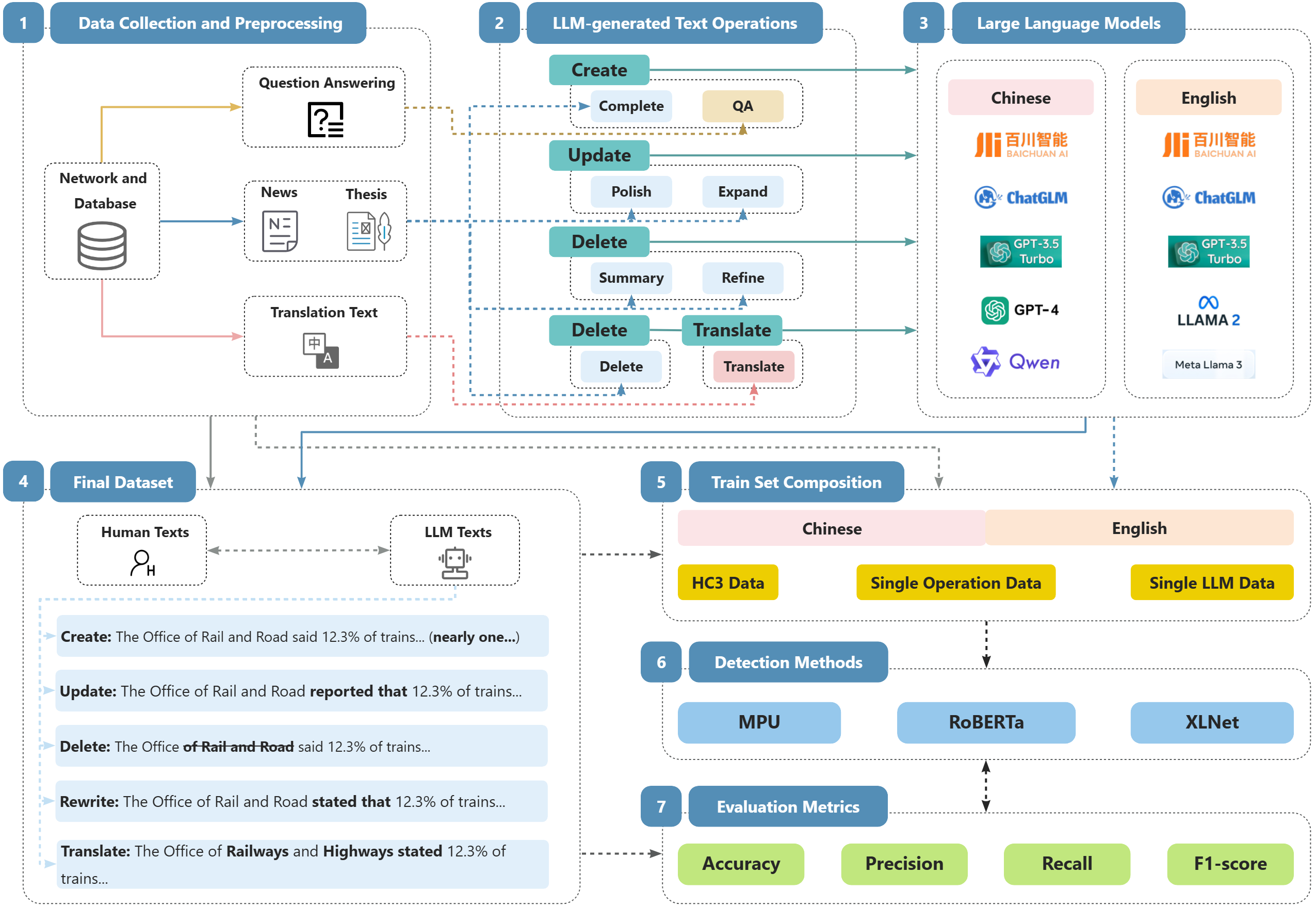}
   \caption{Illustration of the CUDRT evaluation framework.}
   \label{fig1}
\end{figure*}

\section{The CUDRT Evaluation Framework}\label{sec3}

\subsection{Problem Definition}

In this paper, we first curated a comprehensive dataset encompassing various sub-operations of LLM text generation, covering both Chinese and English outputs from diverse LLMs. We then employed widely-used LLM-generated text detectors to assess their performance and generalization abilities across the proposed dataset. The details of the designed CUDRT evaluation framework are illustrated in Figure \ref{fig1}.

Given a dataset $\{(x_i, y_i)\}_{i=1}^N$, where $x_{i}$ represents text content, and $y_i \in \{0, 1\}$ denotes the ground truth label indicating whether the text is LLM-generated, we split the dataset into a training subset $\mathcal{D}_{\text{train}}$ and a testing subset $\mathcal{D}_{\text{test}}$ to evaluate detector generalization. The LLM-generated text detection task is then defined as follows: 
\begin{equation} 
\min_{f} \mathbb{E}_{(x, y) \sim \mathcal{D}_{\text{test}}} \left[ \mathbb{1}\{f(x) \neq y\} \right],
\end{equation} 
where the objective is to find a classifier $f$ that, after training on $\mathcal{D}_{\text{train}}$, minimizes the average misclassification rate on $\mathcal{D}_{\text{test}}$. The indicator function $\mathbb{1}\{f(x_i) \neq y_i\}$ takes the value 1 if the classifier $f$ incorrectly labels the input $x_i$, and 0 otherwise. To capture the distinct distributions, we assume: 
\begin{equation} 
x_i \mid y_i = 0 \sim \mathcal{T}_H \quad \text{and} \quad x_i \mid y_i = 1 \sim \mathcal{T}_A, 
\end{equation} 
where $\mathcal{T}_H$ and $\mathcal{T}_A$ denote the distributions of human-generated and LLM-generated texts, respectively. This formulation enables evaluating the model's performance in detecting LLM-generated text under different distributional conditions, simulating real-world detection scenarios.

\subsection{Data Collection and Preprocessing}

To create our dataset of LLM and human text comparisons, we need to ensure the purity of human texts before generating LLMs texts. Therefore, we selected texts from 2016 and earlier, predating the emergence of LLMs. For ``Complete'', ``Polish'', ``Expand'', ``Summary'', ``Refine'', and ``Rewrite'' operations, we chose news articles and academic papers, as they are widely used and influential fields for LLMs \cite{stokel2023chatgpt, heidt2023artificial}. News articles present facts vividly and straightforwardly \cite{manovich2002language}, while academic papers use specialized terminology for precision \cite{khasawneh2023roles}. This allows for a comprehensive assessment of LLM-generated text detectors in handling different language styles and complexities. However, the tasks of question answering and Chinese-English translation are relatively unique, so we select different human texts from those used in the previous tasks. For QA tasks, we selected data from Community, Wiki, Medicine, and Finance scenarios \cite{guo2023close}, which are prevalent in QA applications \cite{2023evaluating}. For translation, we used ``translation2019zh'' as our source of human texts. For detailed information on the original human texts for each operation, refer to Table A1 in the Appendix.

To ensure data quality and applicability, we filtered 3,000 news articles with word counts between 1,000 and 2,000, and extracted 2,000 segments of academic papers, each around 1,500 words. We removed annotations, special symbols, and references from academic papers, retaining only core content. This aligns with typical human writing conventions. The preprocessing methods for Chinese and English texts were similar. For Arxiv's Latex-formatted texts, we used pydetex.pipelines\footnote{https://github.com/ppizarror/PyDetex} to process equations and removed special symbols while preserving the text. For question answering texts, we selected 1,250 QA pairs from each of four fields and balanced the length of answers.

\subsection{``Create'' Operation in LLM Text Generation}

The ``Create'' operation of LLMs involves generating new content based on user prompts \cite{nam2024using}. This relies on the deep understanding of language and knowledge, utilizing multiple layers of neural networks. By combining learned language patterns and knowledge, LLMs generate grammatically and semantically correct content that is relevant and often creative \cite{liu2024semantic}. We categorize the ``Create'' operation into two sub-operations: ``Complete'' and ``Question Answering''.

\subsubsection{\textbf{``Complete'' as a Sub-operation of ``Create''}}

The ``Complete'' operation refers to LLMs predict subsequent text based on the given text, ensuring that the completed text logically and contextually aligns with the initial input. In the text completion process of LLMs, unfinished text is continued to generate a semantically coherent passage. Providing an incomplete sentence as a prompt to guide the model has a wider range of applications compared to generating complete text directly \cite{marvin2023prompt}. This approach helps the LLMs better grasp the user's specific intent and context, resulting in more relevant and coherent content. Thus, text completion is one of the most commonly used LLM operations.

\begin{figure}[t]
  \centering
  \includegraphics[width=10cm]{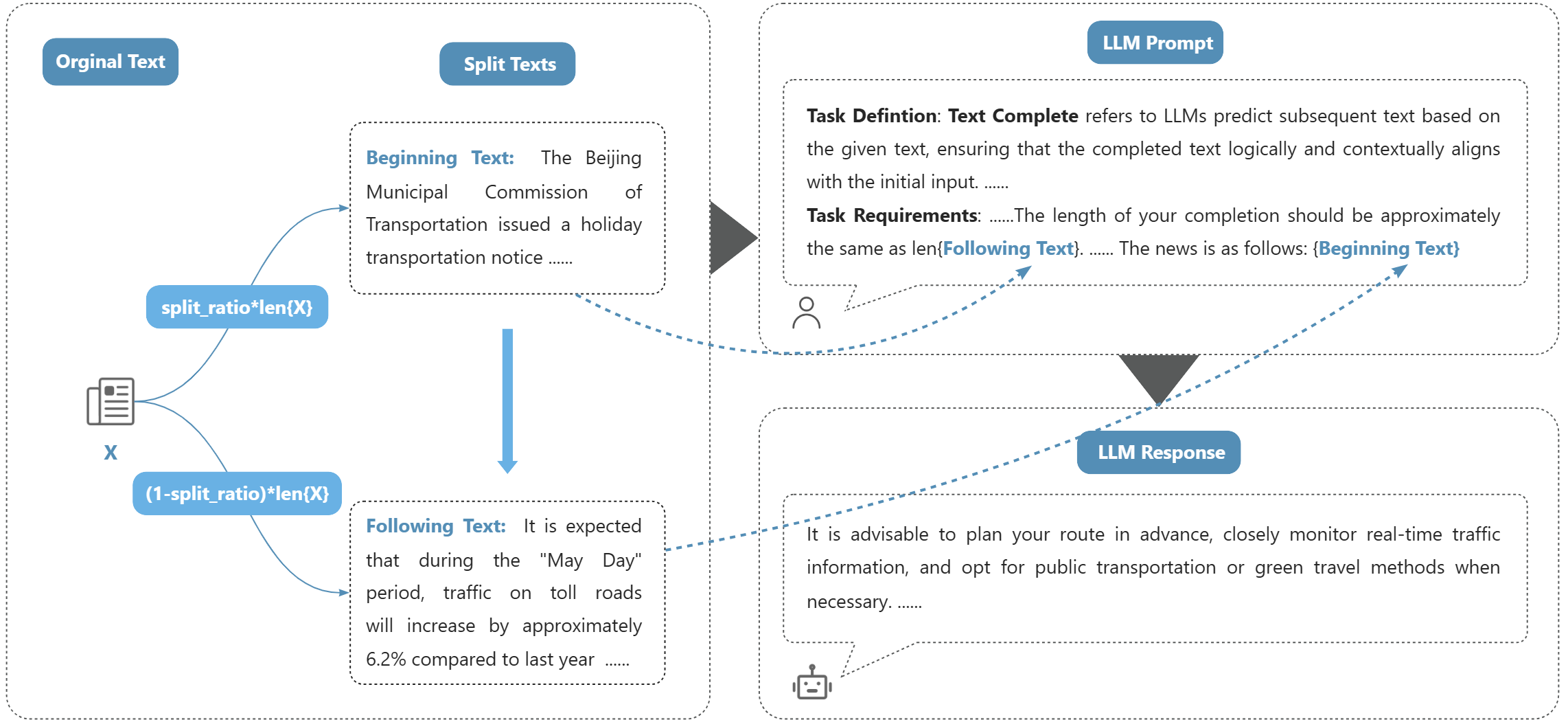}
   \caption{The process of generating text through the ``Complete'' operation of LLMs.}
   \label{fig2}
\end{figure}

When performing ``Complete'' operations, we established three completion ratios: 25\%, 50\%, and 75\%. We used collected news and academic texts, truncating them to these ratios: 75\%, 50\%, and 25\%. LLMs then completed the truncated portions to match the original text's character count, thereby ensuring a consistent basis for fair detection of LLM-generated text. Figure \ref{fig2} illustrates this process. For example, with a 50\% news completion ratio, we split the text into two parts: beginning and following text, with equal word counts. The prompt includes task definition and requirements, specifying the beginning text and the number of words to be completed, ensuring the LLMs understand and perform the task effectively. Additionally, we ensure the completed text adheres to the norms and style of news texts. The process for completing academic papers is similar.

\subsubsection{\textbf{``Question Answering'' as a Sub-operation of ``Create''}}

The ``Question Answering'' operation refers to that LLMs act as an expert, providing a detailed answer to the given question. Human utilization of machine-based question answering has become widespread. From early rule-based systems to context-based querying and now advanced systems built on deep learning and LLMs, the evolution has been remarkable \cite{ni2023recent}. These systems have demonstrated significant value across various domains, including education, healthcare, and finance, providing accurate and timely answers \cite{2023revolutionizing}. Currently, LLM-based question answering systems can essentially replace humans in answering diverse queries, offering accurate and rapid responses in both professional and everyday contexts.

When creating the LLMs' question answering dataset, we covered four domains: Community, Wiki, Medicine, and Finance, encompassing mainstream aspects of daily life. This selection builds a diverse and practical question answering dataset (see section 3.2 for details). Some questions involve complex knowledge and privacy concerns \cite{chen2023can}, which may lead to lower quality answers from LLMs \cite{bai2024benchmarking}. To address this, we instruct LLMs to assume specific roles in prompts. For example, for medical questions: ``Imagine you are a doctor specialized in answering patient inquiries. Provide detailed answers based on your knowledge base, ensuring accuracy''. Adjusting roles for other domains improves response willingness and answer quality, aligning closely with human reasoning.

\subsection{``Update'' Operation in LLM Text Generation}

In this section, we define the ``Update'' operation for LLM-generated text, which involves modifying existing text to improve clarity, completeness, or content richness. We categorize these modifications into ``Polish'' and ``Expand'' operations.

\subsubsection{\textbf{``Polish'' as a Sub-operation of ``Update''}}

The ``Polish'' operation refers to the process where, given input text, the LLMs improve the quality, fluency, and accuracy of the given text to better adhere to language conventions and reader expectations. Text polishing is a common task for humans using LLMs after drafting \cite{zhang2023visar}. LLMs help improve text quality by correcting spelling and grammatical errors, suggesting alternative vocabulary, optimizing sentence structures, and adjusting tone and style, making the text more engaging and readable. In typical scenarios, LLMs serve as auxiliary tools, with LLM-generated text often interspersed within human-generated text. This raises higher requirements for detecting LLM text.

\begin{figure}[t]
  \centering
  \includegraphics[width=10cm]{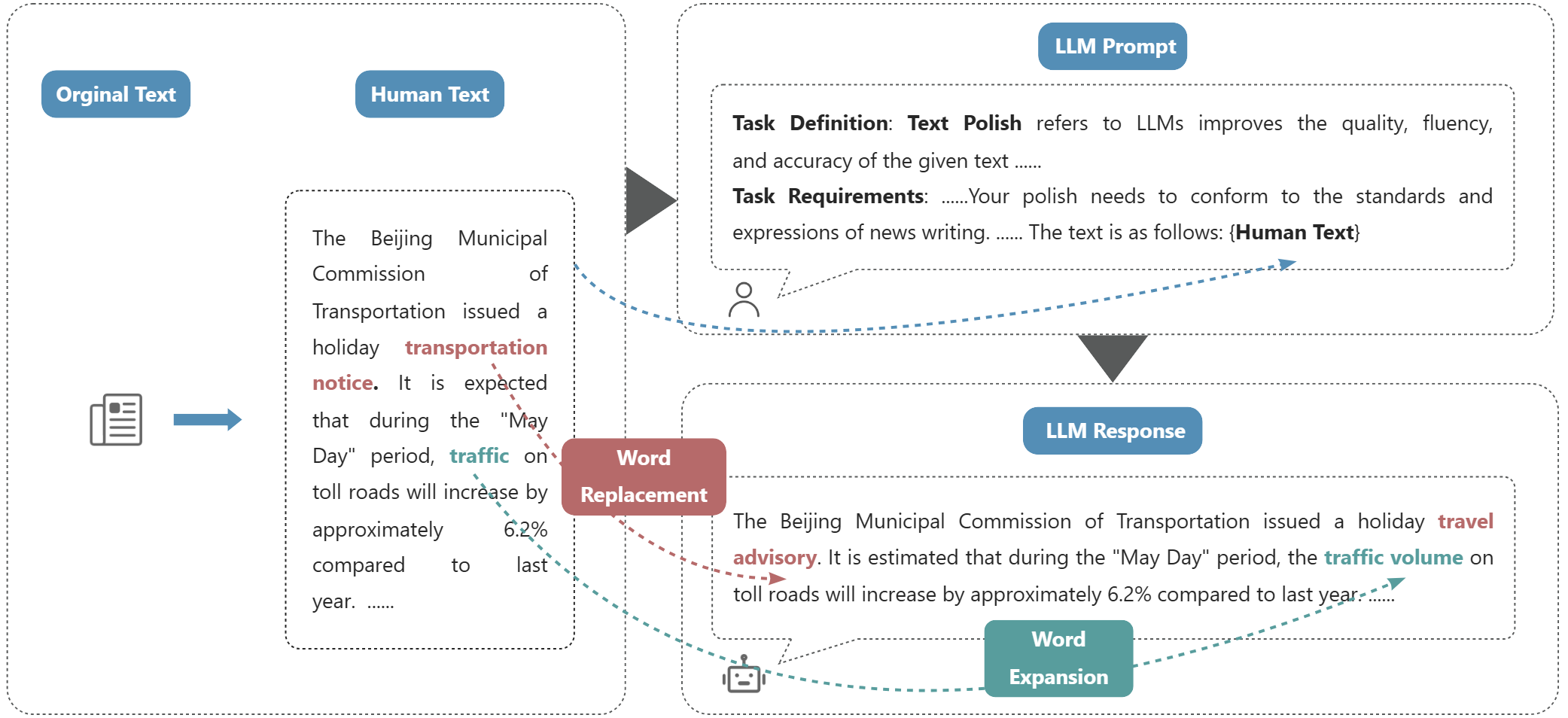}
   \caption{The process of generating text through the ``Polish'' operation of LLMs.}
   \label{fig3}
\end{figure}

Figure \ref{fig3} shows the process of using LLMs for text polishing. The ``Polish'' operation prompts include task definition and task requirements to help LLMs understand and execute tasks better. The figure illustrates changes before and after polishing a news excerpt, showing improvements in word replacement and expansion for higher quality, more fluent text. It's important to note slight differences between polishing news articles and academic papers, necessitating field-specific terminologies in the prompts.

\subsubsection{\textbf{``Expand'' as a Sub-operation of ``Update''}}

The ``Expand'' operation refers to the process where, given an input text, the LLMs generate new text related to the original content but in greater detail, richness, or with more expressed nuances. This operation aims to enrich existing text by making it more comprehensive and detailed. LLMs can add new viewpoints, examples, and details while maintaining style and semantic coherence, resulting in more substantial and persuasive content. Unlike text completion, which fills in missing parts, text expansion focuses on adding additional information to enhance the text's persuasiveness. It involves introducing new viewpoints and providing more examples or details.

\begin{figure}[t]
  \centering
  \includegraphics[width=10cm]{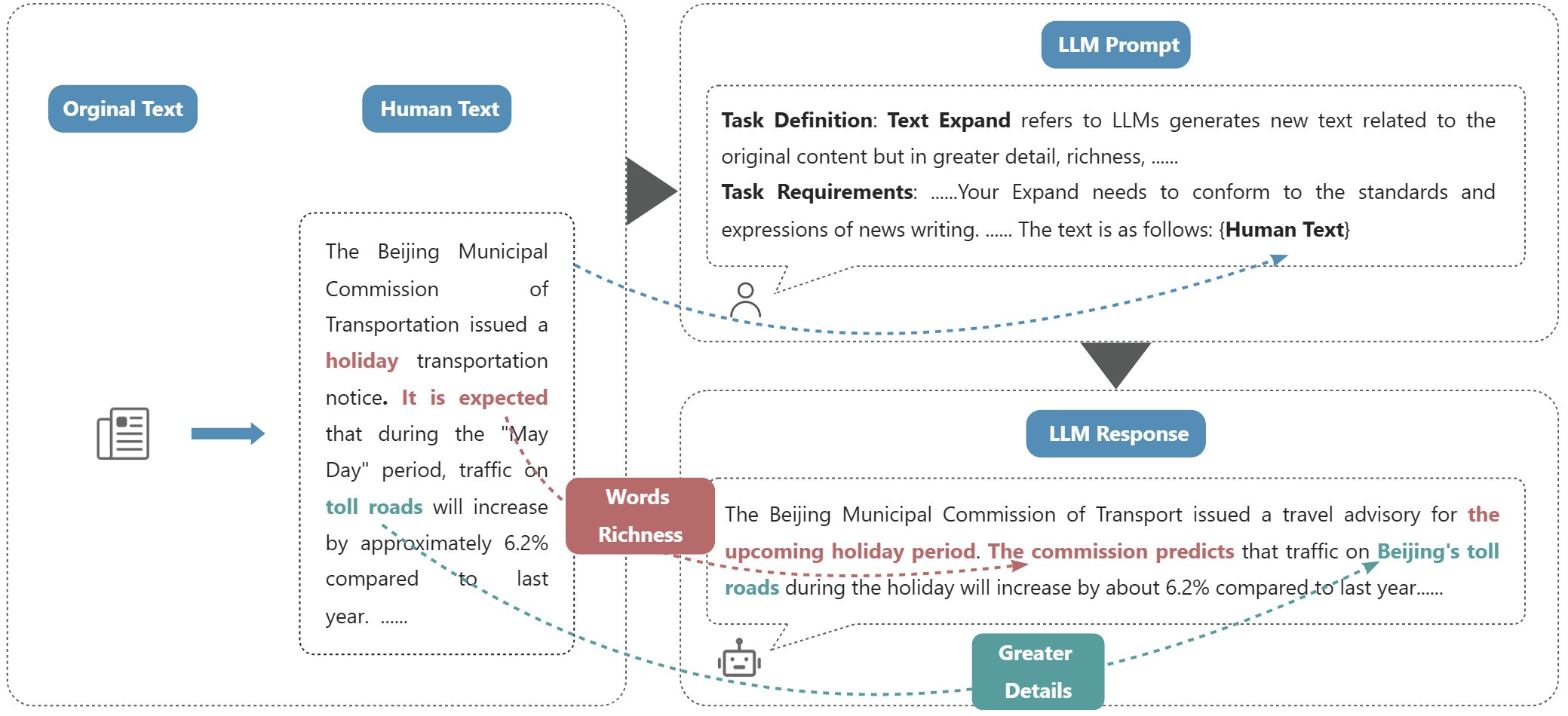}
   \caption{The process of generating text through the ``Expand'' operation of LLMs.}
   \label{fig4}
\end{figure}

Similar to text polishing, text expansion focuses on enriching text at the word level. Figure \ref{fig4} shows the process of using LLMs for text expansion, highlighting the main functions of adding richer words and greater details. It's also necessary to prompt LLMs to recognize the differences between news and academic papers, as their linguistic styles vary during text expansion.

\subsection{``Delete'' Operation in LLM Text Generation}

In modern text processing, the ``Delete'' operation of LLMs effectively identifies and removes redundant text through a deep understanding of the content \cite{kaddour2023challenges}. This semantic-based editing enhances conciseness and readability, not just simple character deletion. LLMs analyze text structure and coherence to identify unnecessary repetitions, overly detailed sections, or deviations from the main topic. They propose modifications or directly delete parts to make the text more compact and clear, while maintaining information integrity and fluency. We categorize the ``Delete'' operation into ``Summary'' and ``Refine'' based on the extent of text deletion.

\subsubsection{\textbf{``Summary'' as a Sub-operation of ``Delete''}}

The ``Summary'' operation refers to the process where, after inputting text, the LLMs summarize the given text into a concise and accurate abstract, highlighting the main information and key points. This improves reading efficiency and makes information delivery more direct and focused. ``Summary'' operations are particularly useful for processing large amounts of data or lengthy articles, supporting quick decision-making and knowledge absorption \cite{hu2022can}. This technology is widely applied in mainstream news websites and information-intensive fields \cite{guo2021conditional}. However, as its use proliferates, identifying LLM-generated summaries has become essential to ensure information source and reliability.

\begin{figure}[t]
  \centering
  \includegraphics[width=10cm]{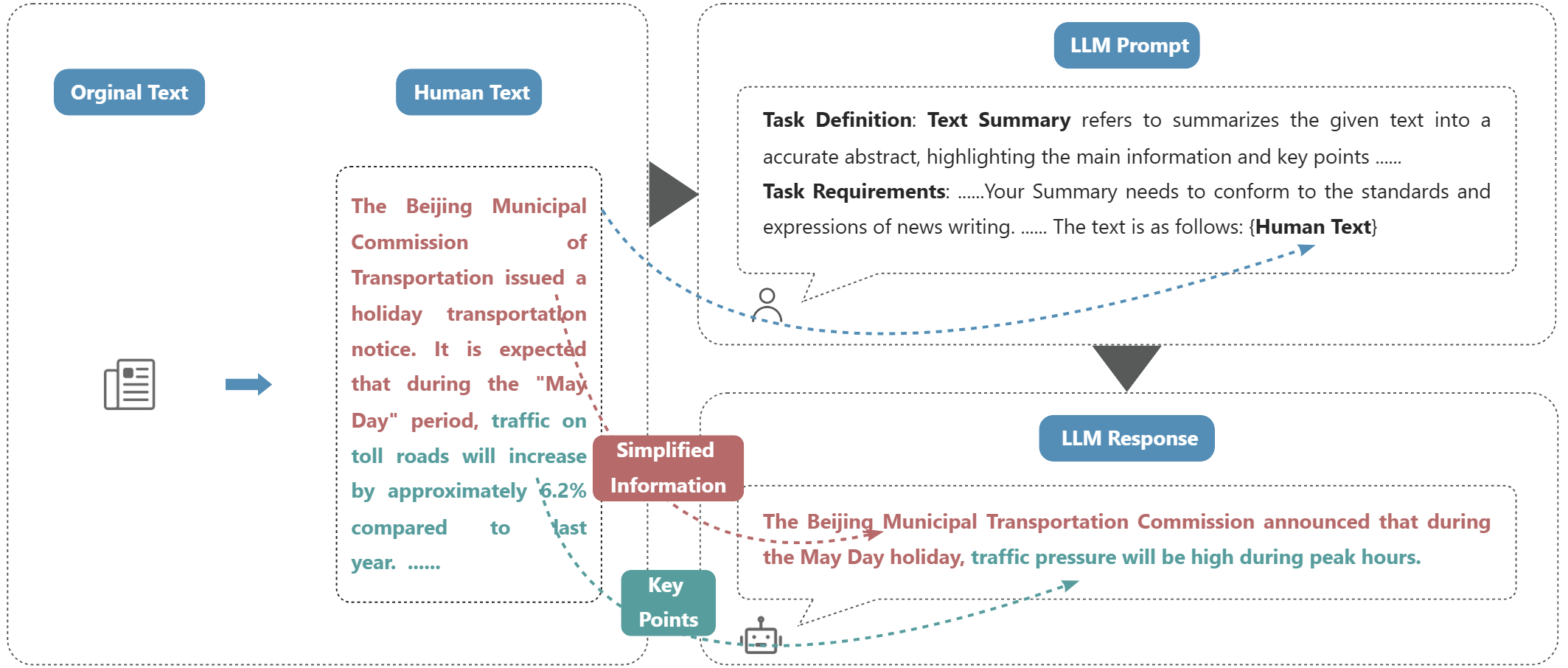}
   \caption{The process of generating text through the ``Summary'' operation of LLMs.}
   \label{fig5}
\end{figure}

Due to token limitations of LLMs, entire papers are usually not input for summarization, so we focus on news summaries. Figure \ref{fig5} illustrates the process using a news excerpt. To ensure fairness in detection experiments, we calculated the word count of human-generated summaries and specified that LLMs generate summaries with similar word counts. This allows accurate comparison between human and LLM-generated summaries, avoiding length differences. Figure \ref{fig5} shows that LLMs reduce redundant information and summarize key points, enhancing text fluency, readability, and accurate information conveyance.

\subsubsection{\textbf{``Refine'' as a Sub-operation of ``Delete''}}

The ``Refine'' operation refers to the process where, after inputting text, the LLM simplifies, compresses, or refines the given text into a form that is more concise, clear, and easy to understand. By identifying and removing unnecessary repetitions, verbose sections, or irrelevant parts, the model improves text flow and readability while retaining key information \cite{shi2023chatgraph}. This enhances text quality and precision of information delivery, especially in business reports or research papers. Refinement prevents readers from losing focus due to lengthy expressions and avoids ambiguity, making it easier to grasp core ideas. Compared to ``Summary'', ``Refine'' involves less deletion, retaining most information and only removing redundant parts for a more concise and fluid text.

\begin{figure}[t]
  \centering
  \includegraphics[width=10cm]{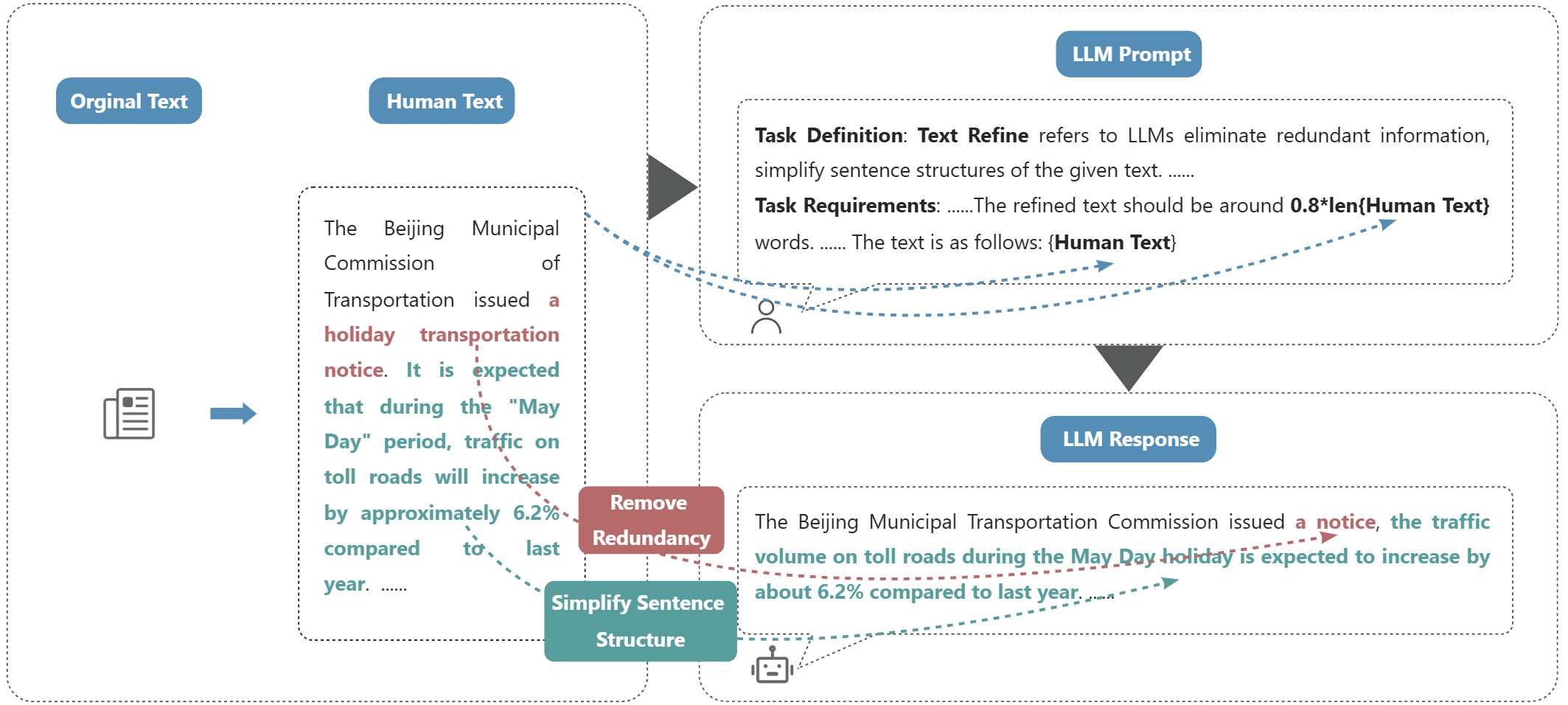}
   \caption{The process of generating text through the ``Refine'' operation of LLMs.}
   \label{fig6}
\end{figure}

Figure \ref{fig6} illustrates the ``Refine'' operation. When creating refined text, we specify that the output length should be between 80\% and 90\% of the original text to avoid it resembling a summary. This aligns with the typical understanding of text refinement. The figure shows detailed changes, including removal of redundancy and simplification of sentence structures. Furthermore, we produce news and paper refinements separately to align with the semantic styles of each field.

\subsection{``Rewrite'' Operation in LLM Text Generation}

The ``Rewrite'' operation refers to the process in which, upon receiving a text input, LLMs adjust word selection and reorganize paragraphs to change the style of the given text, thereby producing a new text that is close in meaning but different in linguistic expression from the original. This enhances fluidity, readability, and stylistic diversity, making the language more vivid or suited to specific requirements. During rewriting, synonyms can be substituted, sentence structures adjusted, and semantics strengthened while maintaining core information and themes. This is useful for academic papers, business copy, or any text requiring style adjustment. Compared to the ``Polish'' operation, which involves minor adjustments like correcting grammatical errors and optimizing word choice, ``Rewrite'' involves more profound changes, such as reorganizing entire sections, rearticulating themes, and altering viewpoints or tones.

\begin{figure}[t]
  \centering
  \includegraphics[width=10cm]{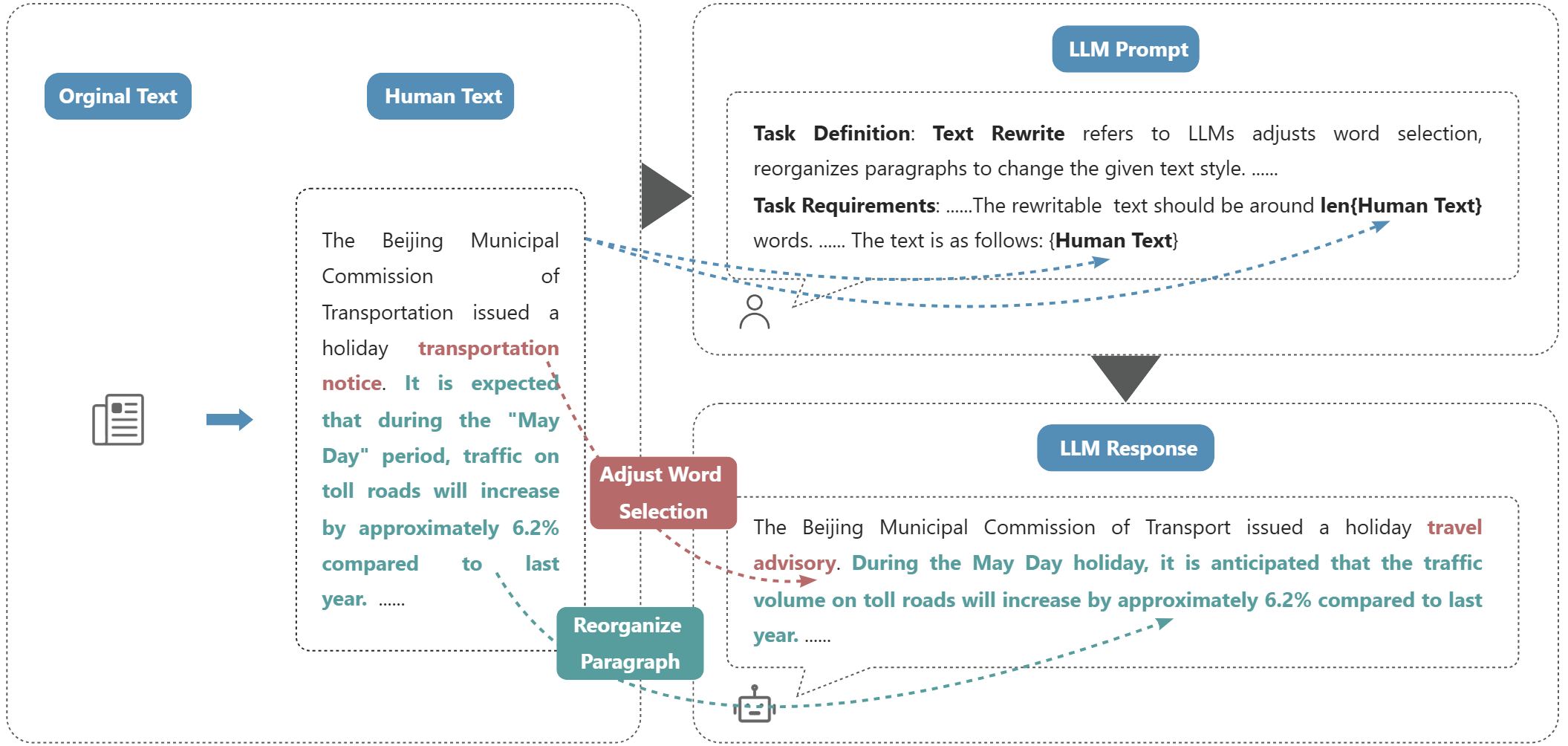}
   \caption{The process of generating text through the ``Rewrite'' operation of LLMs.}
   \label{fig7}
   \vspace{-\baselineskip}
\end{figure}

Figure \ref{fig7} details the ``Rewrite'' operation. To ensure fairness in detection experiments, the word count of the LLM-rewritten text is kept approximately the same as the human text. Using a news excerpt, the figure illustrates the rewriting process at the word and sentence levels. By adjusting word selection and reorganizing paragraph structures, the LLM transform the language style, offering a completely new reading experience.

\subsection{``Translate'' Operation in LLM Text Generation}

The use of LLMs for text translation is a core function. These models capture context, tone, and cultural nuances, making translations more natural and accurate. Their ongoing learning ensures continuous improvement, adapting to new linguistic usages and challenges. The ``Translate'' operation has matured, with translated texts often indistinguishable from human translations. For our dataset, we focused on English-Chinese and Chinese-English translations, having the LLMs translate as needed.

% Table generated by Excel2LaTeX from sheet 'LLM'
\begin{table*}[t]
  \centering
  \caption{Specific parameter details for Chinese and English LLMs.}
  \begin{threeparttable}
  \renewcommand{\arraystretch}{1.2}
  \resizebox{12cm}{!}{
    \begin{tabular}{lllll|lllll}
    \toprule
    \textbf{Chinese LLMs} & \textbf{Parm.} & \textbf{Type} & \textbf{Publisher} & \textbf{Release} & \textbf{English LLMs} & \textbf{Parm.} & \textbf{Type} & \textbf{Publisher} & \textbf{Release} \\
    \midrule
    Baichuan2 \cite{yang2023baichuan} & 13B   & Chat  & Baichuan Inc. & 2023.09 & Baichuan2 \cite{yang2023baichuan} & 13B   & Chat  & Baichuan Inc. & 2023.09 \\
    ChatGLM3-32K \cite{zeng2022glm} & 6B    & Chat  & Tsinghua & 2024.01 & ChatGLM3-32K \cite{zeng2022glm} & 6B    & Chat  & Tsinghua & 2024.01 \\
    GPT-3.5-Turbo \cite{zhao2023survey} & 175B  & Chat  & OpenAI & 2023.03 & GPT-3.5-Turbo \cite{zhao2023survey} & 175B  & Chat  & OpenAI & 2023.03 \\
    GPT-4-1106 \cite{achiam2023gpt} & NaN & Chat  & OpenAI & 2023.11 & Llama3 \cite{llama3} & 8B    & Chat  & Meta  & 2024.04 \\
    Qwen1.5 \cite{bai2023qwen} & 32B   & Chat  & Alibaba & 2024.04 & Llama2 \cite{touvron2023llama} & 13B   & Chat  & Meta  & 2023.07 \\
    \bottomrule
    \end{tabular}}
    \begin{tablenotes}
        \footnotesize
        \item \textit{Notes}: NaN denotes no public data available. 
    \end{tablenotes}
  \end{threeparttable}
  \label{tab2}
  \vspace{-\baselineskip}
\end{table*}%

\subsection{Large Language Models for Text Generation}

Our benchmark includes both Chinese and English materials, so we selected advanced LLMs from China and the U.S. This ensures our dataset keeps pace with the latest technological developments and maintains high-quality texts. The selected LLMs include the GPT, Llama, Qwen, ChatGLM, and Baichuan series. Specific model details are shown in Table \ref{tab2}. GPT-4 and Llama3 represent state-of-the-art developments in closed-source and open-source LLMs \cite{zhao2024explainability}, respectively. For our dataset, we used GPT-4-1106-preview for generating Chinese text and Llama-3-8b-Instruct for English text, due to Llama-3-8b-Instruct's suboptimal performance in Chinese. GPT-3.5-Turbo was also used for both languages due to its versatility and wide user base. For Chinese text, we selected Qwen1.5-32B-Chat from Alibaba, as it performs poorly in English. Before Llama3, Llama2 was the most commonly used open-source English model, used here exclusively for English text. ChatGLM3-6b-32k, developed by Tsinghua University and Zhichart AI, was used for both languages. Finally, we employed the Baichuan2-13B chat model from ``Baichuan Intelligence'' for its multilingual capabilities.

\subsection{LLM-generated Text Detection Methods}

To detect LLM-generated text in both Chinese and English, we employ two types of detectors: model-based and metric-based methods. The model-based method relies on deep learning models trained on extensive data to identify LLM-specific features. The metric-based method uses predefined rules or indicators, such as grammar, vocabulary usage, and logical coherence, for quick identification. Integrating these methods provides a multidimensional analysis, enhancing detection accuracy and reliability.

\subsubsection{\textbf{Metric-based Method}}

Most metric-based models are primarily designed for detecting LLM-generated text in English and often lack effectiveness. In contrast, Multiscale Positive-Unlabeled (MPU) excels in detection and supports both Chinese and English, making it suitable for comprehensive language detection. Therefore, we chose MPU as our metric-based detection method.

\textbf{Multiscale Positive-Unlabeled} \cite{tian2023multiscale}. MPU treats LLM-generated text detection as a partial Positive-Unlabeled (PU) problem. It uses a length-sensitive Multiscale PU loss function and an abstract recurrent model to estimate prior probabilities of positive samples across different text lengths. Additionally, a Text Multiscaling module generates texts of varying lengths through random sentence deletion, enhancing the model's ability to detect short texts while maintaining effectiveness for long texts.

\subsubsection{\textbf{Model-based Method}}

For the model-based method, we select RoBERTa and XLNet, foundational architectures for many language models, excelling in handling complex linguistic structures and context dependencies.

\textbf{RoBERTa} \cite{liu2019roberta}. This classifier, based on the pretrained RoBERTa model, has been fine-tuned for detecting LLM-generated text. It improves on BERT by enhancing data handling and model configuration during training. It is used to detect both Chinese and English texts in this study.

\textbf{XLNet} \cite{yang2019xlnet}. XLNet uses a permutation language model to consider all possible word order combinations during training, addressing BERT's inconsistency issues. It also integrates Transformer-XL technology, enhancing its ability to handle long-range dependencies. It is also used to detect both Chinese and English texts.

\begin{table}[t]
  \centering
  \caption{Operation composition statistics of the proposed dataset.}
  \begin{threeparttable}
  \renewcommand{\arraystretch}{1.0}
  \resizebox{\textwidth}{!}{
    \begin{tabular}{p{5.835em}llp{16.085em}p{26.165em}}
    \toprule
    \multicolumn{2}{p{15.335em}}{\textbf{LLM-generated Text Operations}} & \multicolumn{1}{p{5.085em}}{\textbf{Scale}} & \textbf{Text Composition} & \textbf{Task Description} \\
    \midrule
    \multirow{2}[4]{*}{Create} & \multicolumn{1}{p{9.5em}}{Complete} & 5000  & News, Thesis & LLMs predict subsequent text based on the given partial text, ensuring that the completed text logically and contextually aligns with the initial input. \\
\cmidrule{2-5}    \multicolumn{1}{l}{} & \multicolumn{1}{p{9.5em}}{Question Answering} & 5000  & Baike, Community, Medicine, Finance & LLMs act as an expert, providing a detailed answer to the following question. \\
    \midrule
    \multirow{2}[4]{*}{Update} & \multicolumn{1}{p{9.5em}}{Polish} & 5000  & News, Thesis & LLMs improve the quality, fluency, and accuracy of the given text to make it more in line with linguistic standards and reader expectations \\
\cmidrule{2-5}    \multicolumn{1}{l}{} & \multicolumn{1}{p{9.5em}}{Expand} & 5000  & News, Thesis & LLMs generate new text related to the original content but in greater detail, richness, or with more expressed nuances. \\
    \midrule
    \multirow{2}[4]{*}{Delete} & \multicolumn{1}{p{9.5em}}{Summary} & 3000  & News  & LLMs summarize the given text into a concise and accurate abstract, highlighting the main information and key points of the text. \\
\cmidrule{2-5}    \multicolumn{1}{l}{} & \multicolumn{1}{p{9.5em}}{Refine} & 5000  & News, Thesis & LLMs simplify, compress, or refine the given text into a form that is more concise, clear, and easy to understand. \\
    \midrule
    \multicolumn{2}{p{15.335em}}{Rewrite} & 5000  & News, Thesis & LLMs rephrase the content of the given text to produce a new text that is semantically close to the original but differs in linguistic expression. \\
    \midrule
    \multicolumn{2}{p{15.335em}}{Translate} & 5000  & News  & LLMs are primarily used to perform English-Chinese translation. \\
    \bottomrule
    \end{tabular}}
    \begin{tablenotes}
        \footnotesize
        \item \textit{Notes}: The dataset size per task is for a single language. Multiply by two for the total in Chinese and English. 
    \end{tablenotes}
  \end{threeparttable}
  \label{tab3}%
\end{table}%

\subsection{Dataset Statistics}

Our proposed dataset is generated through a combination of preprocessing and various LLM operations. It integrates multiple operations such as Create, Update, Delete, Rewrite, and Translate, where most tasks—except for Question Answering and Translate—are derived from a shared core dataset of news and thesis texts. This approach ensures that LLMs are evaluated consistently across operations like text complete, polish, expand, and summary. Specific dataset details can be found in Table \ref{tab3}. With 5,000 samples per task per language (Chinese and English), the dataset provides a comprehensive resource for robust model training and evaluation across diverse text generation and manipulation tasks. 

\begin{figure*}[t]
    \centering
    \subfigure[Chinese Complete]{
        \includegraphics[width=0.23\textwidth]{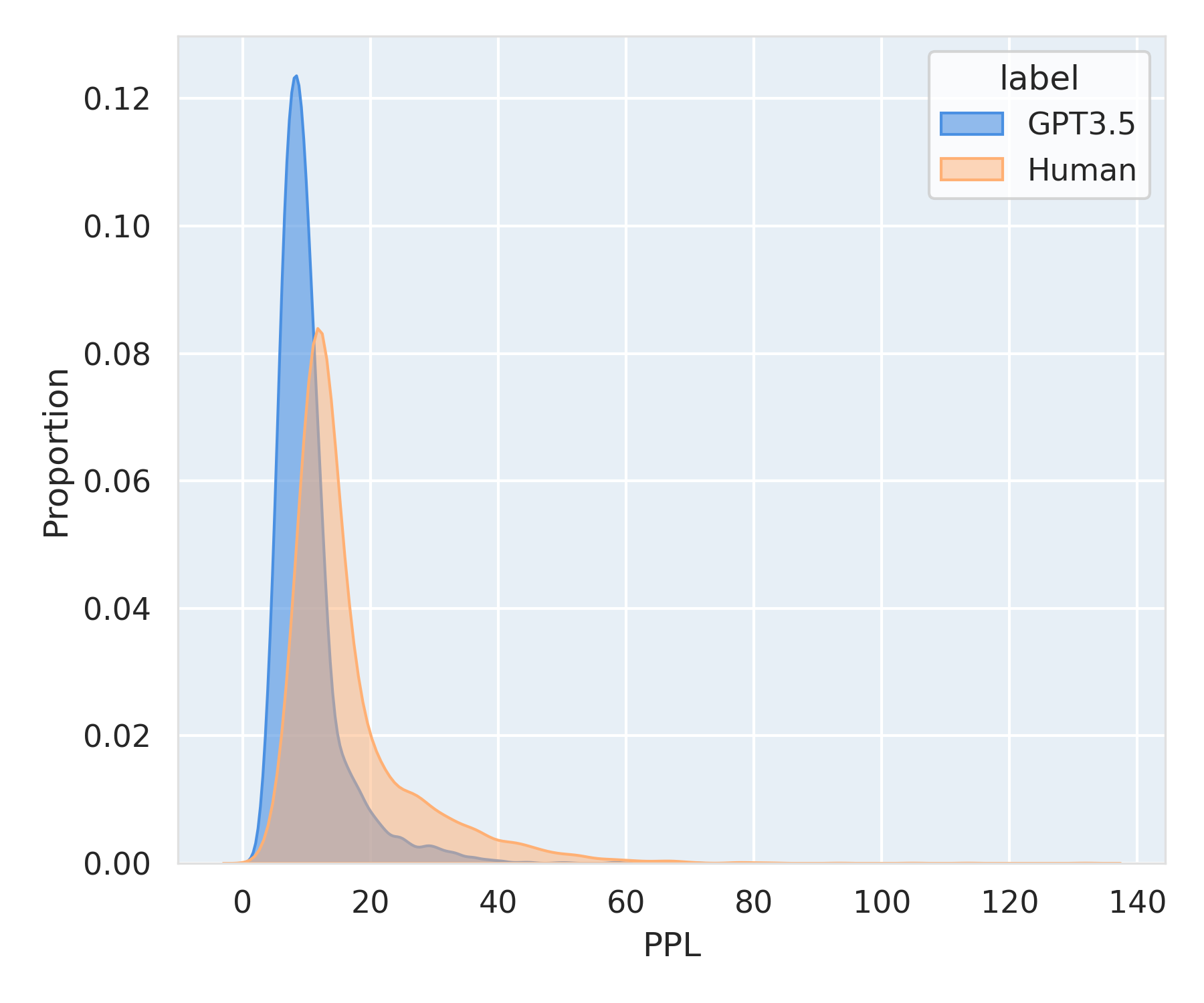}
    }
    \subfigure[Chinese QA]{
        \includegraphics[width=0.23\textwidth]{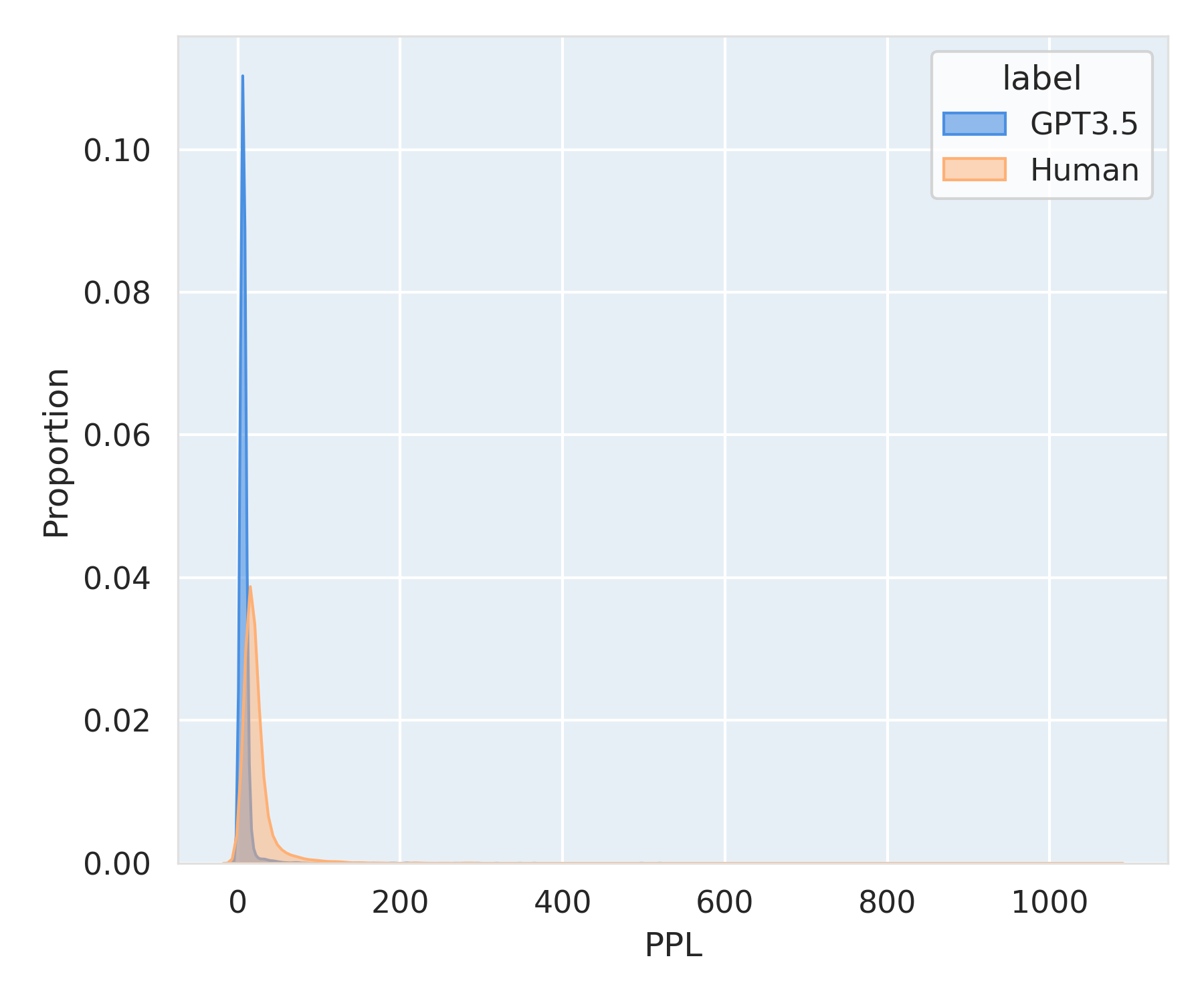}
    }
    \subfigure[Chinese Polish]{
        \includegraphics[width=0.23\textwidth]{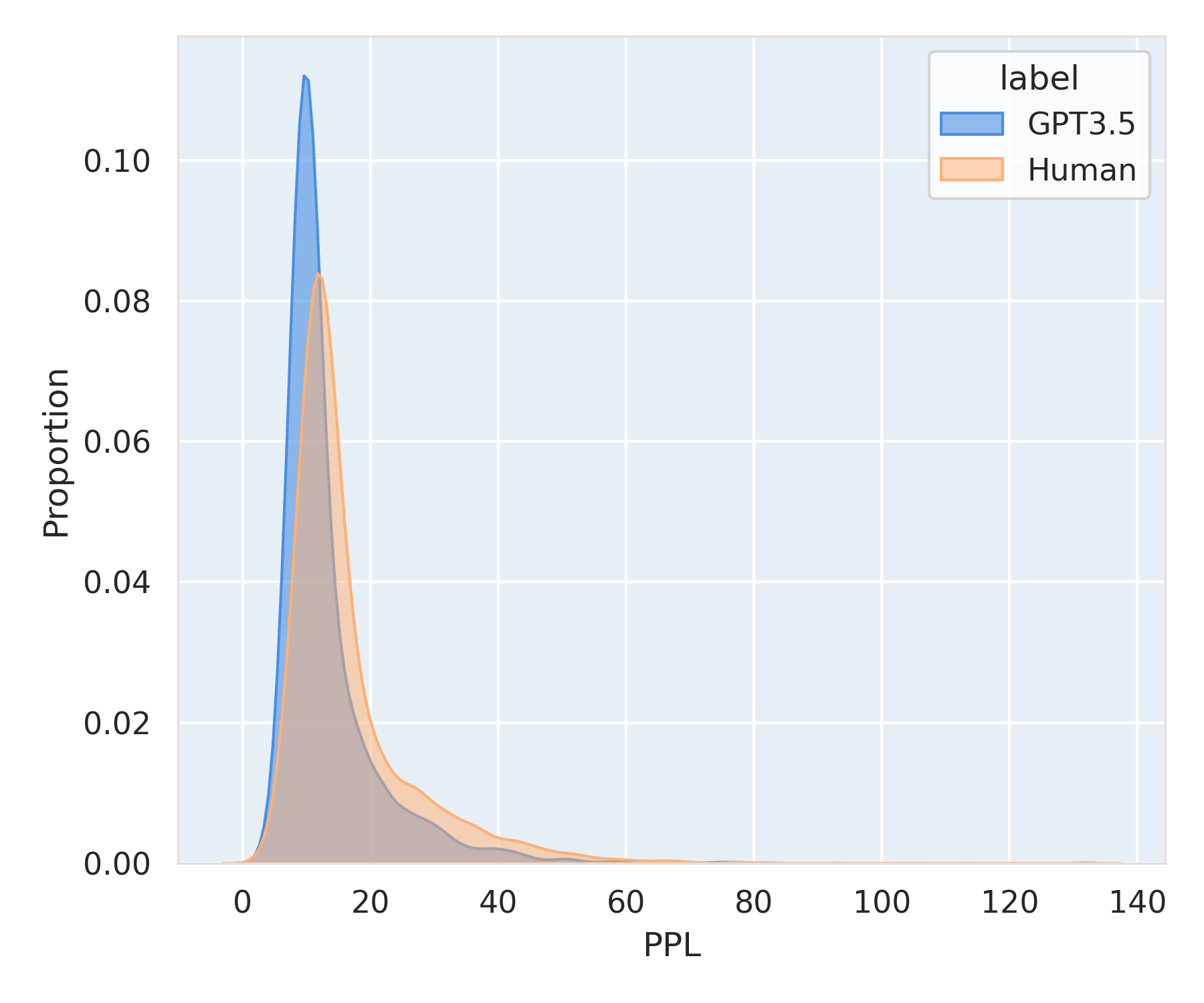}
    }
    \subfigure[Chinese Expand]{
        \includegraphics[width=0.23\textwidth]{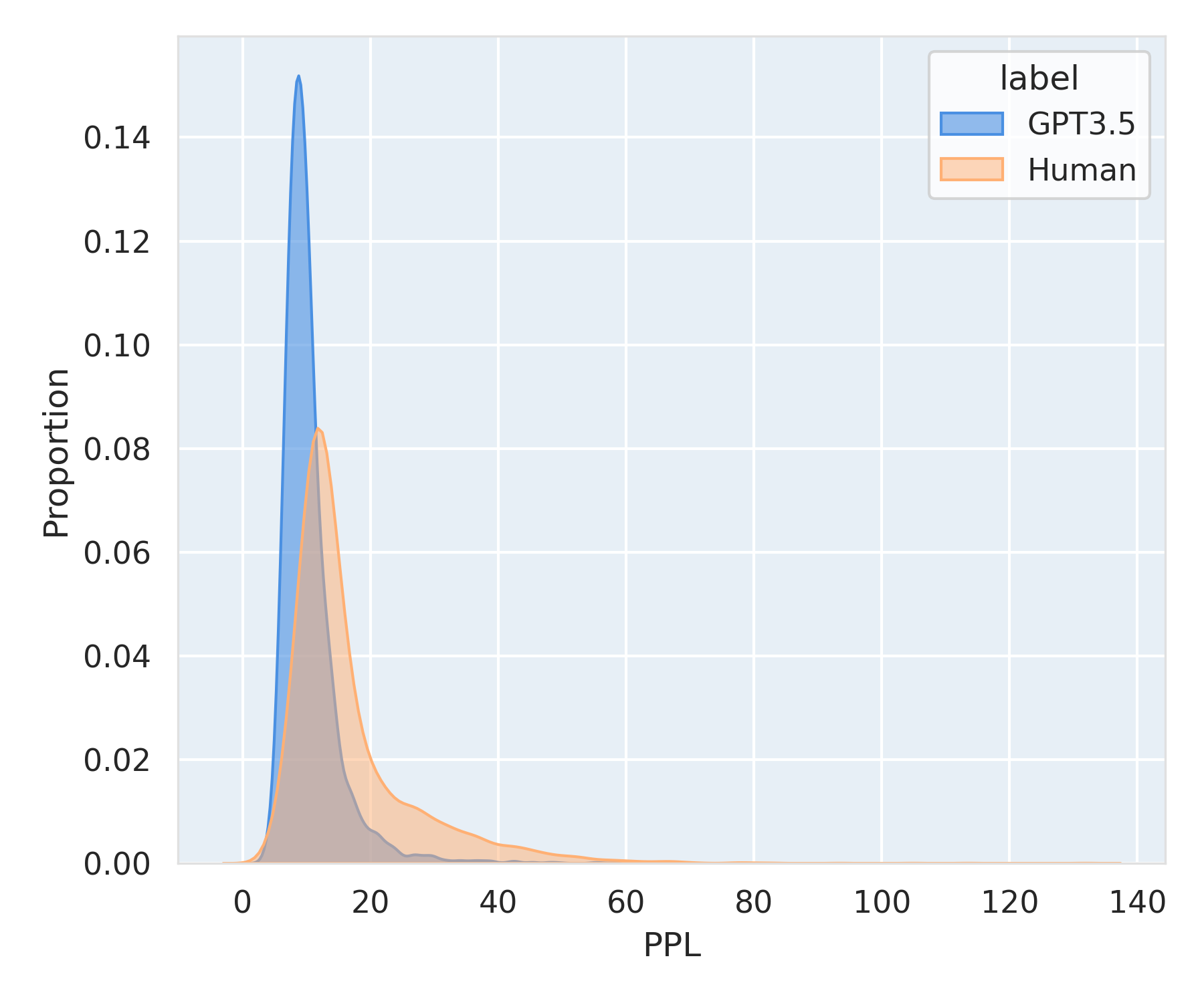}
    }
    
    \subfigure[Chinese Summary]{
        \includegraphics[width=0.23\textwidth]{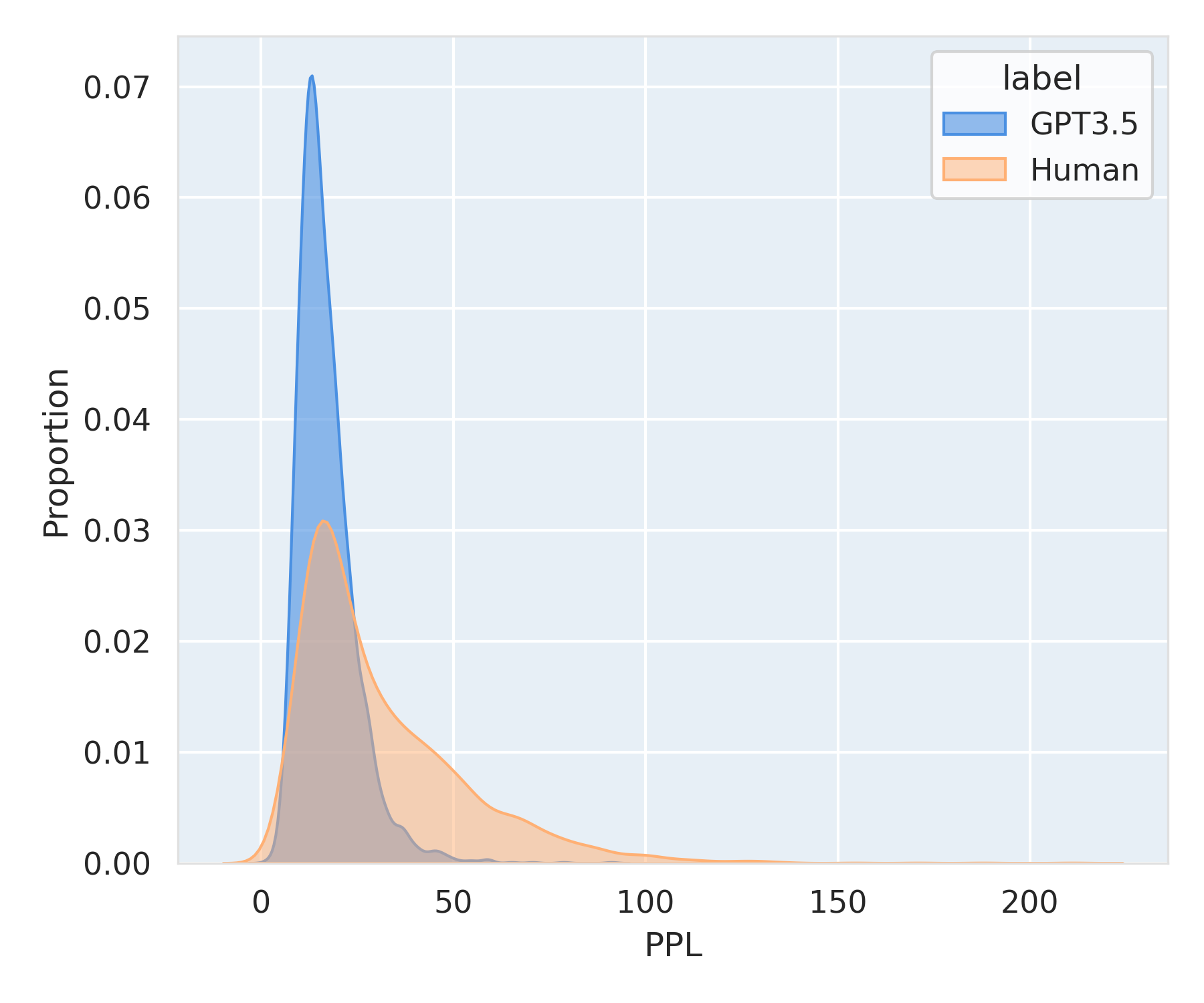}
    }
    \subfigure[Chinese Refine]{
        \includegraphics[width=0.23\textwidth]{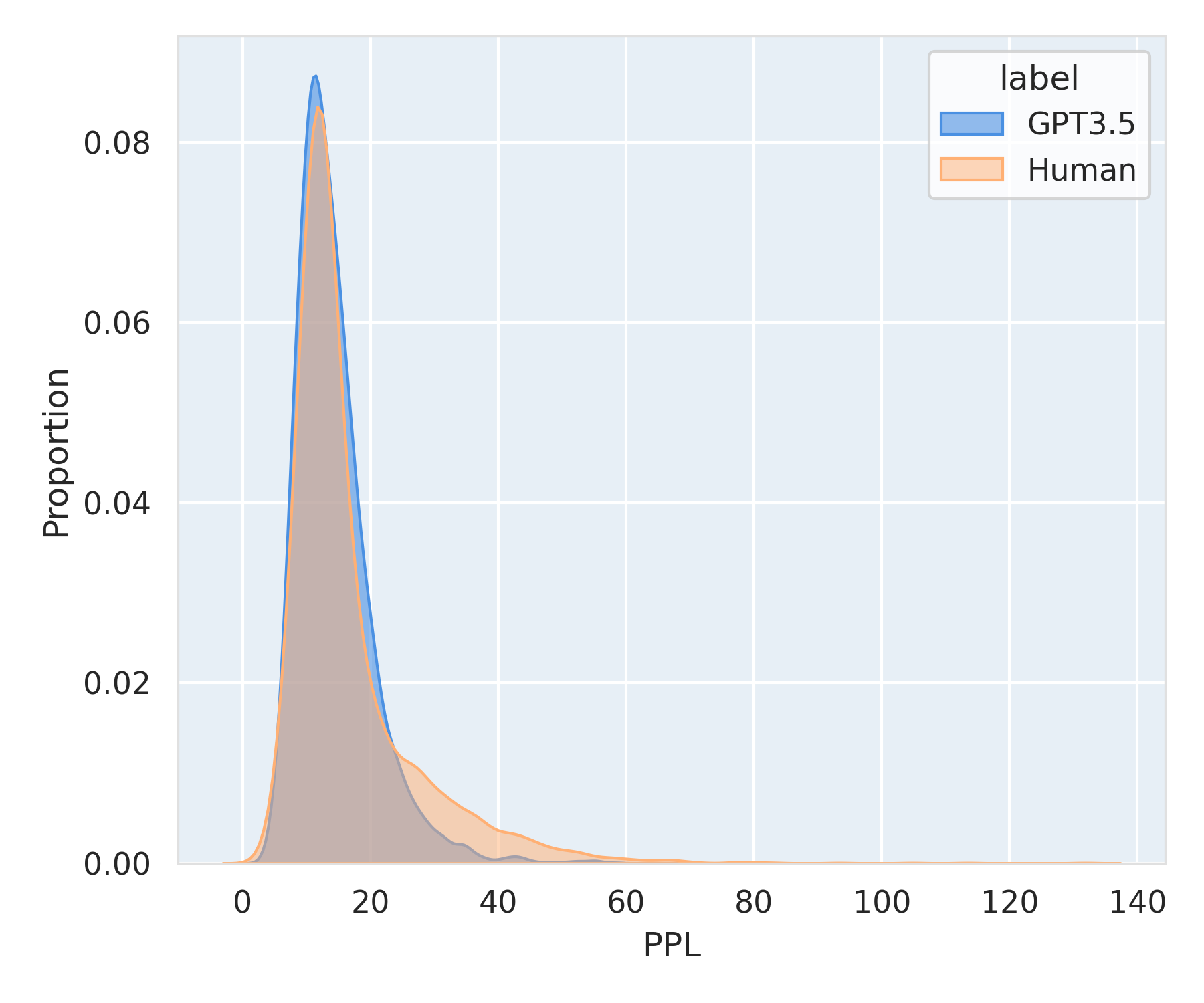}
    }
    \subfigure[Chinese Rewrite]{
        \includegraphics[width=0.23\textwidth]{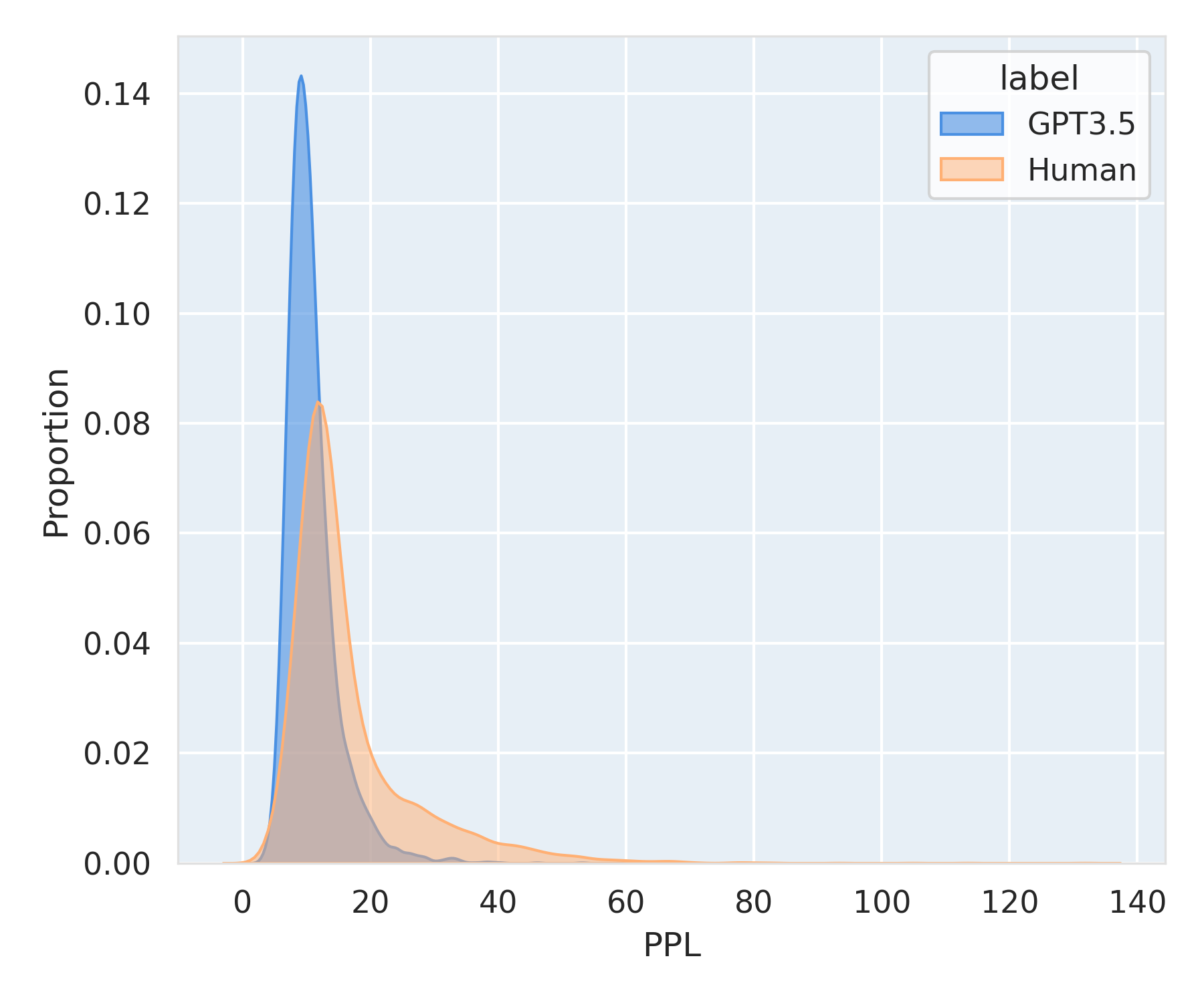}
    }
    \subfigure[Chinese Translate]{
        \includegraphics[width=0.23\textwidth]{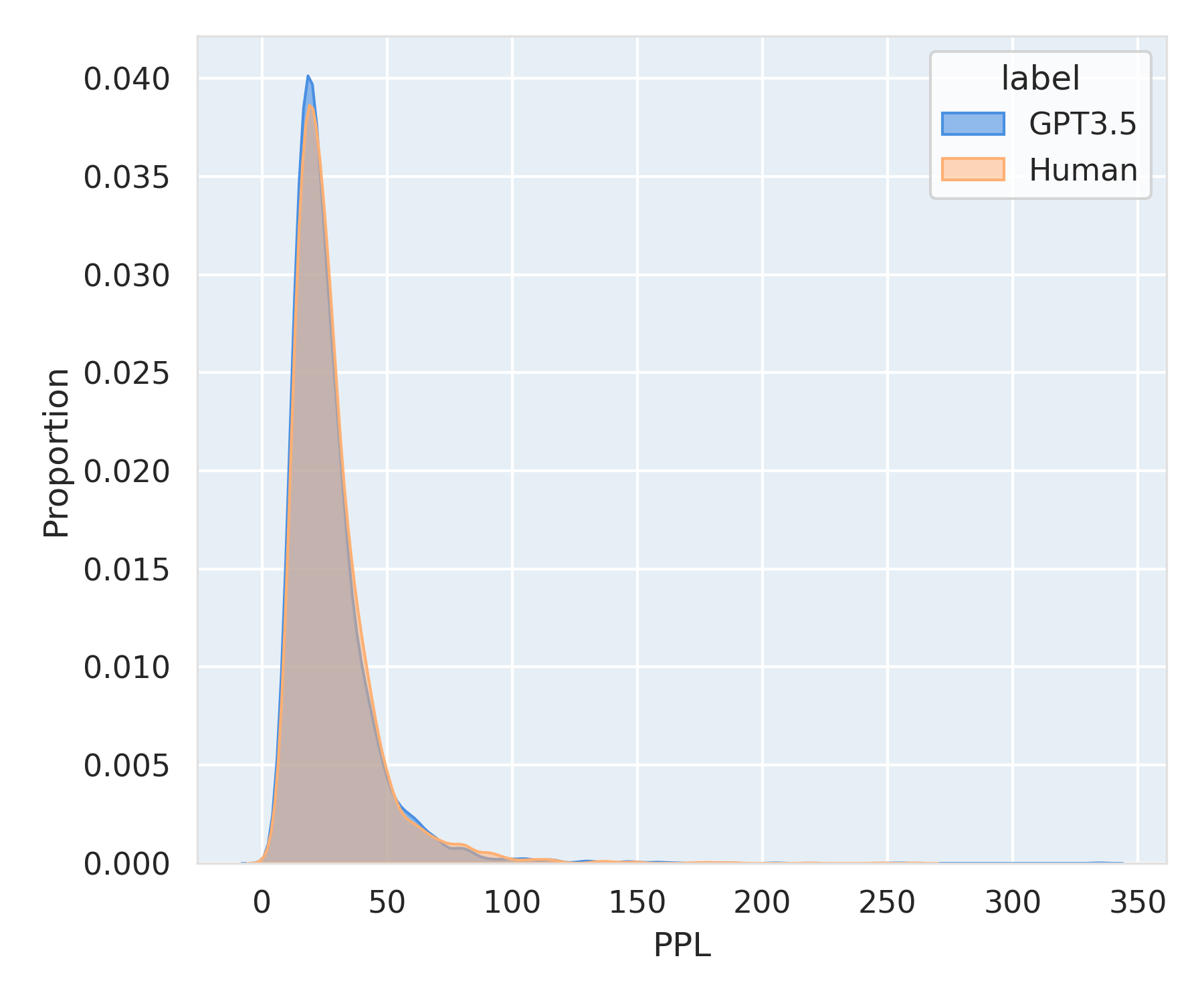}
    }

    \caption{PPL distributions on the proposed Chinese dataset.}
    \label{fig8}
\end{figure*}

To gain a deeper understanding of the variations in text generated by LLMs across multiple operations, we conduct a comparative analysis of the perplexity (PPL) distributions between GPT-3.5-Turbo generated and human-generated texts across various operations. As an effective metric for evaluating textual conformity to common language patterns and structures, PPL allows us to quantify the degree of alignment between LLM-generated and human-generated texts, with lower PPL values indicating closer conformity and higher values suggesting greater deviation \cite{guo2023close}. We use the GPT2-chinese-cluecorpussmall\footnote{https://huggingface.co/uer/gpt2-chinese-cluecorpussmall} model to compute the PPL of Chinese texts generated by humans and GPT-3.5-Turbo. The PPL distributions of various Chinese texts are shown in Figure \ref{fig8}. Similarly, the specific details of the PPL distribution in English are shown in Appendix Figure A1. The PPL values of texts generated by different operations vary considerably, highlighting the distinct characteristics of each task. Except for the ``Translate'' operation, the PPL values of GPT-3.5-Turbo generated texts are generally lower and more concentrated compared to those of human-generated texts. This contrast illustrates the LLMs' relative consistency and confidence across various operations, whereas human-generated texts exhibit greater variability in PPL distributions. Therefore, it is crucial for LLM-generated text detection models to be trained and tested across diverse LLM operation outputs, enabling a more comprehensive assessment of detector performance across various task types.

\section{Experiments}\label{sec4}

\subsection{Experimental Settings}

This section describes our experiments to evaluate the impact of each component in LLM-generated text detection. The system consists of the following components:

\begin{itemize}
\item \textbf{Detection Methods}: The core of the system is the detection model, which analyzes input text to determine if it is generated by LLMs. We evaluated two sets of detection models with different principles.

\item \textbf{Training Data Compositions}: For model-based detection, the training dataset's composition is crucial. This includes data diversity, text generation operations, and LLMs sources. We studied how variations in training data affect model generalization, using a 4:1 training-to-test set ratio.

\item \textbf{LLMs Text Generation Operations}: Different text generation operations involve varying LLM strategies, user prompts, text quality, and semantics, impacting detection results.

\item \textbf{Languages}: Chinese and English differ significantly in structure, vocabulary, syntax, grammar, and cultural background, leading to varying detection outcomes.

\item \textbf{Large Language Models}: The detection system examines texts from various LLMs, whose complexity, coherence, and style can affect detection accuracy.
\end{itemize}

\begin{figure*}[t]
  \centering
  \includegraphics[width=11cm]{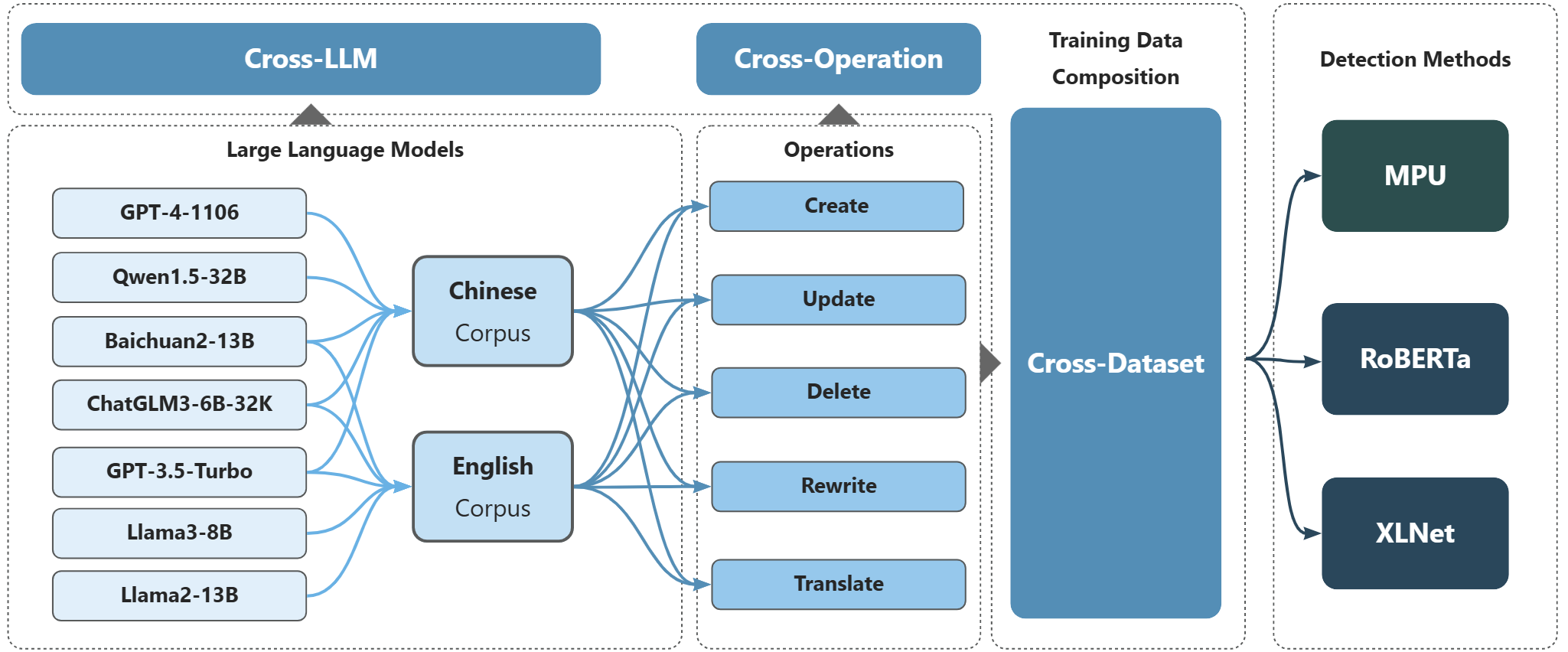}
   \caption{The schematic diagram of the specific experimental process.}
   \label{fig9}
\end{figure*}

To better evaluate the impact of each component in the detection system, we develop an easy-to-extend and use evaluation framework, illustrated in Figure \ref{fig9}. In the experiments, we explore how different component variations affect detection results, focusing primarily on the training data composition. We compare the following values for each component:

\begin{itemize}
\item \textbf{Detection Methods}: MPU, RoBERTa, XLNet.
\item \textbf{Training Data Composition}: Cross-Dataset, Cross-Operation, Cross-LLM.
\item \textbf{LLMs Text Generation Operations}: Create, Update, Delete, Rewrite, Translate.
\item \textbf{Languages}: Chinese, English.
\item \textbf{Large Language Models}: \textbf{Chinese}: Baichuan2-13B, ChatGLM3-6B-32K, GPT-3.5-Turbo, GPT-4-1106, Qwen1.5-32B; \textbf{English}: Baichuan2-13B, ChatGLM3-6B-32K, GPT-3.5-Turbo, Llama3-8B, Llama2-13B.
\end{itemize}

\subsection{Evaluation Metrics}

In this paper, all detectors are binary classification models, determining LLM involvement in text generation. We use common metrics to assess classifier performance: Accuracy, Precision, Recall, and F1-score. Given the true positive (TP), true negative (TN), false positive (FP), and false negative (FN), we calculate the metrics as follows:
\begin{equation}
	Accuracy=\dfrac{TP+TN}{TP+TN+FP+FN},
\end{equation}
\begin{equation}
	Precision=\dfrac{TP}{TP+FP},
\end{equation}
\begin{equation}
	Recall=\dfrac{TP}{TP+FN},
\end{equation}
\begin{equation}
	F1-score=\dfrac{TP}{TP+\dfrac{1}{2}(FP+FN)}.
\end{equation}
These metrics provide a comprehensive overview of the LLM-generated text detector's performance and are commonly used to evaluate binary classification models.

\subsection{Implementation Details}

In our experiments, we employ the Adam optimizer with a learning rate of \(2 \times 10^{-5}\) to optimize model parameters. To improve the training process, we incorporate a StepLR learning rate scheduler with a step size of 2 and a decay factor \(\gamma = 0.8\). The training objective is based on the CrossEntropy loss function. For model-based detection methods, we perform a full update on all model parameters during training. Additionally, in our multi-head attention mechanism, we set the number of attention heads to 8. The models are trained over 30 epochs with a batch size of 64 for the training phase. During optimization, only parameters that require gradients are updated, ensuring computational efficiency. All models are trained and evaluated on two NVIDIA H800 GPUs to leverage parallel computation and enhance training speed.

\subsection{Experimental Results}

To comprehensively evaluate LLM-generated text detection methods, it is crucial to examine specific scenarios where they are employed. This section examines the impact of various training dataset compositions on the performance of detection methods, with a particular focus on model-based approaches, using our proposed dataset. We focus on three primary scenarios: \textbf{Cross-Dataset Detection}, \textbf{Cross-Operation Detection}, and \textbf{Cross-LLM Detection}. These experiments provide a robust framework for understanding the strengths and weaknesses of current detectors, offering insights for optimizing their design and implementation. Our comprehensive evaluation expands both the breadth and depth of assessment. It also provides critical guidance for future research and practical applications in LLM-generated text detection.

\subsubsection{\textbf{Cross-Dataset Detection}}

In this experiment, we select three commonly used and recent LLM-generated text detectors, which were either trained or fine-tuned on the widely adopted HC3 dataset \cite{guo2023close}. To assess their performance on unseen texts, we evaluate the detectors on the proposed dataset without fine-tuning. This approach mirrors real-world scenarios where detectors encounter unfamiliar texts, providing an objective measure of their effectiveness in practical applications. This evaluation also serves to test whether the HC3 dataset sufficiently meets the demands of detecting texts generated by current LLMs.

Table \ref{tab4} shows the detection performance of detectors on various operation generated texts without distinguishing LLMs' texts. For both Chinese and English texts, the MPU method consistently achieves the highest average accuracy, precision, and F1-score across different tasks, underscoring its advanced capabilities as a state-of-the-art metric-based model in detecting LLM-generated text across multiple languages. In contrast, model-based detectors like RoBERTa and XLNet rely on the HC3 training dataset to enhance their performance, which may limit their effectiveness compared to the MPU model. Notably, all three detection models achieved their best results on texts generated from the ``Create'' operation, with MPU and XLNet reaching accuracy levels above 80\%. This performance can be attributed to the composition of the HC3 dataset, which predominantly consists of question-answering data, aligning closely with the design of the MPU model, specifically optimized for detecting such content. In contrast, tasks such as ``Delete'' and ``Translate'' show significantly lower accuracy across all detectors, particularly for RoBERTa, highlighting potential limitations in detecting LLM-generated content in shorter texts.

\begin{table}[t]
  \centering
  \caption{Detection results of detectors on various operation generated texts.}
  \resizebox{12cm}{!}{
    \begin{tabular}{llllllllllll}
    \toprule
    \multirow{2}[2]{*}{\textbf{Detectors}} & \multirow{2}[2]{*}{\textbf{Test Data}} & \multirow{2}[2]{*}{} & \multicolumn{4}{c}{\textbf{Chinese}} & \multirow{2}[2]{*}{} & \multicolumn{4}{c}{\textbf{English}} \\
          &       &       & Accuracy & Precision & Recall & F1-score &       & Accuracy & Precision & Recall & F1-score \\
    \midrule
    \midrule
    \multirow{6}[2]{*}{\textbf{MPU}} & Create & \multirow{6}[2]{*}{} & \cellcolor{lightpink}\textbf{0.8938} & 0.9752  & \cellcolor{lightpink}\textbf{0.8374} & \cellcolor{lightpink}\textbf{0.8779} & \multirow{6}[2]{*}{} & \cellcolor{lightpink}\textbf{0.8330} & \cellcolor{lightpink}\textbf{0.8901} & \cellcolor{lightpink}\textbf{0.7771} & \cellcolor{lightpink}\textbf{0.8257} \\
          & Update &       & 0.7215  & 0.9826  & 0.5472  & 0.6733  &       & 0.6648  & 0.8341  & 0.5146  & 0.6348  \\
          & Delete &       & 0.6859  & 0.9144  & 0.5151  & 0.6546  &       & 0.6520  & 0.7430  & 0.5652  & 0.6299  \\
          & Rewrite &       & 0.7644  & \cellcolor{lightpink}\textbf{0.9920} & 0.6101  & 0.7366  &       & 0.6718  & 0.8500  & 0.5211  & 0.6443  \\
          & Translate &       & 0.6330  & 0.6891  & 0.5675  & 0.6216  &       & 0.6418  & 0.7183  & 0.5554  & 0.6256  \\
          \rowcolor{Ocean}
          & \textbf{All Average} &       & 0.7500  & 0.9282  & 0.6221  & 0.7212  &       & 0.7016  & 0.8128  & 0.5988  & 0.6814  \\
    \midrule
    \midrule
    \multirow{6}[2]{*}{\textbf{RoBERTa}} & Create & \multirow{6}[2]{*}{} & \cellcolor{lightpink}\textbf{0.7618} & \cellcolor{lightpink}\textbf{0.9845} & \cellcolor{lightpink}\textbf{0.6237} & \cellcolor{lightpink}\textbf{0.7172} & \multirow{6}[2]{*}{} & \cellcolor{lightpink}\textbf{0.7234} & \cellcolor{lightpink}\textbf{0.6962} & \cellcolor{lightpink}\textbf{0.6407} & \cellcolor{lightpink}\textbf{0.6547} \\
          & Update &       & 0.5470  & 0.9123  & 0.2769  & 0.4228  &       & 0.5224  & 0.5115  & 0.3115  & 0.3871  \\
          & Delete &       & 0.5538  & 0.9425  & 0.2914  & 0.4443  &       & 0.6102  & 0.5708  & 0.5422  & 0.5505  \\
          & Rewrite &       & 0.5645  & 0.9653  & 0.3027  & 0.4581  &       & 0.5186  & 0.5095  & 0.3071  & 0.3832  \\
          & Translate &       & 0.5229  & 0.7626  & 0.3245  & 0.4529  &       & 0.6485  & 0.6258  & 0.6669  & 0.6457  \\
          \rowcolor{Ocean}
          & \textbf{All Average} &       & 0.6016  & 0.9258  & 0.3764  & 0.5100  &       & 0.6099  & 0.5866  & 0.4954  & 0.5267  \\
    \midrule
    \midrule
    \multirow{6}[2]{*}{\textbf{XLNet}} & Create & \multirow{6}[2]{*}{} & \cellcolor{lightpink}\textbf{0.8377} & \cellcolor{lightpink}\textbf{0.9474} & \cellcolor{lightpink}\textbf{0.7544} & \cellcolor{lightpink}\textbf{0.8226} & \multirow{6}[2]{*}{} & \cellcolor{lightpink}\textbf{0.5848} & \cellcolor{lightpink}\textbf{0.5535} & \cellcolor{lightpink}\textbf{0.4654} & \cellcolor{lightpink}\textbf{0.4894} \\
          & Update &       & 0.6090  & 0.7721  & 0.4298  & 0.5490  &       & 0.5018  & 0.5009  & 0.2578  & 0.3404  \\
          & Delete &       & 0.6088  & 0.8335  & 0.4153  & 0.5520  &       & 0.5502  & 0.5279  & 0.3751  & 0.4309  \\
          & Rewrite &       & 0.6320  & 0.8128  & 0.4609  & 0.5847  &       & 0.4981  & 0.4990  & 0.2546  & 0.3372  \\
          & Translate &       & 0.5776  & 0.6891  & 0.4260  & 0.5256  &       & 0.5220  & 0.5113  & 0.3385  & 0.4073  \\
          \rowcolor{Ocean}
          & \textbf{All Average} &       & 0.6530  & 0.8110  & 0.4973  & 0.6068  &       & 0.5314  & 0.5185  & 0.3383  & 0.4010  \\
    \bottomrule
    \end{tabular}}
  \label{tab4}%
\end{table}%

% Table generated by Excel2LaTeX from sheet 'Cross_Dataset_LLM'
\begin{table}[t]
  \centering
  \caption{Detection results of detectors on various LLM-generated texts.}
  \renewcommand{\arraystretch}{1.2}
  \resizebox{12cm}{!}{
    \begin{tabular}{l|lllll|lllll}
    \toprule
    \multirow{2}[2]{*}{\textbf{Detectors}} & \multirow{2}[2]{*}{\textbf{Test Data}} & \multicolumn{4}{c|}{\textbf{Chinese}} & \multirow{2}[2]{*}{\textbf{Test Data}} & \multicolumn{4}{c}{\textbf{English}} \\
          &       & Accuracy & Precision & Recall & F1-score &       & Accuracy & Precision & Recall & F1-score \\
    \midrule
    \midrule
    \multirow{6}[2]{*}{\textbf{MPU}} & Baichuan2-13B & 0.6534  & 0.9182  & 0.4721  & 0.6003  & Baichuan2-13B & 0.6380  & 0.7633  & 0.5063  & 0.6009  \\
          & ChatGLM3-6B-32K & 0.6675  & 0.9313  & 0.4915  & 0.6155  & ChatGLM3-6B-32K & 0.6897  & 0.8184  & 0.5763  & 0.6696  \\
          & GPT-3.5-Turbo & 0.7683  & 0.9288  & 0.6493  & 0.7480  & GPT-3.5-Turbo & 0.6822  & 0.7849  & 0.5808  & 0.6580  \\
          & GPT-4-1106  & \cellcolor{lightpink}\textbf{0.8534} & \cellcolor{lightpink}\textbf{0.9386} & \cellcolor{lightpink}\textbf{0.7816} & \cellcolor{lightpink}\textbf{0.8478} & Llama3-8B & 0.7488 & 0.8481  & 0.6633  & \cellcolor{lightpink}\textbf{0.7401} \\
          & Qwen1.5-32B  & 0.8073  & 0.9241  & 0.7160  & 0.7946  & Llama2-13B & \cellcolor{lightpink}\textbf{0.7496}  & \cellcolor{lightpink}\textbf{0.8495} & \cellcolor{lightpink}\textbf{0.6673} & 0.7381  \\
          \rowcolor{Ocean}
          & \textbf{All Average} & 0.7500  & 0.9282  & 0.6221  & 0.7212  & \textbf{All Average} & 0.7016  & 0.8128  & 0.5988  & 0.6814  \\
    \midrule
    \midrule
    \multirow{6}[2]{*}{\textbf{RoBERTa}} & Baichuan2-13B & 0.5838  & 0.9185  & 0.3415  & 0.4755  & Baichuan2-13B & 0.6071  & 0.5849  & 0.4911  & 0.5237  \\
          & ChatGLM3-6B-32K & 0.6059  & 0.9339  & 0.3751  & 0.5085  & ChatGLM3-6B-32K & 0.5995  & 0.5807  & 0.4875  & 0.5197  \\
          & GPT-3.5-Turbo & 0.6012  & 0.9225  & 0.3702  & 0.5021  & GPT-3.5-Turbo & 0.6109  & 0.5862  & 0.4941  & 0.5256  \\
          & GPT-4-1106  & \cellcolor{lightpink}\textbf{0.6519} & \cellcolor{lightpink}\textbf{0.9537} & \cellcolor{lightpink}\textbf{0.4459} & \cellcolor{lightpink}\textbf{0.5861} & Llama3-8B & 0.6109  & 0.5884  & 0.4942  & 0.5277  \\
          & Qwen1.5-32B  & 0.5872  & 0.9004  & 0.3492  & 0.4776  & Llama2-13B & \cellcolor{lightpink}\textbf{0.6211} & \cellcolor{lightpink}\textbf{0.5926} & \cellcolor{lightpink}\textbf{0.5099} & \cellcolor{lightpink}\textbf{0.5368} \\
          \rowcolor{Ocean}
          & \textbf{All Average} & 0.6016  & 0.9258  & 0.3764  & 0.5100  & \textbf{All Average} & 0.6099  & 0.5866  & 0.4954  & 0.5267  \\
    \midrule
    \midrule
    \multirow{6}[2]{*}{\textbf{XLNet}} & Baichuan2-13B & 0.6206  & 0.7956  & 0.4460  & 0.5624  & Baichuan2-13B & 0.5378  & 0.5223  & 0.3485  & 0.4087  \\
          & ChatGLM3-6B-32K & 0.6449  & 0.8141  & 0.4811  & 0.5937  & ChatGLM3-6B-32K & 0.5242  & 0.5154  & 0.3418  & 0.4012  \\
          & GPT-3.5-Turbo & 0.6683  & 0.8390  & 0.5131  & 0.6254  & GPT-3.5-Turbo & 0.5398  & 0.5234  & 0.3495  & 0.4092  \\
          & GPT-4-1106  & \cellcolor{lightpink}\textbf{0.7428} & \cellcolor{lightpink}\textbf{0.8844} & \cellcolor{lightpink}\textbf{0.6224} & \cellcolor{lightpink}\textbf{0.7205} & Llama3-8B & 0.5387  & 0.5226  & 0.3479  & 0.4085  \\
          & Qwen1.5-32B  & 0.6489  & 0.7967  & 0.4911  & 0.5964  & Llama2-13B & \cellcolor{lightpink}\textbf{0.5433} & \cellcolor{lightpink}\textbf{0.5255} &\cellcolor{lightpink}\textbf{0.3560} & \cellcolor{lightpink}\textbf{0.4136} \\
          \rowcolor{Ocean}
          & \textbf{All Average} & 0.6530  & 0.8110  & 0.4973  & 0.6068  & \textbf{All Average} & 0.5314  & 0.5185  & 0.3383  & 0.4010  \\
    \bottomrule
    \end{tabular}}
  \label{tab5}%
\end{table}%

Table \ref{tab5} shows the detection performance of detectors on various LLM-generated texts without distinguishing LLM operations' texts. This table presents the average performance of these detectors in detecting text generated by five different LLMs in both Chinese and English. For Chinese texts, all detection methods achieve their highest Accuracy and F1-score on GPT-4-1106-generated texts, with the MPU model performing exceptionally well, reaching an Accuracy of 0.8534. This result is primarily due to the HC3 dataset being largely composed of texts generated by the GPT series models. As a result, the three detectors fine-tuned on this dataset are more familiar with the patterns in GPT-generated content. Furthermore, all three detectors show their lowest performance on texts generated by Baichuan2-13B, suggesting that this LLM-generated content differs substantially from that of the GPT series. For English texts,  all three detectors achieve their highest performance on texts generated by the Llama series. This likely reflects the structured and predictable nature of Llama-generated text, which is more readily captured by the detectors. In contrast, detection performance declines significantly on texts generated by Baichuan2-13B and ChatGLM3-6B-32K, particularly for XLNet, underscoring potential limitations when encountering content with patterns that diverge from the detectors' familiar training contexts. Additionally, it is noteworthy that all three detectors exhibit higher precision than recall, likely due to the imbalance of positive and negative samples in the HC3 training dataset, leading to a disparity between these metrics during testing.

\subsubsection{\textbf{Cross-Operation Detection}}

Since the internal mechanisms of metric-based models like MPU are predefined, we use only RoBERTa and XLNet for cross-operation testing on our proposed dataset. To assess their adaptability, we pre-train the models on texts generated from each of the five main operations individually and then test their detection performance across other operations. This approach leverages limited data from a single operation to assess how effectively detectors can generalize, providing insights into optimizing detector performance under constrained resources.

\begin{figure*}[t]
    \centering
    \subfigure[Chinese Accuracy]{
        \includegraphics[width=0.23\textwidth]{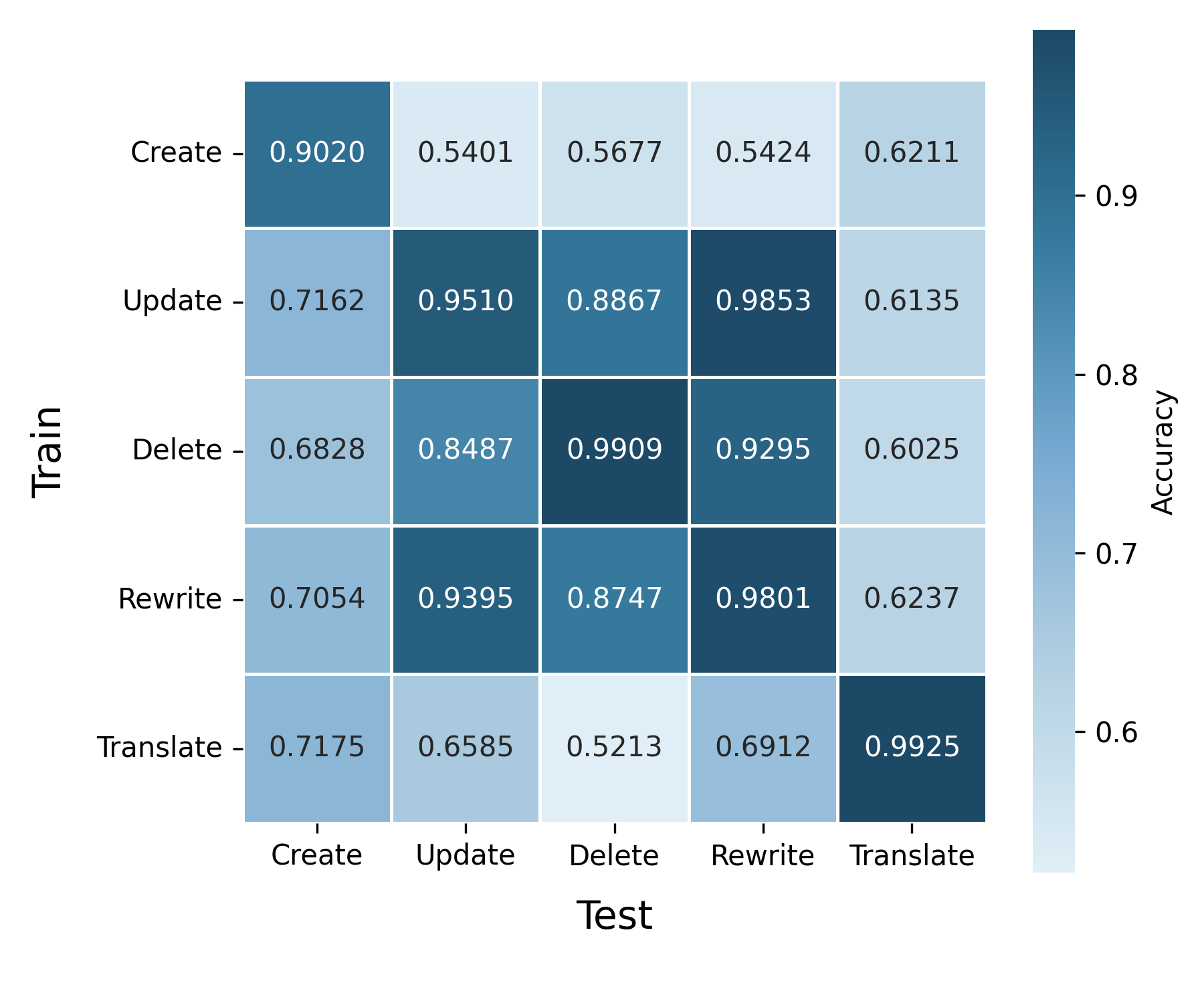}
    }
    \subfigure[Chinese F1-score]{
        \includegraphics[width=0.23\textwidth]{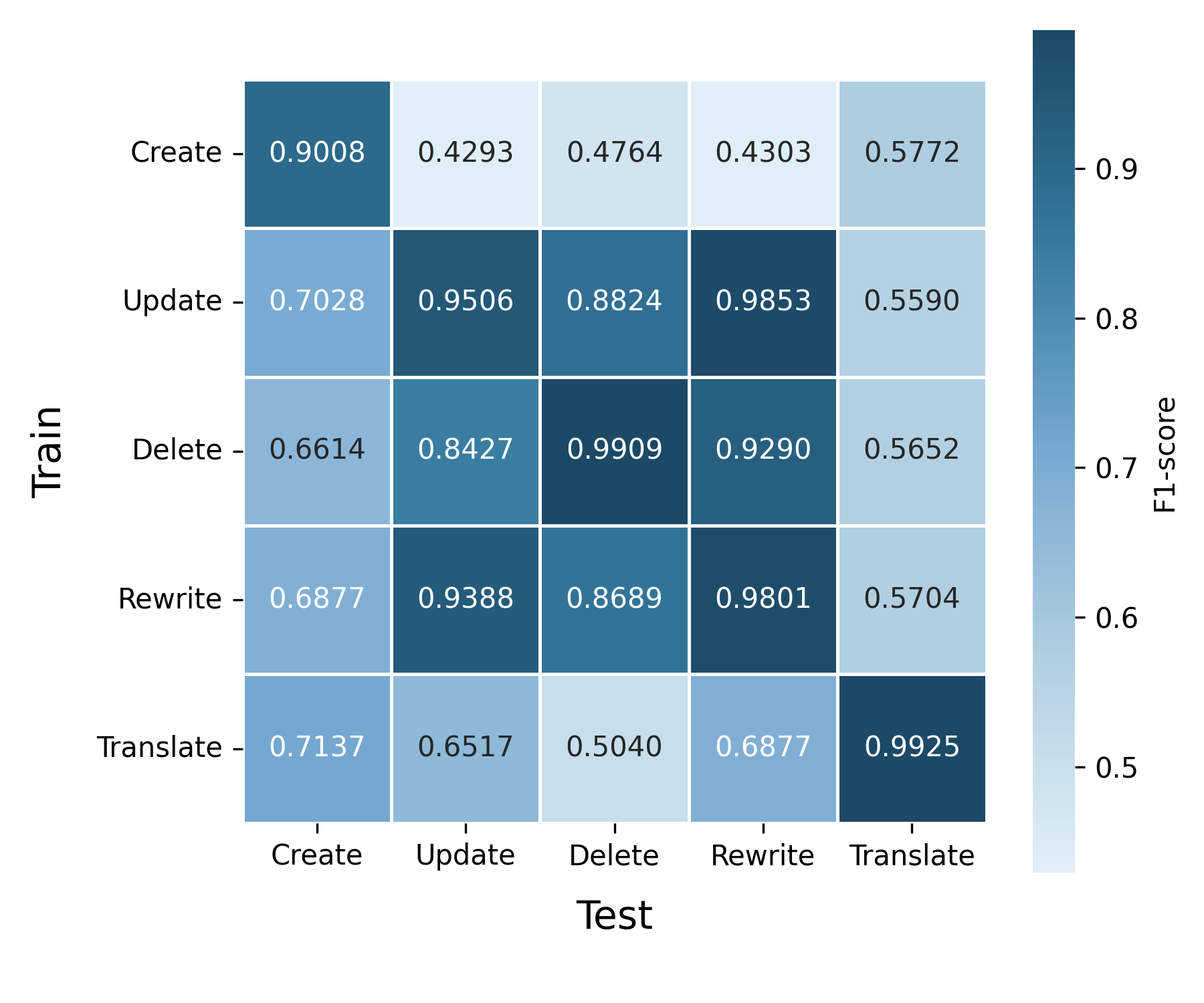}
    }
    \subfigure[English Accuracy]{
        \includegraphics[width=0.23\textwidth]{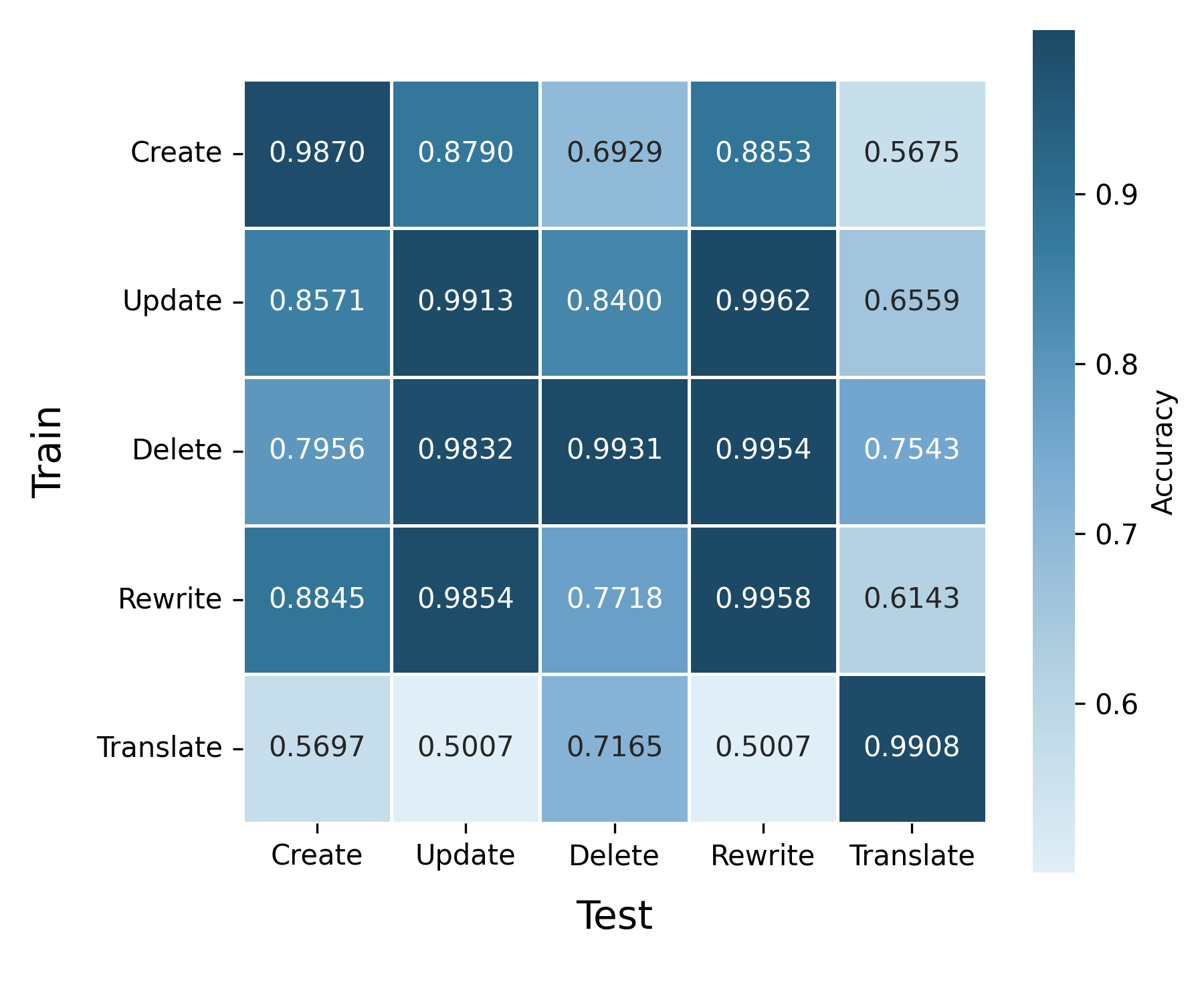}
    }
    \subfigure[English F1-score]{
        \includegraphics[width=0.23\textwidth]{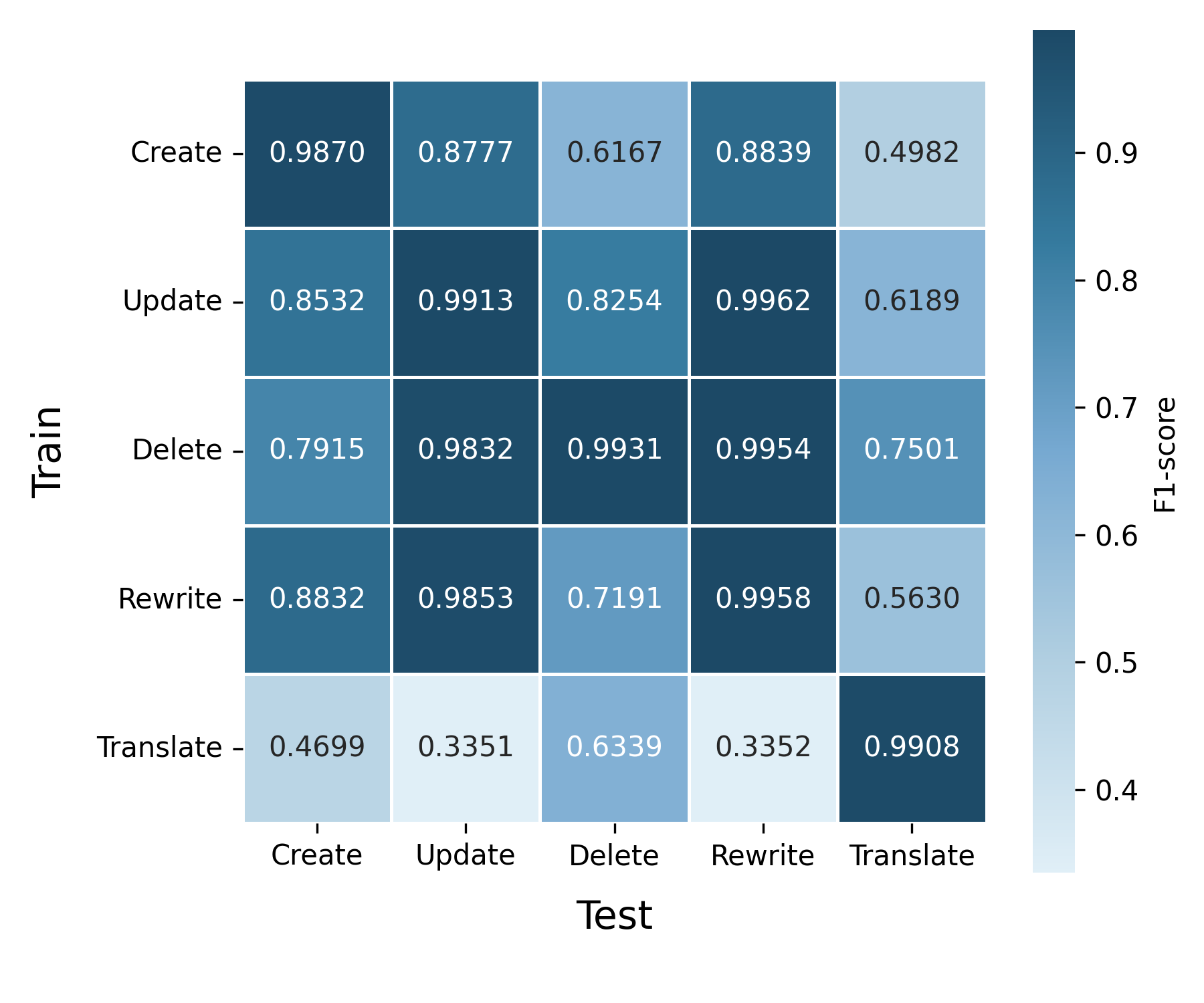}
    }
    \caption{Heatmaps of RoBERTa model trained on single operation data and tested on other operations.}
    \label{fig10}
\end{figure*}

\begin{figure}[t]
    \centering
    \subfigure[Chinese Accuracy]{
        \includegraphics[width=0.23\textwidth]{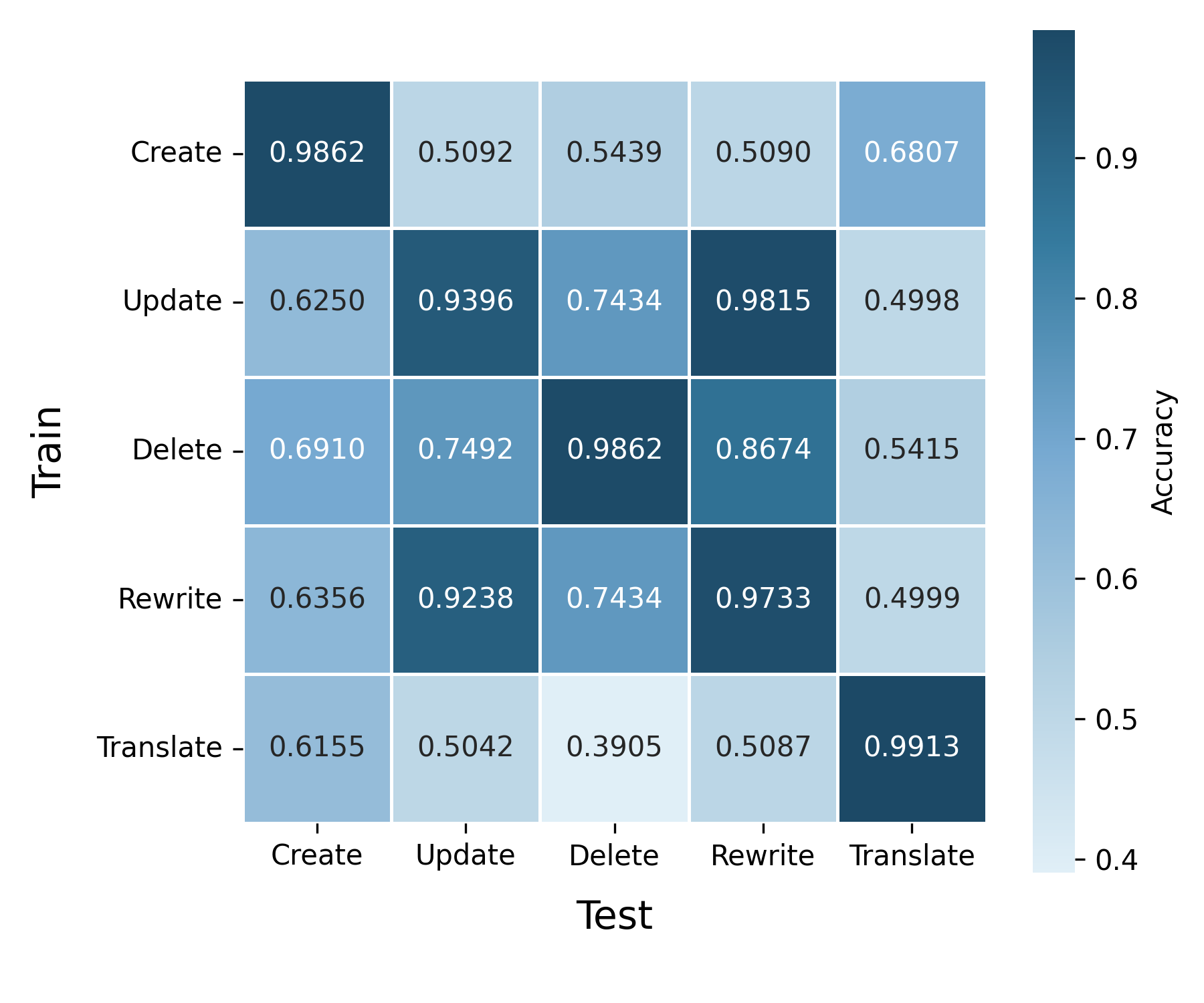}
    }
    \subfigure[Chinese F1-score]{
        \includegraphics[width=0.23\textwidth]{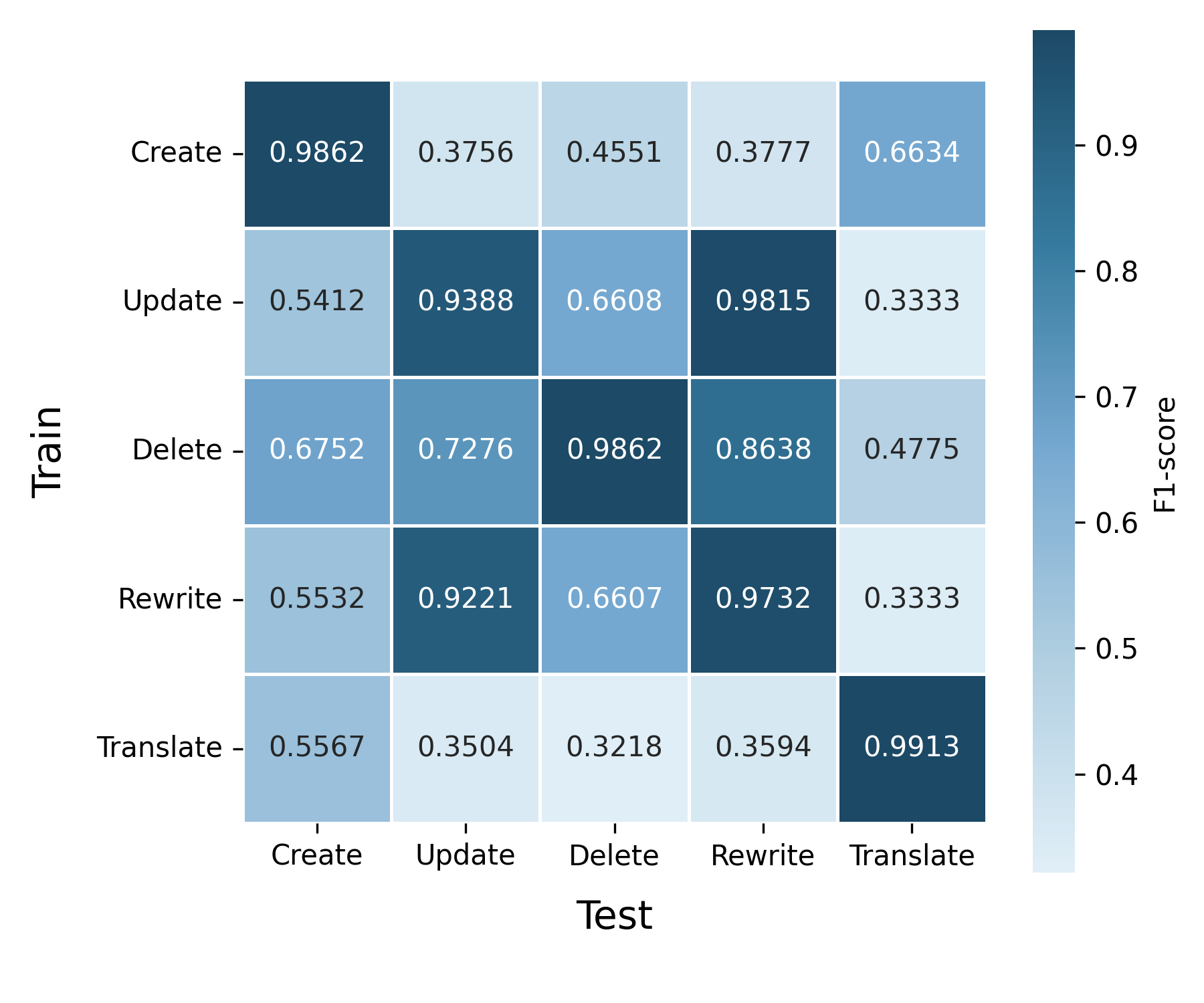}
    }
    \subfigure[English Accuracy]{
        \includegraphics[width=0.23\textwidth]{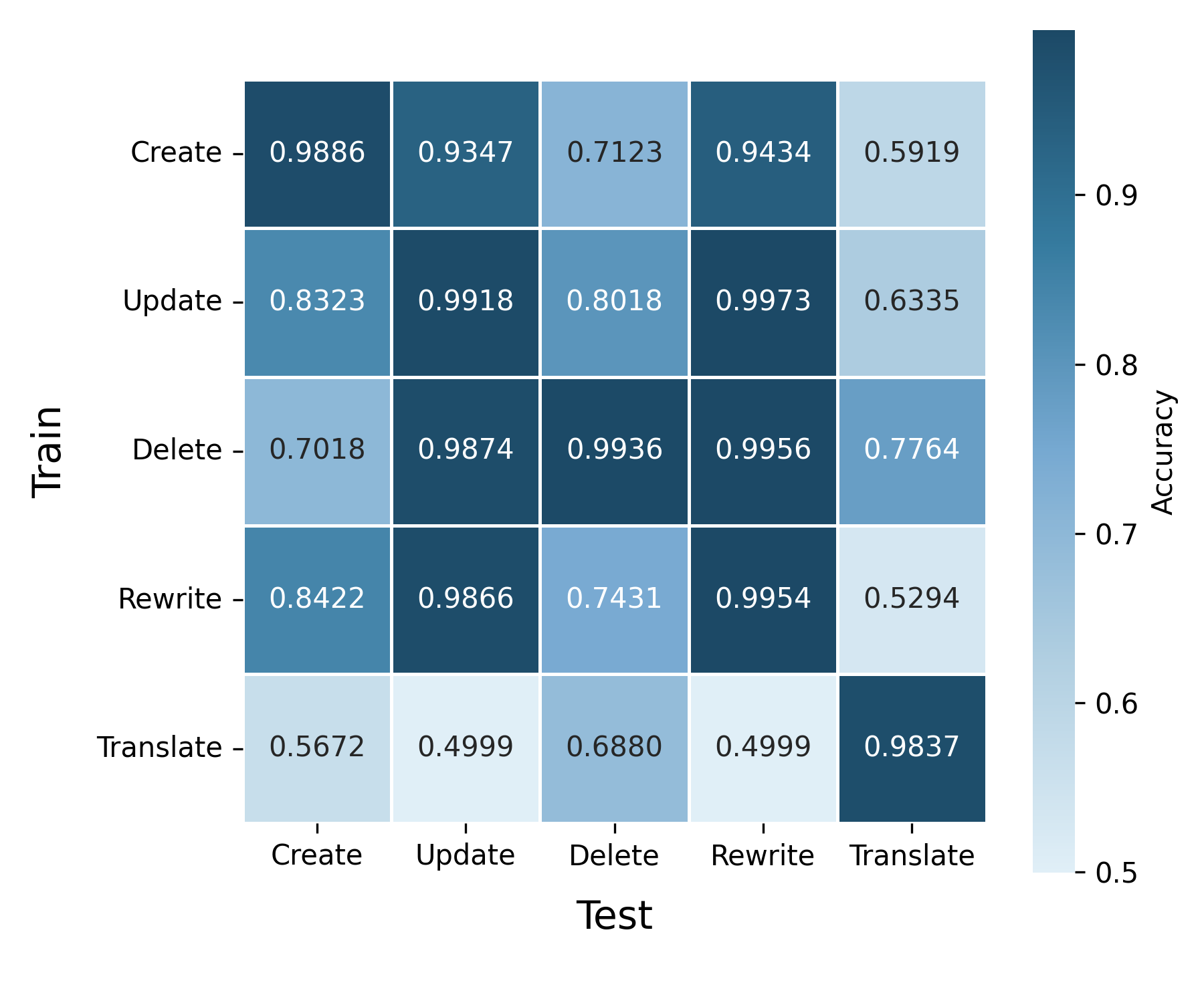}
    }
    \subfigure[English F1-score]{
        \includegraphics[width=0.23\textwidth]{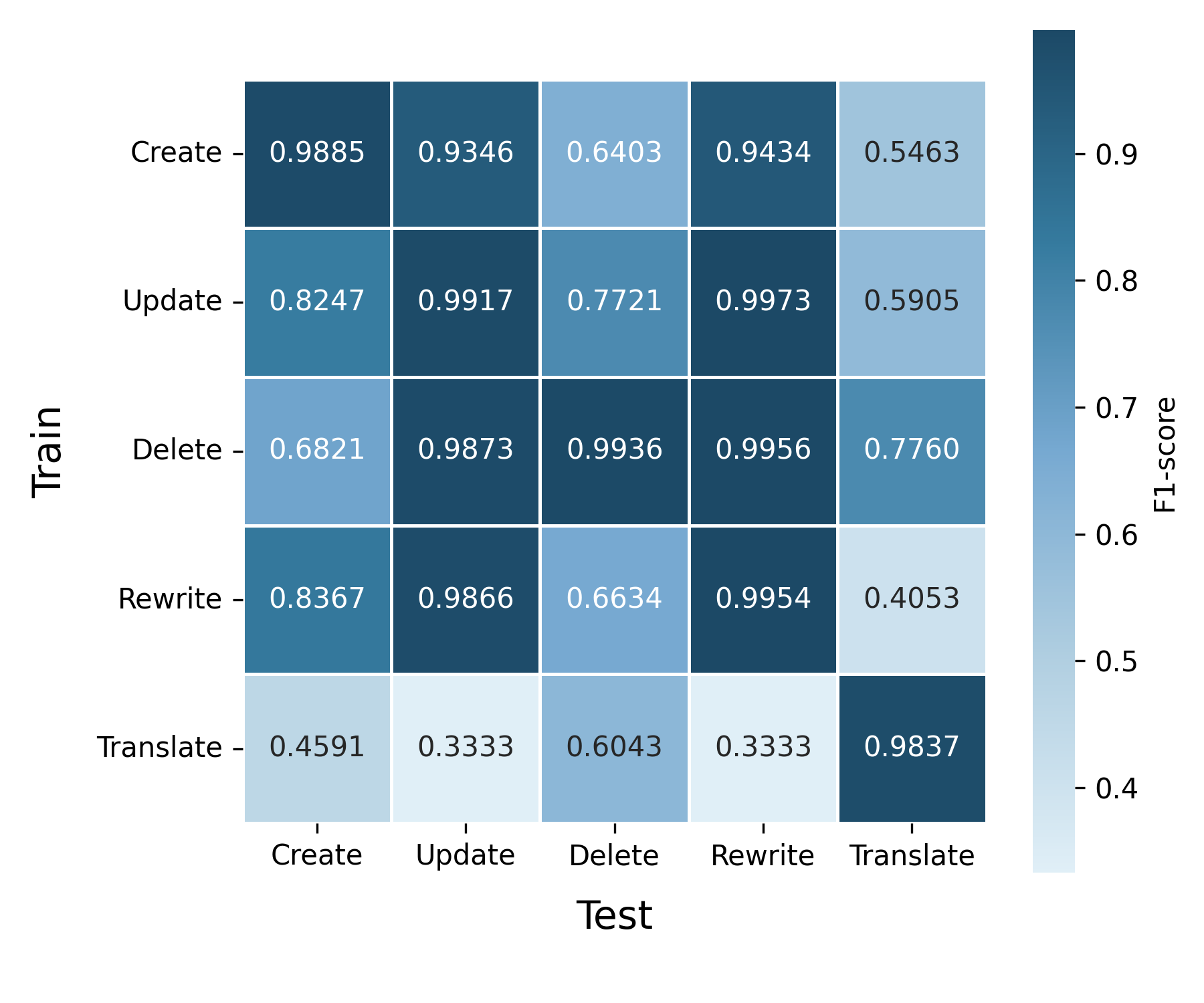}
    }
    \caption{Heatmaps of XLNet model trained on single operation data and tested on other operations.}
    \label{fig11}
\end{figure}

Figure \ref{fig10} illustrates the RoBERTa model's performance across Chinese and English contexts through Accuracy and F1-score, showcasing significant variations when trained and tested on different operations. For Chinese texts, RoBERTa demonstrates poor performance when trained on ``Create'' and ``Translate'' operations. While the model achieves relatively high Accuracy and F1-score within the same operation, its detection Accuracy and F1-score on other operations rarely exceed 0.75. This finding suggests that these two operation types are less suitable for training the model due to RoBERTa's limited generalization capabilities across tasks. This may be attributed to the substantial differences between the texts generated by these two operations and those of other types. In contrast, for the other three types of texts, RoBERTa performs consistently well across different test operations. Except for slightly weaker detection on the ``Translate'' text, the model achieves Accuracy and F1-score above 0.7 in most cases. This indicates that RoBERTa is better able to capture and learn the subtle distinctions between human-written and LLM-generated texts in these three types. For English texts, RoBERTa performs slightly better than on Chinese texts. Except for the ``Translate'' operation, the model achieves satisfactory Accuracy and F1-score when trained on other types of texts. Notably, when trained on ``Delete'' texts, RoBERTa attains Accuracy and F1-score exceeding 0.75 across various test operations, with results surpassing 0.98 on ``Update'', ``Delete'', and ``Rewrite'' tests. This suggests that, in resource-constrained environments, ``Delete'' texts serve as an effective training dataset for the RoBERTa model. Similarly, Figure \ref{fig11} presents heatmaps of XLNet's performance on training and testing across cross-operation texts. For Chinese texts, the overall trend in XLNet's detection results is consistent with RoBERTa, showing poor performance when trained on ``Create'' and ``Translate'' texts, while achieving stronger generalization when trained on ``Update'', ``Delete'', and ``Rewrite'' texts. This further indicates that these three types of Chinese texts are highly suitable for training model-based detectors. The performance on English texts follows a similar pattern, with ``Delete'' being the most effective training text. We speculate that this may be due to LLMs making more substantial and characteristic modifications to human texts in ``Delete'' operations, thereby better highlighting the differences between human and LLM-generated texts.

In addition, we conduct an intriguing set of experiments by training on a single type of operation text and testing across various LLM-generated texts. This approach aims to examine the generalization capability of models trained on limited single-operation data when applied to different LLM-generated texts. The experimental results are presented in Appendix Tables A2 and A3 for RoBERTa and XLNet, respectively. For Chinese texts, RoBERTa, when trained on different operations, consistently performs best on detecting texts generated by GPT-4-1106 and Qwen1.5-32B. This suggests that the texts generated by these models exhibit more noticeable differences from human texts, making them easier to detect. However, the variation in detection performance across different LLMs is smaller compared to the performance differences resulting from training on different operation types. This further supports the validity of our approach in categorizing LLMs outputs by different operations. For English texts, the detection results of RoBERTa across different LLM-generated texts indicate that Llama2-13B yields the best performance overall, although the differences are not particularly significant. For the XLNet model, the overall trend in detection results is largely similar to that of RoBERTa, though the performance is generally inferior. This indicates that RoBERTa demonstrates stronger capabilities when trained on single-operation texts compared to XLNet.

\subsubsection{\textbf{Cross-LLM Detection}}

To account for the distinct characteristics of texts generated by various LLMs, we perform cross-LLM detection experiments. Specifically, we trained our detectors on the full range of operation-specific texts produced by a given LLM and evaluated their detection accuracy and generalization on texts generated by other LLMs. For these experiments, we employ two model-based detectors, RoBERTa and XLNet, to assess their cross-LLM adaptability and gain insights into optimizing detector performance in diverse LLMs contexts.

\begin{figure}[t]
    \centering
    \subfigure[Chinese Accuracy]{
        \includegraphics[width=0.23\textwidth]{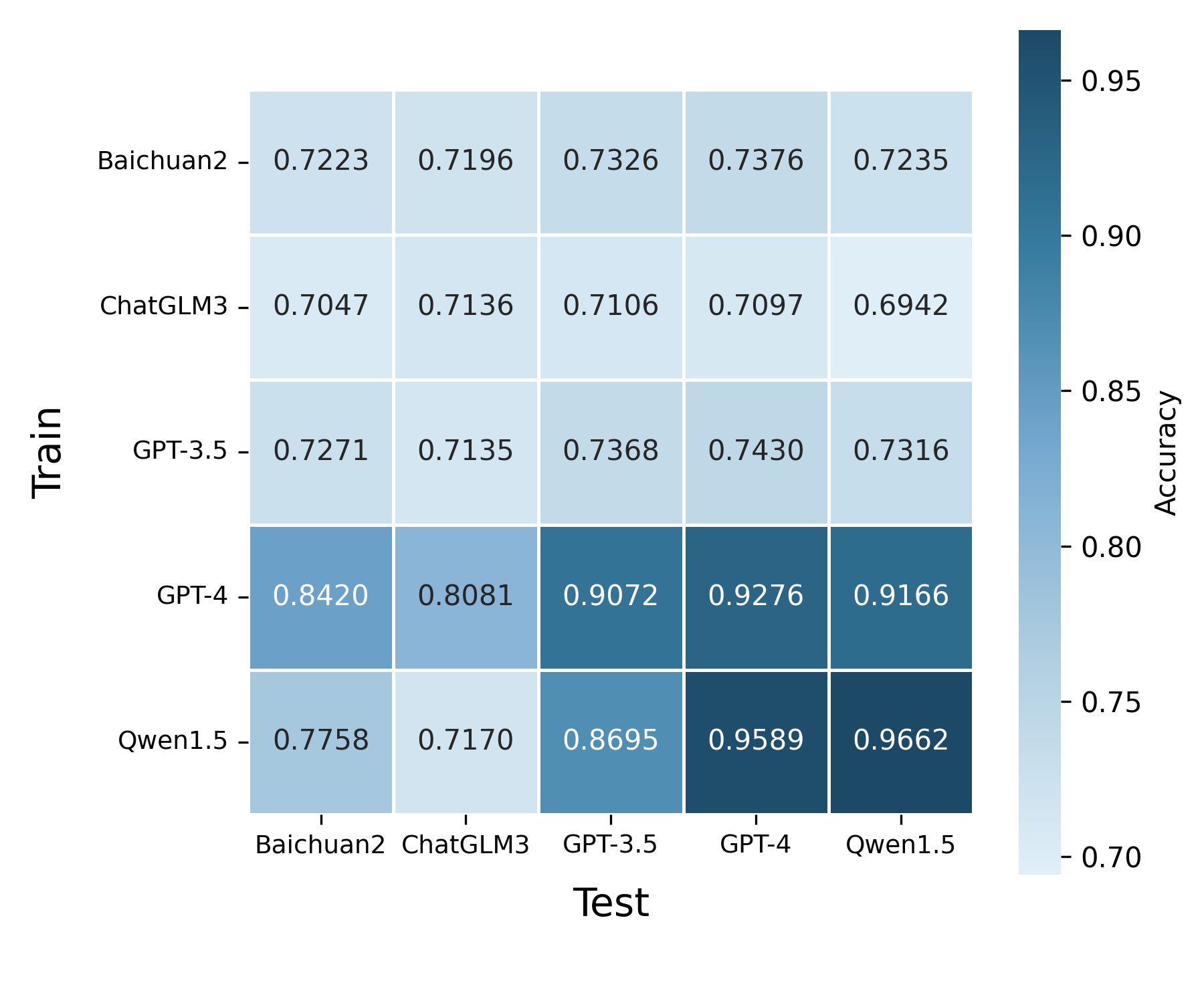}
    }
    \subfigure[Chinese F1-score]{
        \includegraphics[width=0.23\textwidth]{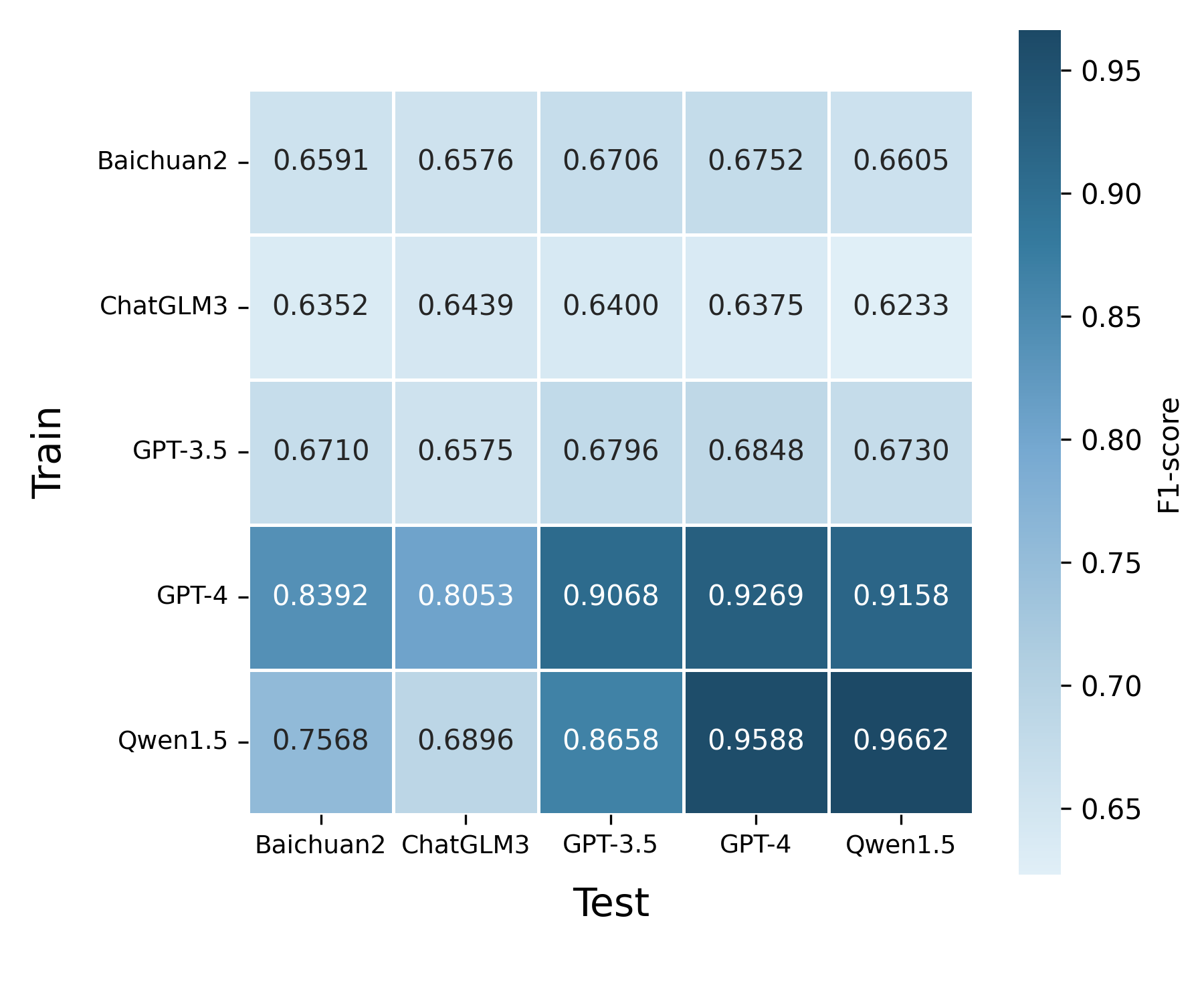}
    }
    \subfigure[English Accuracy]{
        \includegraphics[width=0.23\textwidth]{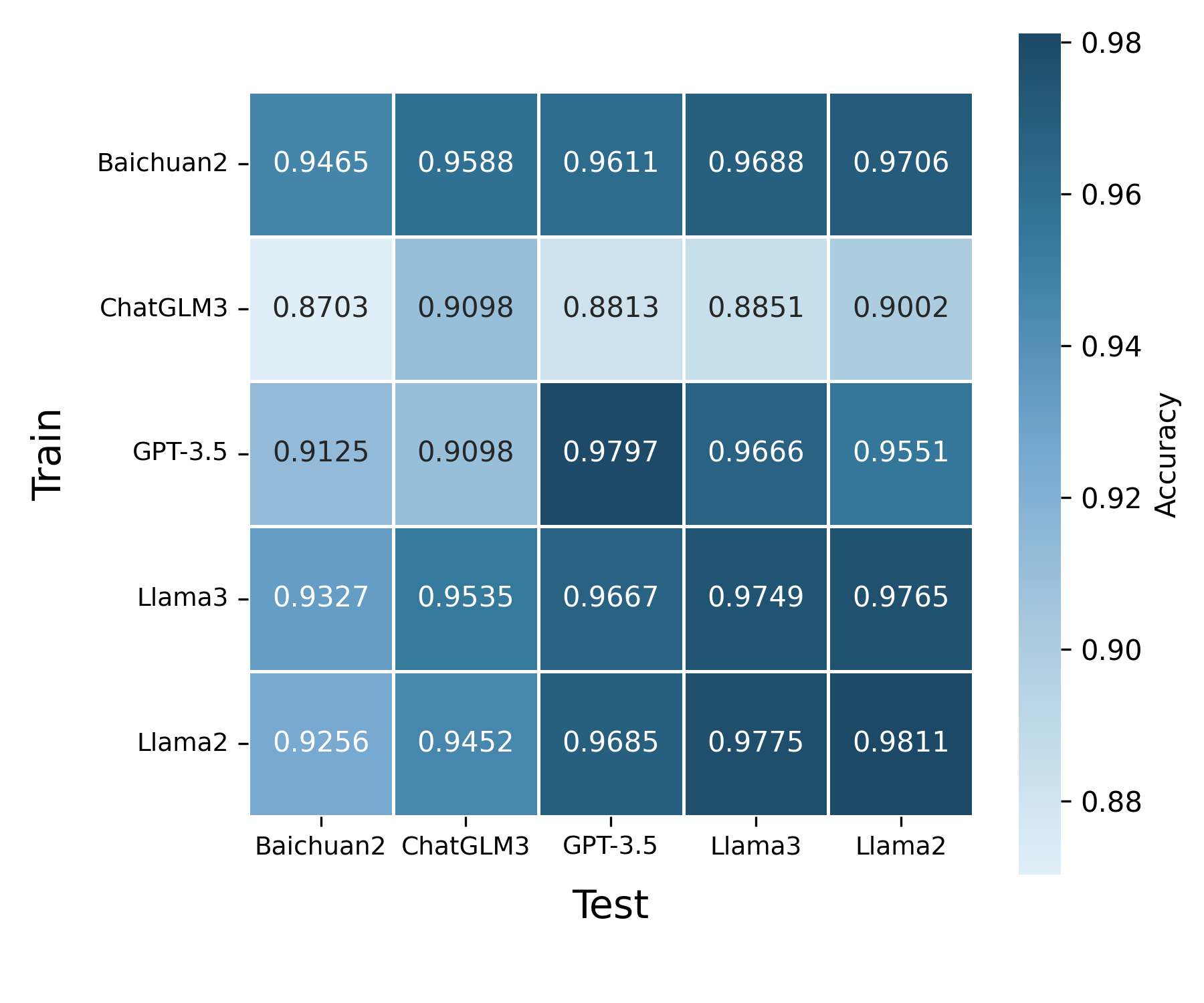}
    }
    \subfigure[English F1-score]{
        \includegraphics[width=0.23\textwidth]{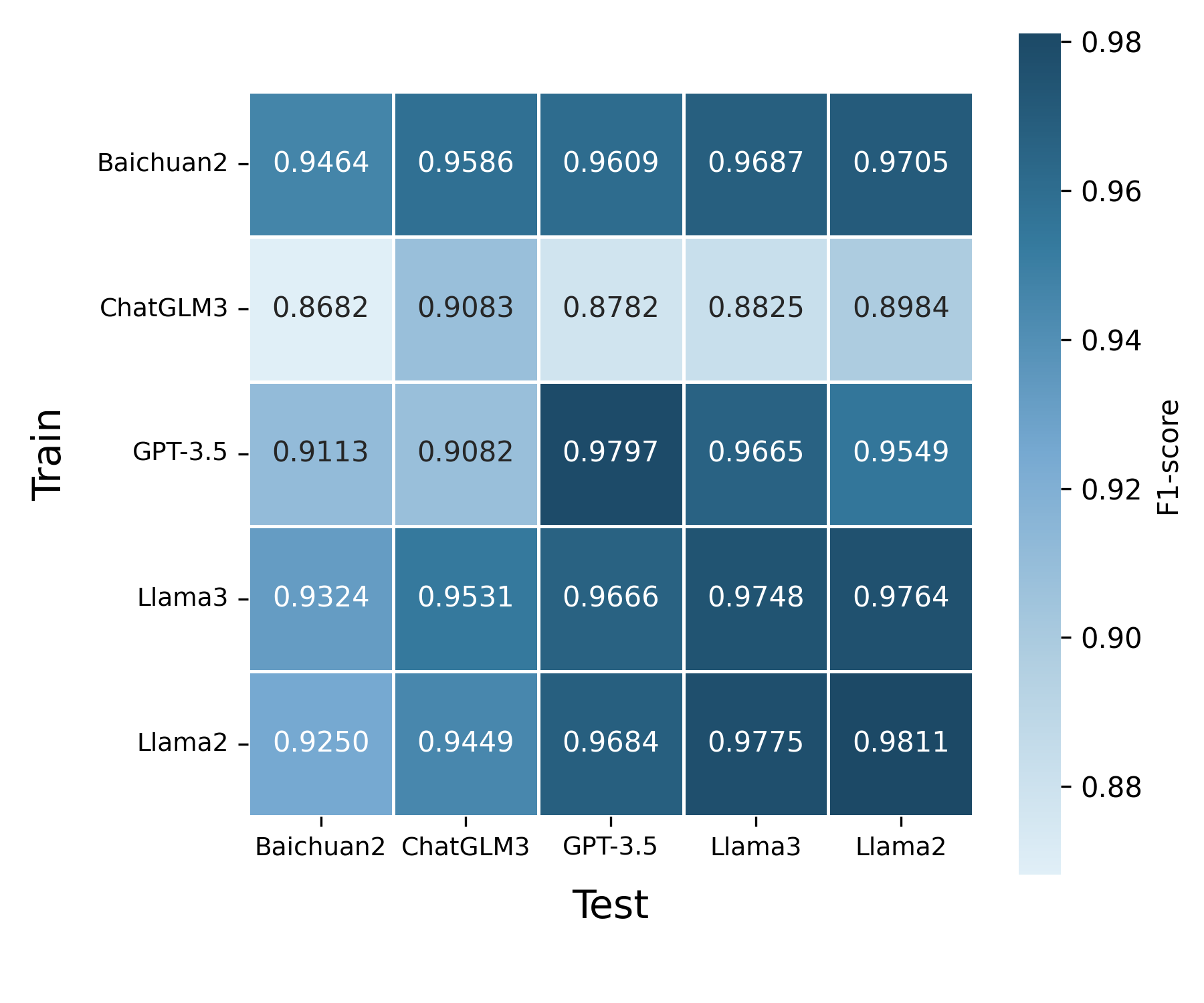}
    }
    \caption{Heatmap results of RoBERTa model trained on single LLM data and tested on other LLMs.}
    \vspace{-\baselineskip}
    \label{fig12}
\end{figure}

\begin{figure}[t]
    \centering
    \subfigure[Chinese Accuracy]{
        \includegraphics[width=0.23\textwidth]{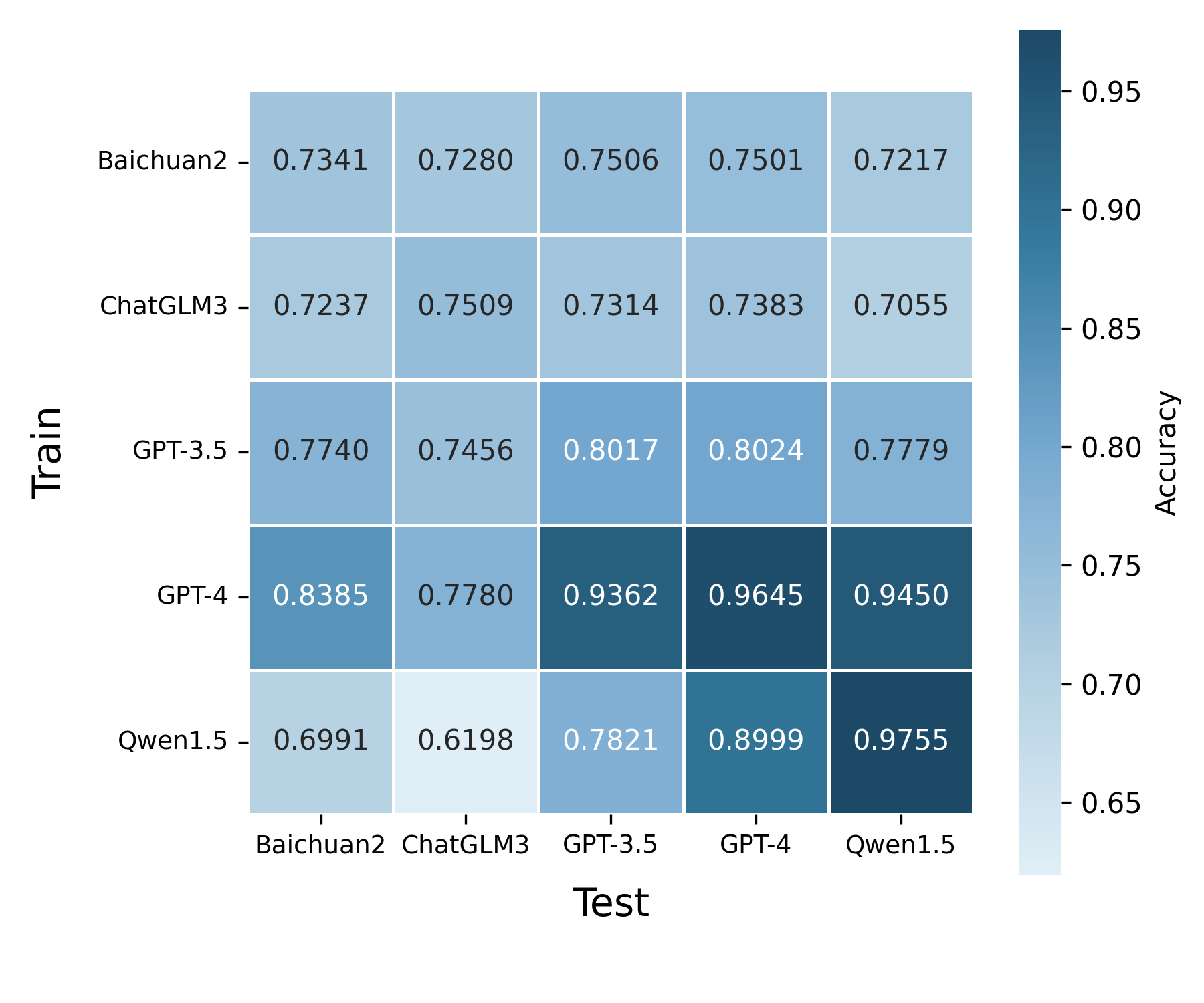}
    }
    \subfigure[Chinese F1-score]{
        \includegraphics[width=0.23\textwidth]{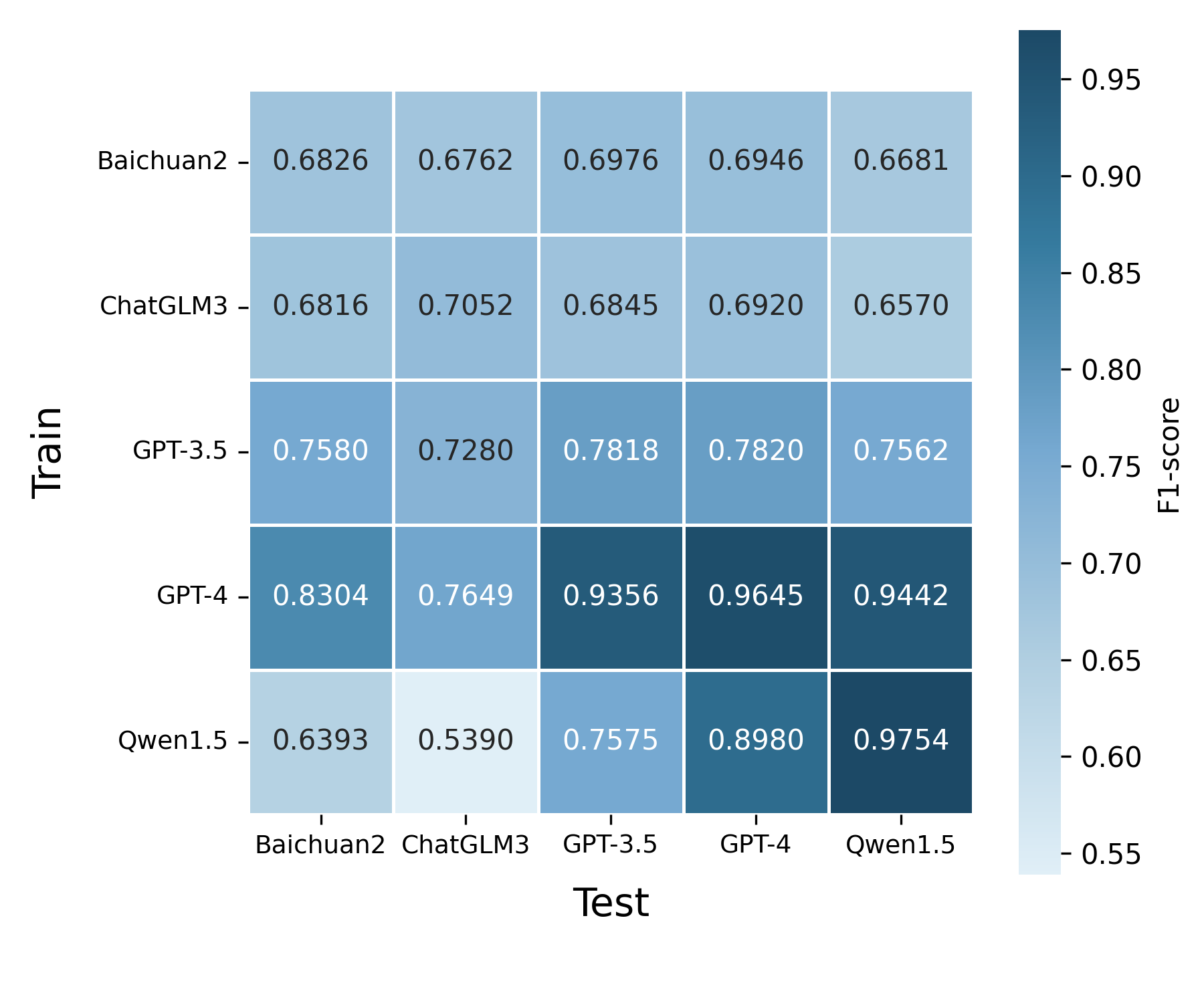}
    }
    \subfigure[English Accuracy]{
        \includegraphics[width=0.23\textwidth]{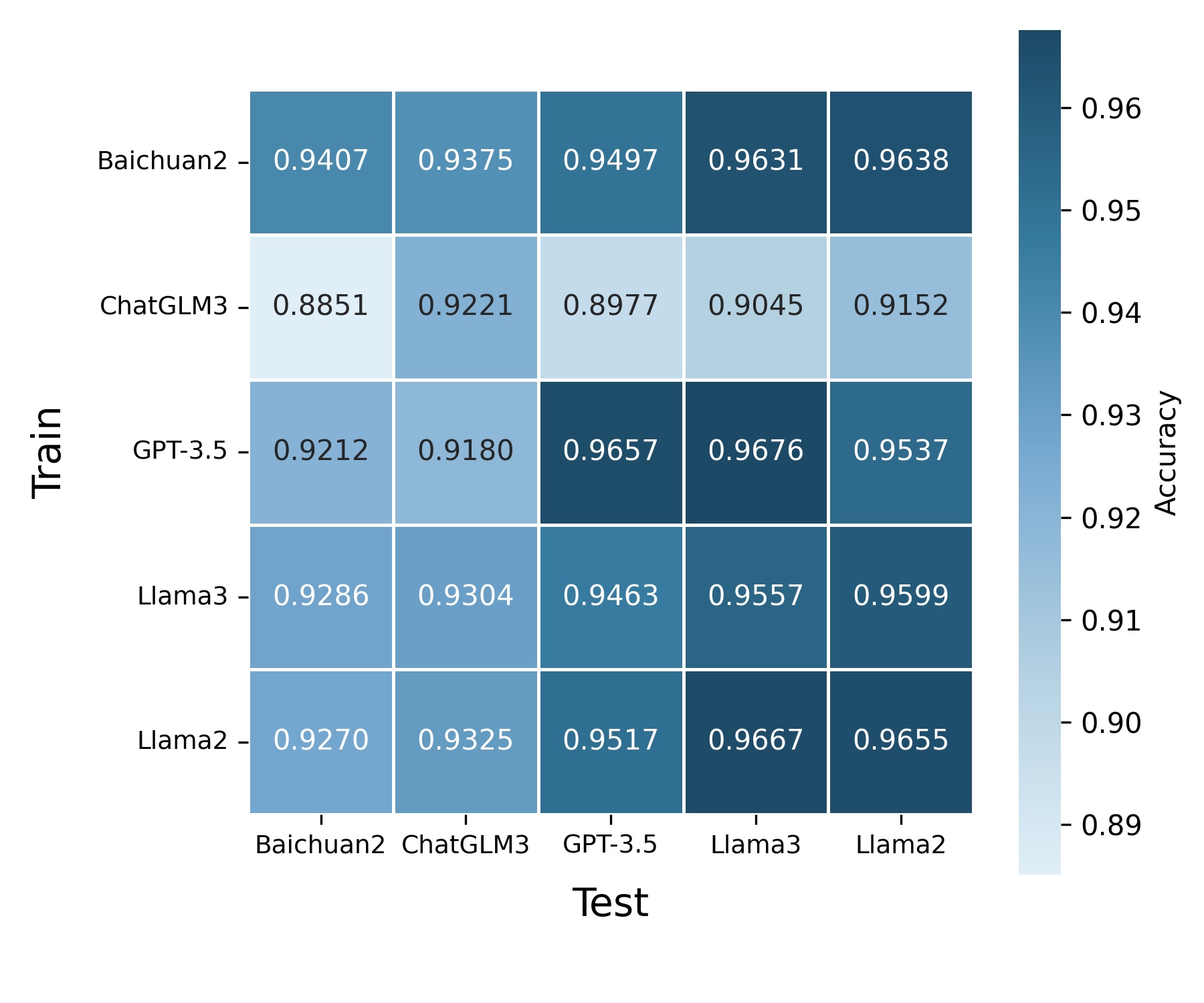}
    }
    \subfigure[English F1-score]{
        \includegraphics[width=0.23\textwidth]{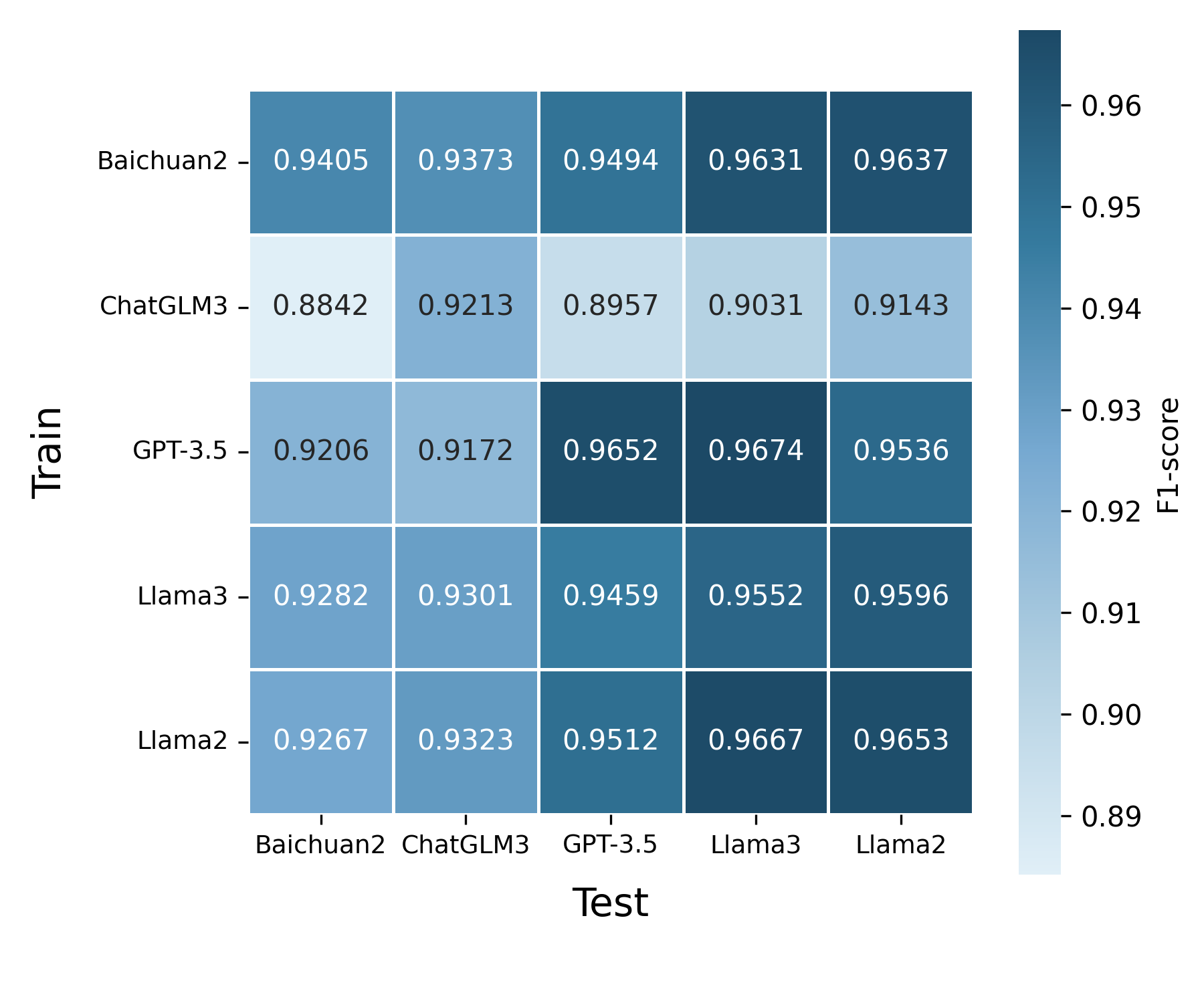}
    }
    \caption{Heatmap results of XLNet model trained on single LLM data and tested on other LLMs.}
    \label{fig13}
\end{figure}

We first investigate the generalization performance of detectors trained on texts from a single LLM across texts generated by other LLMs, without differentiating between operation types. Figure \ref{fig12} presents heatmaps of the RoBERTa model’s performance across Chinese and English contexts, revealing significant variations when trained on texts from one LLM and tested on texts from other LLMs. For Chinese texts, RoBERTa achieves optimal performance when trained on texts generated by GPT-4-1106 and Qwen1.5-32B, with Accuracy and F1-score generally exceeding 0.8 across nearly all other LLM-generated texts. As GPT-4-1106 and Qwen1.5-32B represent the highest level of current Chinese LLMs, training on these texts enables the detector to better capture the intrinsic differences between human and LLM-generated texts. In contrast, when trained on the other three types of texts, RoBERTa’s performance is relatively modest, indicating that these texts do not effectively leverage RoBERTa’s detection capabilities. For English texts, RoBERTa’s performance is notably stronger than with Chinese texts. Except for a slightly weaker result when trained on ChatGLM3-6B-32K texts, RoBERTa achieves high Accuracy and F1-score above 0.9 with training on other text sources. This indirectly suggests that the intrinsic mechanisms of Chinese texts are more challenging to capture compared to English texts. Figure \ref{fig13} presents heatmaps of the XLNet model’s performance across Chinese and English contexts. For Chinese texts, XLNet’s detection results exhibit a distribution largely consistent with RoBERTa’s, though with overall superior performance. This suggests that cross-LLM experiments more effectively harness XLNet’s detection potential. It is noteworthy that XLNet’s generalization performance shows a slight decline when trained on Qwen1.5-32B, with its F1-score on ChatGLM3-6B-32K texts falling to just 0.5390. For English texts, XLNet also demonstrates strong performance, with both Accuracy and F1-score exceeding 0.88. This underscores the need for the rapid development of detection models specifically tailored to Chinese LLM-generated texts.

% Table generated by Excel2LaTeX from sheet 'Cross_LLM_opera_roberta'

Furthermore, we conduct an intriguing experiment by training on texts generated by a single LLM and testing across texts generated by different operations, without distinguishing between LLMs. The experimental results are presented in Appendix Tables A4 and A5 for RoBERTa and XLNet, respectively. The similar distribution of results across both detectors suggests that differences in training data affect each detector in comparable ways. For Chinese texts, both detection models perform best on ``Translate'' operation texts, with Accuracy and F1-score exceeding 0.8. For Chinese texts, both detection models perform best on Translate operation texts, achieving Accuracy and F1-score above 0.8. This indicates that the structural distribution of Translate texts across different LLMs is consistent. In contrast, the detectors perform poorly on ``Update'', ``Delete'', and ``Rewrite'' texts, suggesting significant differences in the operation mechanisms of these three types across different LLMs. Both detectors demonstrate strong performance on English texts, with Accuracy and F1-score exceeding 0.9. This suggests that the underlying mechanisms for different text operations in English LLMs are relatively similar.

\begin{figure}[b]
    \centering
    \subfigure[Cross-Operation Chinese]{
        \includegraphics[width=0.48\textwidth]{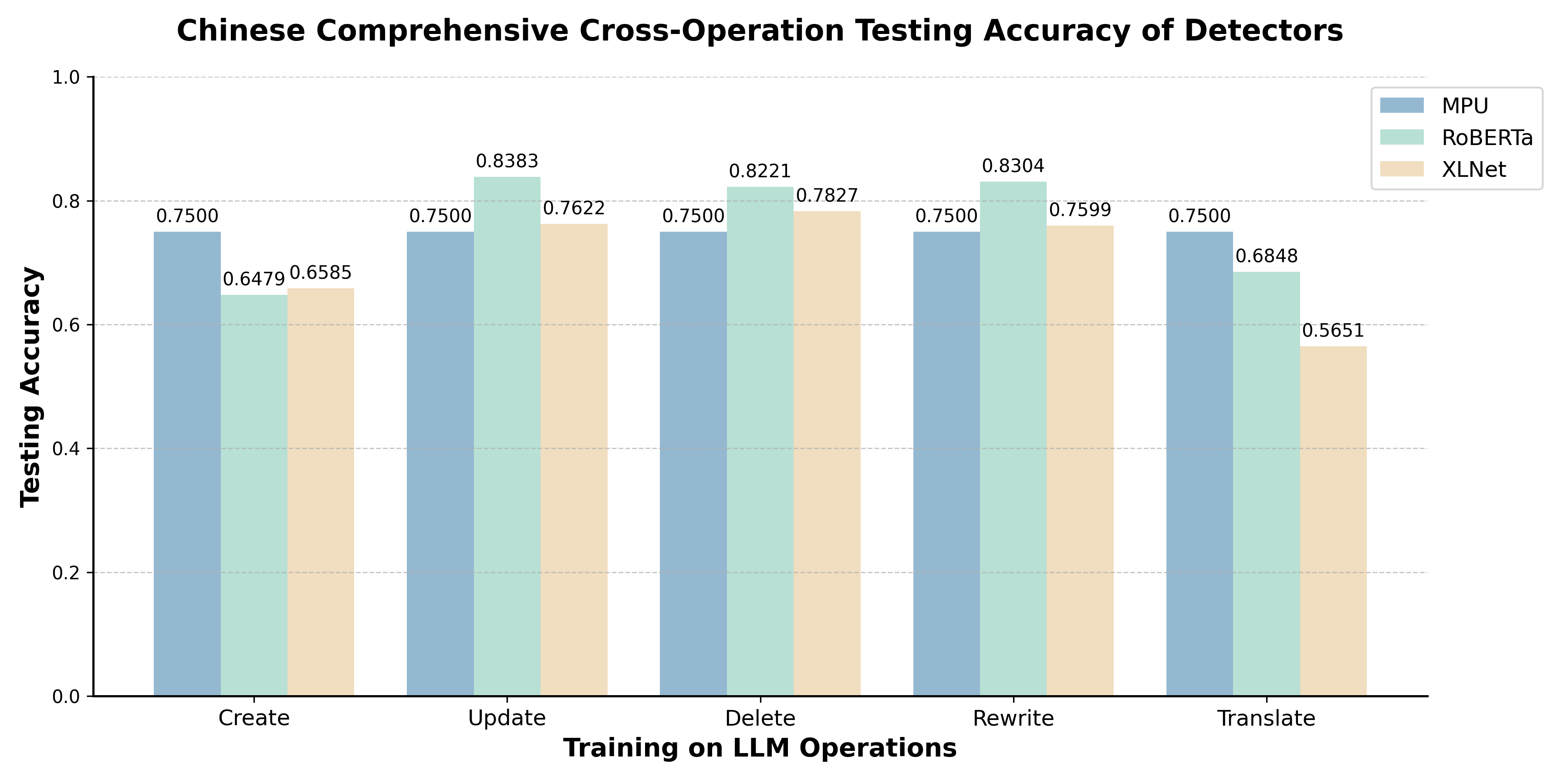}
    }
    \subfigure[Cross-Operation English]{
        \includegraphics[width=0.48\textwidth]{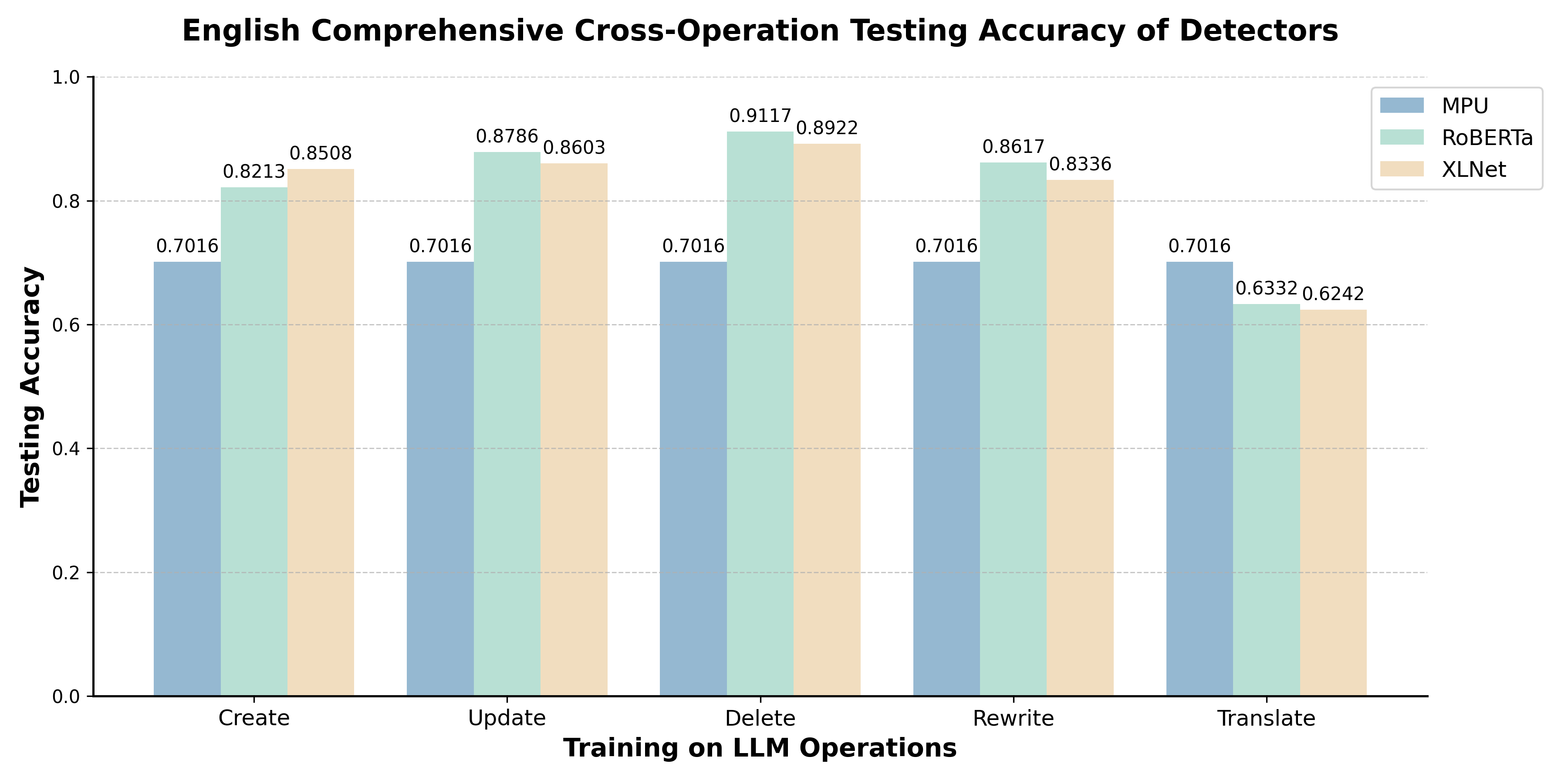}
    }
    
    \subfigure[Cross-LLM Chinese]{
        \includegraphics[width=0.48\textwidth]{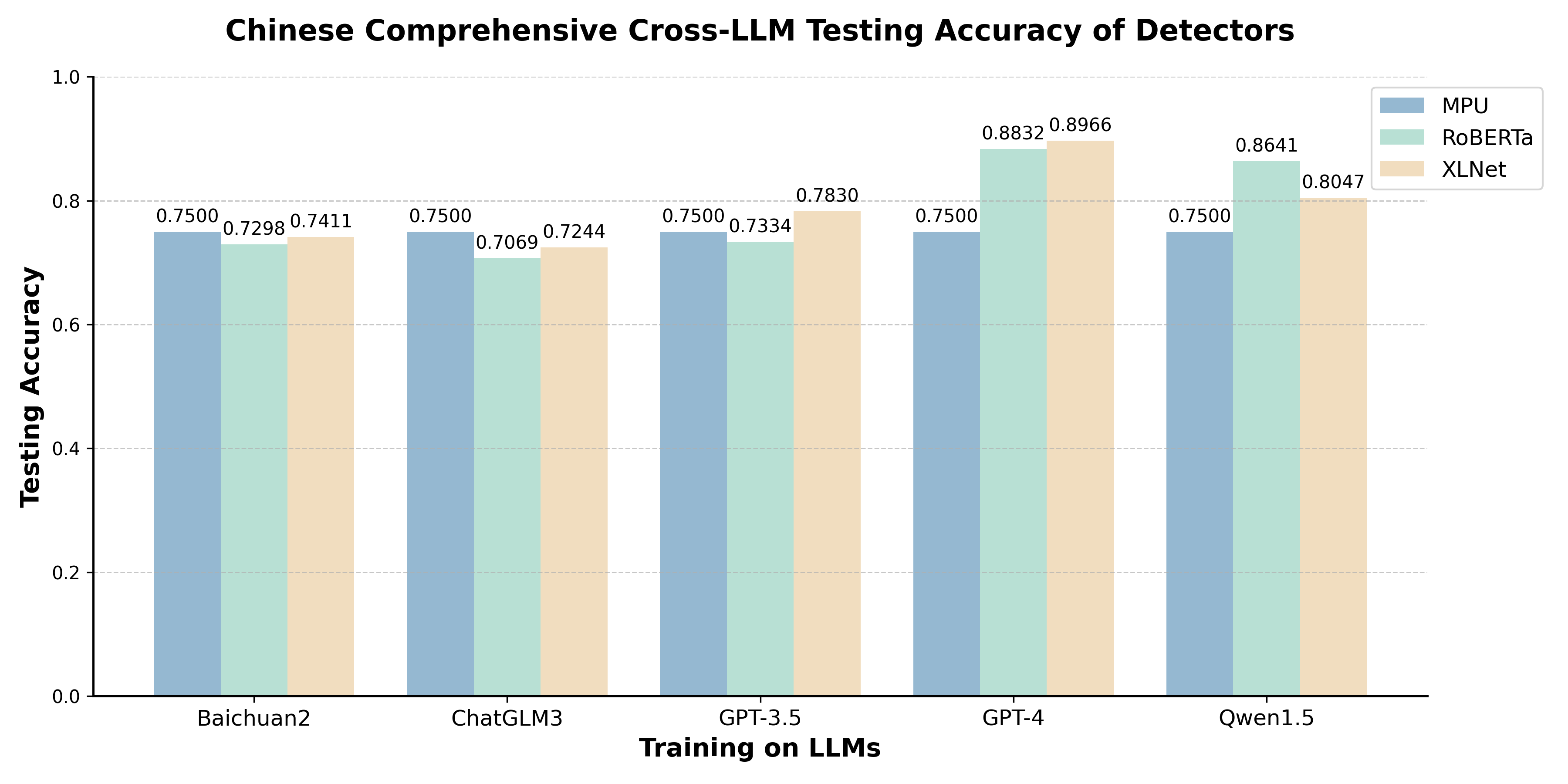}
    }
    \subfigure[Cross-LLM English]{
        \includegraphics[width=0.48\textwidth]{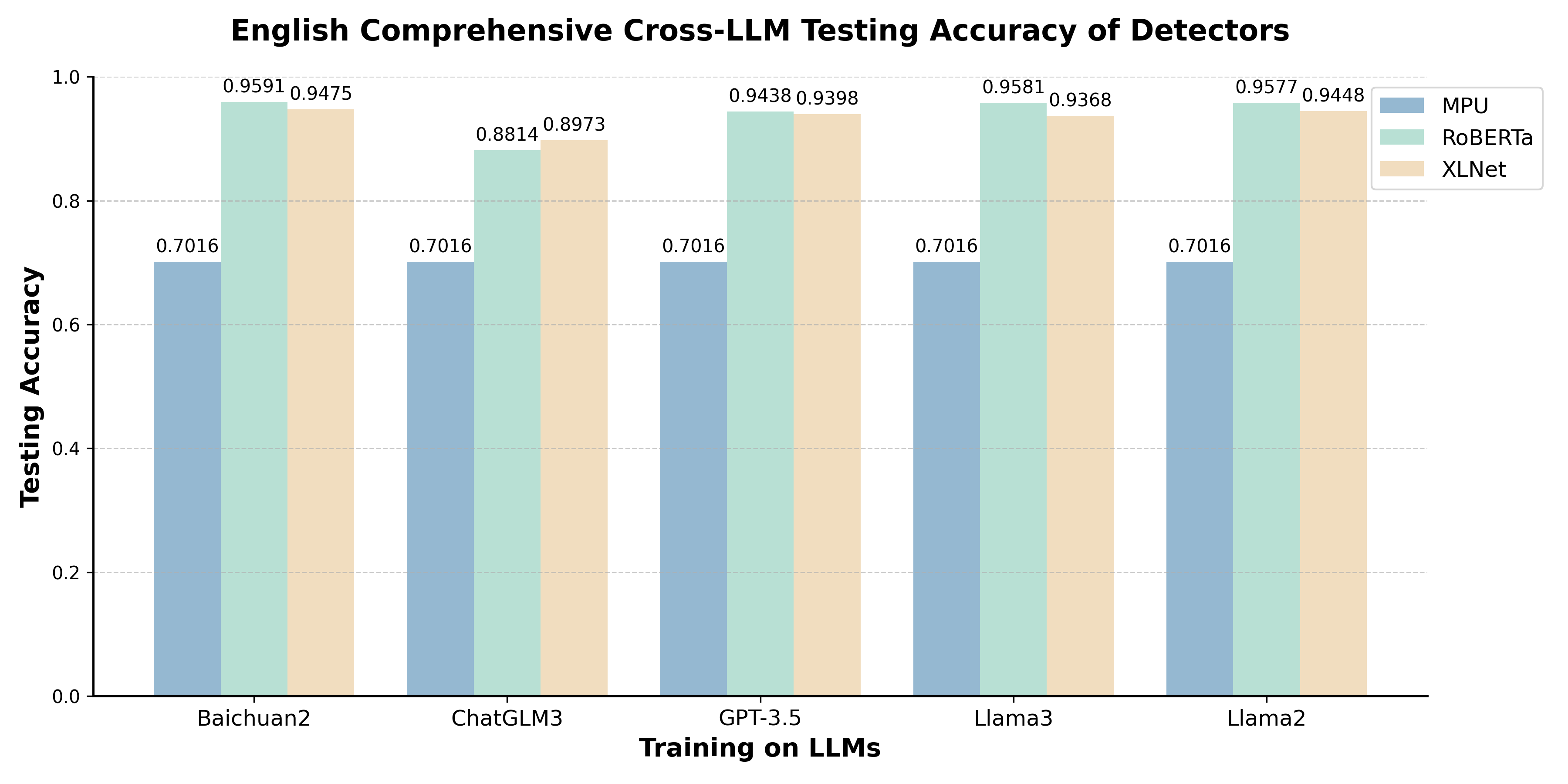}
    }
    \caption{Cross-Operation and Cross-LLM testing Accuracy of detectors across different training configurations.}
    \label{fig14}
    \vspace{-\baselineskip}
\end{figure}

\subsection{Discussions and Suggestions}

Our comprehensive evaluation of LLM-generated text detection systems highlights several key findings for the advancement of this field. The experiments focus on \textbf{Cross-Dataset}, \textbf{Cross-Operation}, and \textbf{Cross-LLM} detection, providing insights into the strengths and limitations of various models and configurations. In the \textbf{Cross-Dataset} experiments, MPU outperforms RoBERTa and XLNet in Accuracy and F1-score, particularly on ``Create'' operations, likely due to its alignment with the HC3 dataset's structure. However, all detection models display limitations in handling ``Translate'' and ``Delete'' operations, reflecting the challenge of detecting LLM-generated content in shorter or structurally divergent texts. The \textbf{Cross-Operation} experiments reveal that training on ``Delete'' texts leads to more generalized performance across tasks, with both RoBERTa and XLNet performing consistently well on various operations. This result suggests that ``Delete'' operations may highlight distinctive LLM-generated text features, such as structural changes and content reduction, which enhance model generalization across other operation types. The findings indicate that certain operation types may capture fundamental LLM characteristics more effectively, providing robust detection insights that aid in broader cross-operation adaptability and model resilience, even in resource-limited settings. Lastly, in the \textbf{Cross-LLM} experiments, models trained on widely used LLMs like GPT-4-1106 and Qwen1.5-32B excel in detecting texts generated by other LLMs, suggesting that these models’ widespread adoption and consistent structural patterns make their outputs easier to recognize. However, detection accuracy drops noticeably for texts from less common models, such as Baichuan2-13B, pointing to challenges in identifying texts with unique or less familiar stylistic patterns. These results emphasize the need for diverse training datasets to enhance the adaptability of detection models across a broader range of LLM-generated content. In addition, we summary the results of the three detectors in the Cross-Operation and Cross-LLM experiments, with specific details shown in Figure \ref{fig14}. Since the MPU model does not require retraining, its test results remain constant throughout. Our findings indicate that model-based detectors outperform metric-based detectors after being trained on ``Update'', ``Delete'', and ``Rewrite'' texts, suggesting that these three operation types are particularly effective for training purposes in both Chinese and English contexts. Detectors trained on GPT-4-1106 and Qwen1.5-32B Chinese texts also exhibit performance improvements, though the overall differences between detectors trained on various LLM texts are relatively minor. These results underscore the substantial differences in internal mechanisms across different LLM operations, which merit attention, and demonstrate the validity of the experimental design in capturing such nuances.

To enhance the robustness and accuracy of LLM-generated text detection systems, several key strategies should be pursued. First, focusing on operation-specific training, particularly with operation types such as ``Delete'', ``Update'', and ``Rewrite'', has proven effective in improving detection performance across different operations. Therefore, emphasizing targeted operation-based training can significantly enhance detector performance in real-world scenarios. Secondly, expanding the diversity of training datasets is crucial, particularly by incorporating more data from advanced LLMs and a broader range of operations, such as GPT-4-1106 and ``Delete''. This would improve generalization capabilities and help address the challenges posed by unfamiliar or divergent text structures. However, future work should also prioritize improving adaptability to more unique or irregular outputs, like those from Baichuan2-13B. Lastly, continued evaluation across multilingual contexts is necessary to further validate the efficacy of these strategies, as language differences can considerably influence detection outcomes. These efforts will collectively ensure that detection models remain effective and adaptable, even as LLM technologies evolve.

\vspace{-\baselineskip}
\section{Conclusion and Future Work}\label{sec5}

In this study, we propose a comprehensive evaluation framework for LLM-generated text detection, \textbf{CUDRT}, that spans diverse languages, text operations, and datasets. Through extensive experiments involving Cross-Dataset, Cross-Operation, and Cross-LLM detection, we evaluate both metric-based and model-based detection methods across multiple scenarios. Our findings reveal that specific text operations, particularly ``Delete'', ``Update'', and ``Rewrite'', are instrumental in enhancing detection model generalization, highlighting their importance as training data in resource-limited environments. Additionally, the results underscore the need for diversified training data, as detectors demonstrated substantial variance in performance across different LLMs and languages, particularly in distinguishing unique stylistic patterns. This work offers valuable insights for enhancing the accuracy and robustness of LLM-generated text detectors while establishing a scalable framework for real-world, multilingual applications. By addressing diverse operational scenarios, CUDRT enables the development of adaptable detection systems suited for intelligent applications like online content moderation, educational tools, and multilingual information retrieval.

Future research should focus on integrating CUDRT with real-world intelligent systems, such as real-time content filtering and adaptive educational platforms. Developing lightweight, self-learning models capable of rapid inference will enhance adaptability to evolving LLM patterns and multilingual contexts. Emphasizing cross-lingual transfer learning and resource-efficient implementations can bridge performance gaps between high and low resource languages. By addressing critical challenges in multilingual LLM-generated text detection, CUDRT serves as a foundational tool for advancing intelligent systems, improving interaction, and supporting applications like multilingual interfaces and social intelligence systems.

\begin{acks}
This work was supported by the National Natural Science Foundation of China (Grant No. 72271233), the Fundamental Research Funds for the Central Universities (2024110591), Suzhou Key Laboratory of Artificial Intelligence and Social Governance Technologies (SZS2023007), Smart Social Governance Technology and Innovative Application Platform (YZCXPT2023101).
\end{acks}

\bibliographystyle{ACM-Reference-Format}
\bibliography{sample-base}


\begin{thebibliography}{57}

%%% the Web address.

\ifx \showCODEN    \undefined \def \showCODEN     #1{\unskip}     \fi
\ifx \showDOI      \undefined \def \showDOI       #1{#1}\fi
\ifx \showISBNx    \undefined \def \showISBNx     #1{\unskip}     \fi
\ifx \showISBNxiii \undefined \def \showISBNxiii  #1{\unskip}     \fi
\ifx \showISSN     \undefined \def \showISSN      #1{\unskip}     \fi
\ifx \showLCCN     \undefined \def \showLCCN      #1{\unskip}     \fi
\ifx \shownote     \undefined \def \shownote      #1{#1}          \fi
\ifx \showarticletitle \undefined \def \showarticletitle #1{#1}   \fi
\ifx \showURL      \undefined \def \showURL       {\relax}        \fi
% invisible to TeX
\providecommand\bibfield[2]{#2}
\providecommand\bibinfo[2]{#2}
\providecommand\natexlab[1]{#1}
\providecommand\showeprint[2][]{arXiv:#2}

\bibitem[Achiam et~al\mbox{.}(2023)]%
        {achiam2023gpt}
\bibfield{author}{\bibinfo{person}{Josh Achiam}, \bibinfo{person}{Steven Adler}, \bibinfo{person}{Sandhini Agarwal}, \bibinfo{person}{Lama Ahmad}, \bibinfo{person}{Ilge Akkaya}, \bibinfo{person}{Florencia~Leoni Aleman}, \bibinfo{person}{Diogo Almeida}, \bibinfo{person}{Janko Altenschmidt}, \bibinfo{person}{Sam Altman}, \bibinfo{person}{Shyamal Anadkat}, {et~al\mbox{.}}} \bibinfo{year}{2023}\natexlab{}.
\newblock \showarticletitle{Gpt-4 technical report}.
\newblock \bibinfo{journal}{\emph{arXiv preprint arXiv:2303.08774}} (\bibinfo{year}{2023}).
\newblock


\bibitem[Alowais et~al\mbox{.}(2023)]%
        {2023revolutionizing}
\bibfield{author}{\bibinfo{person}{Shuroug~A Alowais}, \bibinfo{person}{Sahar~S Alghamdi}, \bibinfo{person}{Nada Alsuhebany}, \bibinfo{person}{Tariq Alqahtani}, \bibinfo{person}{Abdulrahman~I Alshaya}, \bibinfo{person}{Sumaya~N Almohareb}, \bibinfo{person}{Atheer Aldairem}, \bibinfo{person}{Mohammed Alrashed}, \bibinfo{person}{Khalid Bin~Saleh}, \bibinfo{person}{Hisham~A Badreldin}, {et~al\mbox{.}}} \bibinfo{year}{2023}\natexlab{}.
\newblock \showarticletitle{Revolutionizing healthcare: the role of artificial intelligence in clinical practice}.
\newblock \bibinfo{journal}{\emph{BMC medical education}} \bibinfo{volume}{23}, \bibinfo{number}{1} (\bibinfo{year}{2023}), \bibinfo{pages}{689}.
\newblock


\bibitem[Bahak et~al\mbox{.}(2023)]%
        {2023evaluating}
\bibfield{author}{\bibinfo{person}{Hossein Bahak}, \bibinfo{person}{Farzaneh Taheri}, \bibinfo{person}{Zahra Zojaji}, {and} \bibinfo{person}{Arefeh Kazemi}.} \bibinfo{year}{2023}\natexlab{}.
\newblock \showarticletitle{Evaluating ChatGPT as a Question Answering System: A Comprehensive Analysis and Comparison with Existing Models}.
\newblock \bibinfo{journal}{\emph{arXiv preprint arXiv:2312.07592}} (\bibinfo{year}{2023}).
\newblock


\bibitem[Bai et~al\mbox{.}(2023)]%
        {bai2023qwen}
\bibfield{author}{\bibinfo{person}{Jinze Bai}, \bibinfo{person}{Shuai Bai}, \bibinfo{person}{Yunfei Chu}, \bibinfo{person}{Zeyu Cui}, \bibinfo{person}{Kai Dang}, \bibinfo{person}{Xiaodong Deng}, \bibinfo{person}{Yang Fan}, \bibinfo{person}{Wenbin Ge}, \bibinfo{person}{Yu Han}, \bibinfo{person}{Fei Huang}, {et~al\mbox{.}}} \bibinfo{year}{2023}\natexlab{}.
\newblock \showarticletitle{Qwen technical report}.
\newblock \bibinfo{journal}{\emph{arXiv preprint arXiv:2309.16609}} (\bibinfo{year}{2023}).
\newblock


\bibitem[Bai et~al\mbox{.}(2024)]%
        {bai2024benchmarking}
\bibfield{author}{\bibinfo{person}{Yushi Bai}, \bibinfo{person}{Jiahao Ying}, \bibinfo{person}{Yixin Cao}, \bibinfo{person}{Xin Lv}, \bibinfo{person}{Yuze He}, \bibinfo{person}{Xiaozhi Wang}, \bibinfo{person}{Jifan Yu}, \bibinfo{person}{Kaisheng Zeng}, \bibinfo{person}{Yijia Xiao}, \bibinfo{person}{Haozhe Lyu}, {et~al\mbox{.}}} \bibinfo{year}{2024}\natexlab{}.
\newblock \showarticletitle{Benchmarking foundation models with language-model-as-an-examiner}.
\newblock \bibinfo{journal}{\emph{Advances in Neural Information Processing Systems}}  \bibinfo{volume}{36} (\bibinfo{year}{2024}).
\newblock


\bibitem[Chang et~al\mbox{.}(2024)]%
        {chang2024survey}
\bibfield{author}{\bibinfo{person}{Yupeng Chang}, \bibinfo{person}{Xu Wang}, \bibinfo{person}{Jindong Wang}, \bibinfo{person}{Yuan Wu}, \bibinfo{person}{Linyi Yang}, \bibinfo{person}{Kaijie Zhu}, \bibinfo{person}{Hao Chen}, \bibinfo{person}{Xiaoyuan Yi}, \bibinfo{person}{Cunxiang Wang}, \bibinfo{person}{Yidong Wang}, {et~al\mbox{.}}} \bibinfo{year}{2024}\natexlab{}.
\newblock \showarticletitle{A survey on evaluation of large language models}.
\newblock \bibinfo{journal}{\emph{ACM Transactions on Intelligent Systems and Technology}} \bibinfo{volume}{15}, \bibinfo{number}{3} (\bibinfo{year}{2024}), \bibinfo{pages}{1--45}.
\newblock


\bibitem[Chen et~al\mbox{.}(2023)]%
        {chen2023can}
\bibfield{author}{\bibinfo{person}{Yufan Chen}, \bibinfo{person}{Arjun Arunasalam}, {and} \bibinfo{person}{Z~Berkay Celik}.} \bibinfo{year}{2023}\natexlab{}.
\newblock \showarticletitle{Can large language models provide security \& privacy advice? measuring the ability of llms to refute misconceptions}. In \bibinfo{booktitle}{\emph{Proceedings of the 39th Annual Computer Security Applications Conference}}. \bibinfo{pages}{366--378}.
\newblock


\bibitem[Clusmann et~al\mbox{.}(2023)]%
        {clusmann2023future}
\bibfield{author}{\bibinfo{person}{Jan Clusmann}, \bibinfo{person}{Fiona~R Kolbinger}, \bibinfo{person}{Hannah~Sophie Muti}, \bibinfo{person}{Zunamys~I Carrero}, \bibinfo{person}{Jan-Niklas Eckardt}, \bibinfo{person}{Narmin~Ghaffari Laleh}, \bibinfo{person}{Chiara Maria~Lavinia L{\"o}ffler}, \bibinfo{person}{Sophie-Caroline Schwarzkopf}, \bibinfo{person}{Michaela Unger}, \bibinfo{person}{Gregory~P Veldhuizen}, {et~al\mbox{.}}} \bibinfo{year}{2023}\natexlab{}.
\newblock \showarticletitle{The future landscape of large language models in medicine}.
\newblock \bibinfo{journal}{\emph{Communications medicine}} \bibinfo{volume}{3}, \bibinfo{number}{1} (\bibinfo{year}{2023}), \bibinfo{pages}{141}.
\newblock


\bibitem[Du et~al\mbox{.}(2022)]%
        {glm}
\bibfield{author}{\bibinfo{person}{Zhengxiao Du}, \bibinfo{person}{Yujie Qian}, \bibinfo{person}{Xiao Liu}, \bibinfo{person}{Ming Ding}, \bibinfo{person}{Jiezhong Qiu}, \bibinfo{person}{Zhilin Yang}, {and} \bibinfo{person}{Jie Tang}.} \bibinfo{year}{2022}\natexlab{}.
\newblock \showarticletitle{{GLM}: General Language Model Pretraining with Autoregressive Blank Infilling}. In \bibinfo{booktitle}{\emph{Proceedings of the 60th Annual Meeting of the Association for Computational Linguistics (Volume 1: Long Papers)}}. \bibinfo{publisher}{Association for Computational Linguistics}, \bibinfo{pages}{320--335}.
\newblock


\bibitem[Gehrmann et~al\mbox{.}(2019)]%
        {gltr}
\bibfield{author}{\bibinfo{person}{Sebastian Gehrmann}, \bibinfo{person}{Hendrik Strobelt}, {and} \bibinfo{person}{Alexander Rush}.} \bibinfo{year}{2019}\natexlab{}.
\newblock \showarticletitle{{GLTR}: Statistical Detection and Visualization of Generated Text}. In \bibinfo{booktitle}{\emph{Proceedings of the 57th Annual Meeting of the Association for Computational Linguistics: System Demonstrations}}. \bibinfo{publisher}{Association for Computational Linguistics}, \bibinfo{pages}{111--116}.
\newblock


\bibitem[Guo et~al\mbox{.}(2021)]%
        {guo2021conditional}
\bibfield{author}{\bibinfo{person}{Bin Guo}, \bibinfo{person}{Hao Wang}, \bibinfo{person}{Yasan Ding}, \bibinfo{person}{Wei Wu}, \bibinfo{person}{Shaoyang Hao}, \bibinfo{person}{Yueqi Sun}, {and} \bibinfo{person}{Zhiwen Yu}.} \bibinfo{year}{2021}\natexlab{}.
\newblock \showarticletitle{Conditional text generation for harmonious human-machine interaction}.
\newblock \bibinfo{journal}{\emph{ACM Transactions on Intelligent Systems and Technology (TIST)}} \bibinfo{volume}{12}, \bibinfo{number}{2} (\bibinfo{year}{2021}), \bibinfo{pages}{1--50}.
\newblock


\bibitem[Guo et~al\mbox{.}(2023)]%
        {guo2023close}
\bibfield{author}{\bibinfo{person}{Biyang Guo}, \bibinfo{person}{Xin Zhang}, \bibinfo{person}{Ziyuan Wang}, \bibinfo{person}{Minqi Jiang}, \bibinfo{person}{Jinran Nie}, \bibinfo{person}{Yuxuan Ding}, \bibinfo{person}{Jianwei Yue}, {and} \bibinfo{person}{Yupeng Wu}.} \bibinfo{year}{2023}\natexlab{}.
\newblock \showarticletitle{How close is chatgpt to human experts? comparison corpus, evaluation, and detection}.
\newblock \bibinfo{journal}{\emph{arXiv preprint arXiv:2301.07597}} (\bibinfo{year}{2023}).
\newblock


\bibitem[He et~al\mbox{.}(2023b)]%
        {he-etal-2023-blind}
\bibfield{author}{\bibinfo{person}{Tianxing He}, \bibinfo{person}{Jingyu Zhang}, \bibinfo{person}{Tianle Wang}, \bibinfo{person}{Sachin Kumar}, \bibinfo{person}{Kyunghyun Cho}, \bibinfo{person}{James Glass}, {and} \bibinfo{person}{Yulia Tsvetkov}.} \bibinfo{year}{2023}\natexlab{b}.
\newblock \showarticletitle{On the Blind Spots of Model-Based Evaluation Metrics for Text Generation}. \bibinfo{publisher}{Proceedings of the 61st Annual Meeting of the Association for Computational Linguistics (Volume 1: Long Papers)}, \bibinfo{pages}{12067--12097}.
\newblock


\bibitem[He et~al\mbox{.}(2023a)]%
        {he2023mgtbench}
\bibfield{author}{\bibinfo{person}{Xinlei He}, \bibinfo{person}{Xinyue Shen}, \bibinfo{person}{Zeyuan Chen}, \bibinfo{person}{Michael Backes}, {and} \bibinfo{person}{Yang Zhang}.} \bibinfo{year}{2023}\natexlab{a}.
\newblock \showarticletitle{Mgtbench: Benchmarking machine-generated text detection}.
\newblock \bibinfo{journal}{\emph{arXiv preprint arXiv:2303.14822}} (\bibinfo{year}{2023}).
\newblock


\bibitem[Heidt(2023)]%
        {heidt2023artificial}
\bibfield{author}{\bibinfo{person}{Amanda Heidt}.} \bibinfo{year}{2023}\natexlab{}.
\newblock \showarticletitle{Artificial-intelligence search engines wrangle academic literature}.
\newblock \bibinfo{journal}{\emph{Nature}} \bibinfo{volume}{620}, \bibinfo{number}{7973} (\bibinfo{year}{2023}), \bibinfo{pages}{456--457}.
\newblock


\bibitem[Hu et~al\mbox{.}(2022)]%
        {hu2022can}
\bibfield{author}{\bibinfo{person}{Yang Hu}, \bibinfo{person}{Adriane Chapman}, \bibinfo{person}{Guihua Wen}, {and} \bibinfo{person}{Dame~Wendy Hall}.} \bibinfo{year}{2022}\natexlab{}.
\newblock \showarticletitle{What can knowledge bring to machine learning?—a survey of low-shot learning for structured data}.
\newblock \bibinfo{journal}{\emph{ACM Transactions on Intelligent Systems and Technology (TIST)}} \bibinfo{volume}{13}, \bibinfo{number}{3} (\bibinfo{year}{2022}), \bibinfo{pages}{1--45}.
\newblock


\bibitem[Kaddour et~al\mbox{.}(2023)]%
        {kaddour2023challenges}
\bibfield{author}{\bibinfo{person}{Jean Kaddour}, \bibinfo{person}{Joshua Harris}, \bibinfo{person}{Maximilian Mozes}, \bibinfo{person}{Herbie Bradley}, \bibinfo{person}{Roberta Raileanu}, {and} \bibinfo{person}{Robert McHardy}.} \bibinfo{year}{2023}\natexlab{}.
\newblock \showarticletitle{Challenges and applications of large language models}.
\newblock \bibinfo{journal}{\emph{arXiv preprint arXiv:2307.10169}} (\bibinfo{year}{2023}).
\newblock


\bibitem[Khasawneh and jadallah~abed Khasawneh(2023)]%
        {khasawneh2023roles}
\bibfield{author}{\bibinfo{person}{Mohamad Ahmad~Saleem Khasawneh} {and} \bibinfo{person}{Yusra jadallah~abed Khasawneh}.} \bibinfo{year}{2023}\natexlab{}.
\newblock \showarticletitle{The Roles Of Formulaic Sequences And Discourse Markers In Academic Writing; Insights From Lecturers And Other Researchers}.
\newblock \bibinfo{journal}{\emph{Journal of Namibian Studies: History Politics Culture}}  \bibinfo{volume}{34} (\bibinfo{year}{2023}), \bibinfo{pages}{7102--7122}.
\newblock


\bibitem[Li et~al\mbox{.}(2024)]%
        {li2024flexkbqa}
\bibfield{author}{\bibinfo{person}{Zhenyu Li}, \bibinfo{person}{Sunqi Fan}, \bibinfo{person}{Yu Gu}, \bibinfo{person}{Xiuxing Li}, \bibinfo{person}{Zhichao Duan}, \bibinfo{person}{Bowen Dong}, \bibinfo{person}{Ning Liu}, {and} \bibinfo{person}{Jianyong Wang}.} \bibinfo{year}{2024}\natexlab{}.
\newblock \showarticletitle{Flexkbqa: A flexible llm-powered framework for few-shot knowledge base question answering}. \bibinfo{publisher}{Proceedings of the AAAI Conference on Artificial Intelligence}, \bibinfo{pages}{18608--18616}.
\newblock


\bibitem[Liang et~al\mbox{.}(2024)]%
        {liang2023uhgeval}
\bibfield{author}{\bibinfo{person}{Xun Liang}, \bibinfo{person}{Shichao Song}, \bibinfo{person}{Simin Niu}, \bibinfo{person}{Zhiyu Li}, \bibinfo{person}{Feiyu Xiong}, \bibinfo{person}{Bo Tang}, \bibinfo{person}{Yezhaohui Wang}, \bibinfo{person}{Dawei He}, \bibinfo{person}{Cheng Peng}, \bibinfo{person}{Zhonghao Wang}, {and} \bibinfo{person}{Haiying Deng}.} \bibinfo{year}{2024}\natexlab{}.
\newblock \showarticletitle{{UHGE}val: Benchmarking the Hallucination of {C}hinese Large Language Models via Unconstrained Generation}. In \bibinfo{booktitle}{\emph{Proceedings of the 62nd Annual Meeting of the Association for Computational Linguistics (Volume 1: Long Papers)}}. \bibinfo{publisher}{Association for Computational Linguistics}, \bibinfo{pages}{5266--5293}.
\newblock


\bibitem[Liu et~al\mbox{.}(2024)]%
        {liu2024semantic}
\bibfield{author}{\bibinfo{person}{Guangyuan Liu}, \bibinfo{person}{Hongyang Du}, \bibinfo{person}{Dusit Niyato}, \bibinfo{person}{Jiawen Kang}, \bibinfo{person}{Zehui Xiong}, \bibinfo{person}{Dong~In Kim}, {and} \bibinfo{person}{Xuemin Shen}.} \bibinfo{year}{2024}\natexlab{}.
\newblock \showarticletitle{Semantic communications for artificial intelligence generated content (AIGC) toward effective content creation}.
\newblock \bibinfo{journal}{\emph{IEEE Network}} (\bibinfo{year}{2024}).
\newblock


\bibitem[Liu et~al\mbox{.}(2023)]%
        {2023-coco}
\bibfield{author}{\bibinfo{person}{Xiaoming Liu}, \bibinfo{person}{Zhaohan Zhang}, \bibinfo{person}{Yichen Wang}, \bibinfo{person}{Hang Pu}, \bibinfo{person}{Yu Lan}, {and} \bibinfo{person}{Chao Shen}.} \bibinfo{year}{2023}\natexlab{}.
\newblock \showarticletitle{{C}o{C}o: Coherence-Enhanced Machine-Generated Text Detection Under Low Resource With Contrastive Learning}. In \bibinfo{booktitle}{\emph{Proceedings of the 2023 Conference on Empirical Methods in Natural Language Processing}}. \bibinfo{publisher}{Association for Computational Linguistics}, \bibinfo{pages}{16167--16188}.
\newblock


\bibitem[Liu et~al\mbox{.}(2019)]%
        {liu2019roberta}
\bibfield{author}{\bibinfo{person}{Yinhan Liu}, \bibinfo{person}{Myle Ott}, \bibinfo{person}{Naman Goyal}, \bibinfo{person}{Jingfei Du}, \bibinfo{person}{Mandar Joshi}, \bibinfo{person}{Danqi Chen}, \bibinfo{person}{Omer Levy}, \bibinfo{person}{Mike Lewis}, \bibinfo{person}{Luke Zettlemoyer}, {and} \bibinfo{person}{Veselin Stoyanov}.} \bibinfo{year}{2019}\natexlab{}.
\newblock \showarticletitle{Roberta: A robustly optimized bert pretraining approach}.
\newblock \bibinfo{journal}{\emph{arXiv preprint arXiv:1907.11692}} (\bibinfo{year}{2019}).
\newblock


\bibitem[Macko et~al\mbox{.}(2023)]%
        {macko2023multitude}
\bibfield{author}{\bibinfo{person}{Dominik Macko}, \bibinfo{person}{Robert Moro}, \bibinfo{person}{Adaku Uchendu}, \bibinfo{person}{Jason Lucas}, \bibinfo{person}{Michiharu Yamashita}, \bibinfo{person}{Mat{\'u}{\v{s}} Pikuliak}, \bibinfo{person}{Ivan Srba}, \bibinfo{person}{Thai Le}, \bibinfo{person}{Dongwon Lee}, \bibinfo{person}{Jakub Simko}, {and} \bibinfo{person}{Maria Bielikova}.} \bibinfo{year}{2023}\natexlab{}.
\newblock \showarticletitle{{MULTIT}u{DE}: Large-Scale Multilingual Machine-Generated Text Detection Benchmark}. In \bibinfo{booktitle}{\emph{Proceedings of the 2023 Conference on Empirical Methods in Natural Language Processing}}. \bibinfo{publisher}{Association for Computational Linguistics}, \bibinfo{pages}{9960--9987}.
\newblock


\bibitem[Manovich(2002)]%
        {manovich2002language}
\bibfield{author}{\bibinfo{person}{Lev Manovich}.} \bibinfo{year}{2002}\natexlab{}.
\newblock \bibinfo{booktitle}{\emph{The language of new media}}.
\newblock \bibinfo{publisher}{MIT press}.
\newblock


\bibitem[Marvin et~al\mbox{.}(2023)]%
        {marvin2023prompt}
\bibfield{author}{\bibinfo{person}{Ggaliwango Marvin}, \bibinfo{person}{Nakayiza Hellen}, \bibinfo{person}{Daudi Jjingo}, {and} \bibinfo{person}{Joyce Nakatumba-Nabende}.} \bibinfo{year}{2023}\natexlab{}.
\newblock \showarticletitle{Prompt Engineering in Large Language Models}. \bibinfo{publisher}{International Conference on Data Intelligence and Cognitive Informatics}, \bibinfo{pages}{387--402}.
\newblock


\bibitem[Mitchell et~al\mbox{.}(2023)]%
        {2023detectgpt}
\bibfield{author}{\bibinfo{person}{Eric Mitchell}, \bibinfo{person}{Yoonho Lee}, \bibinfo{person}{Alexander Khazatsky}, \bibinfo{person}{Christopher~D Manning}, {and} \bibinfo{person}{Chelsea Finn}.} \bibinfo{year}{2023}\natexlab{}.
\newblock \showarticletitle{Detectgpt: Zero-shot machine-generated text detection using probability curvature}. \bibinfo{publisher}{International Conference on Machine Learning}, \bibinfo{pages}{24950--24962}.
\newblock


\bibitem[Nam et~al\mbox{.}(2024)]%
        {nam2024using}
\bibfield{author}{\bibinfo{person}{Daye Nam}, \bibinfo{person}{Andrew Macvean}, \bibinfo{person}{Vincent Hellendoorn}, \bibinfo{person}{Bogdan Vasilescu}, {and} \bibinfo{person}{Brad Myers}.} \bibinfo{year}{2024}\natexlab{}.
\newblock \showarticletitle{Using an llm to help with code understanding}. \bibinfo{publisher}{Proceedings of the IEEE/ACM 46th International Conference on Software Engineering}, \bibinfo{pages}{1--13}.
\newblock


\bibitem[Ni et~al\mbox{.}(2023)]%
        {ni2023recent}
\bibfield{author}{\bibinfo{person}{Jinjie Ni}, \bibinfo{person}{Tom Young}, \bibinfo{person}{Vlad Pandelea}, \bibinfo{person}{Fuzhao Xue}, {and} \bibinfo{person}{Erik Cambria}.} \bibinfo{year}{2023}\natexlab{}.
\newblock \showarticletitle{Recent advances in deep learning based dialogue systems: A systematic survey}.
\newblock \bibinfo{journal}{\emph{Artificial intelligence review}} \bibinfo{volume}{56}, \bibinfo{number}{4} (\bibinfo{year}{2023}), \bibinfo{pages}{3055--3155}.
\newblock


\bibitem[Peng et~al\mbox{.}(2024)]%
        {peng2024hidding}
\bibfield{author}{\bibinfo{person}{Xinlin Peng}, \bibinfo{person}{Ying Zhou}, \bibinfo{person}{Ben He}, \bibinfo{person}{Le Sun}, {and} \bibinfo{person}{Yingfei Sun}.} \bibinfo{year}{2024}\natexlab{}.
\newblock \showarticletitle{Hidding the Ghostwriters: An Adversarial Evaluation of AI-Generated Student Essay Detection}.
\newblock \bibinfo{journal}{\emph{arXiv preprint arXiv:2402.00412}} (\bibinfo{year}{2024}).
\newblock


\bibitem[Pr{\"o}hl et~al\mbox{.}(2024)]%
        {prohl2024benchmarking}
\bibfield{author}{\bibinfo{person}{Thorsten Pr{\"o}hl}, \bibinfo{person}{Erik Putzier}, {and} \bibinfo{person}{R{\"u}diger Zarnekow}.} \bibinfo{year}{2024}\natexlab{}.
\newblock \showarticletitle{Benchmarking of LLM Detection: Comparing Two Competing Approaches}.
\newblock \bibinfo{journal}{\emph{arXiv preprint arXiv:2406.11670}} (\bibinfo{year}{2024}).
\newblock


\bibitem[Rouf et~al\mbox{.}(2024)]%
        {rouf2024instantops}
\bibfield{author}{\bibinfo{person}{Raphael Rouf}, \bibinfo{person}{Mohammadreza Rasolroveicy}, \bibinfo{person}{Marin Litoiu}, \bibinfo{person}{Seema Nagar}, \bibinfo{person}{Prateeti Mohapatra}, \bibinfo{person}{Pranjal Gupta}, {and} \bibinfo{person}{Ian Watts}.} \bibinfo{year}{2024}\natexlab{}.
\newblock \showarticletitle{InstantOps: A Joint Approach to System Failure Prediction and Root Cause Identification in Microserivces Cloud-Native Applications}. \bibinfo{publisher}{Proceedings of the 15th ACM/SPEC International Conference on Performance Engineering}, \bibinfo{pages}{119--129}.
\newblock


\bibitem[Sarzaeim et~al\mbox{.}(2023)]%
        {sarzaeim2023framework}
\bibfield{author}{\bibinfo{person}{Paria Sarzaeim}, \bibinfo{person}{Arya Doshi}, {and} \bibinfo{person}{Qusay Mahmoud}.} \bibinfo{year}{2023}\natexlab{}.
\newblock \showarticletitle{A Framework for Detecting AI-Generated Text in Research Publications}. \bibinfo{publisher}{Proceedings of the International Conference on Advanced Technologies}, \bibinfo{pages}{121--127}.
\newblock


\bibitem[Shi et~al\mbox{.}(2023)]%
        {shi2023chatgraph}
\bibfield{author}{\bibinfo{person}{Yucheng Shi}, \bibinfo{person}{Hehuan Ma}, \bibinfo{person}{Wenliang Zhong}, \bibinfo{person}{Qiaoyu Tan}, \bibinfo{person}{Gengchen Mai}, \bibinfo{person}{Xiang Li}, \bibinfo{person}{Tianming Liu}, {and} \bibinfo{person}{Junzhou Huang}.} \bibinfo{year}{2023}\natexlab{}.
\newblock \showarticletitle{Chatgraph: Interpretable text classification by converting chatgpt knowledge to graphs}. \bibinfo{publisher}{IEEE International Conference on Data Mining Workshops (ICDMW)}, \bibinfo{pages}{515--520}.
\newblock


\bibitem[Shu et~al\mbox{.}(2024)]%
        {shu2024rewritelm}
\bibfield{author}{\bibinfo{person}{Lei Shu}, \bibinfo{person}{Liangchen Luo}, \bibinfo{person}{Jayakumar Hoskere}, \bibinfo{person}{Yun Zhu}, \bibinfo{person}{Yinxiao Liu}, \bibinfo{person}{Simon Tong}, \bibinfo{person}{Jindong Chen}, {and} \bibinfo{person}{Lei Meng}.} \bibinfo{year}{2024}\natexlab{}.
\newblock \showarticletitle{Rewritelm: An instruction-tuned large language model for text rewriting}. \bibinfo{publisher}{Proceedings of the AAAI Conference on Artificial Intelligence}, \bibinfo{pages}{18970--18980}.
\newblock


\bibitem[Stokel-Walker and Van~Noorden(2023)]%
        {stokel2023chatgpt}
\bibfield{author}{\bibinfo{person}{Chris Stokel-Walker} {and} \bibinfo{person}{Richard Van~Noorden}.} \bibinfo{year}{2023}\natexlab{}.
\newblock \showarticletitle{What ChatGPT and generative AI mean for science}.
\newblock \bibinfo{journal}{\emph{Nature}} \bibinfo{volume}{614}, \bibinfo{number}{7947} (\bibinfo{year}{2023}), \bibinfo{pages}{214--216}.
\newblock


\bibitem[Su et~al\mbox{.}(2023b)]%
        {su2023detectllm}
\bibfield{author}{\bibinfo{person}{Jinyan Su}, \bibinfo{person}{Terry Zhuo}, \bibinfo{person}{Di Wang}, {and} \bibinfo{person}{Preslav Nakov}.} \bibinfo{year}{2023}\natexlab{b}.
\newblock \showarticletitle{{D}etect{LLM}: Leveraging Log Rank Information for Zero-Shot Detection of Machine-Generated Text}. In \bibinfo{booktitle}{\emph{Findings of the Association for Computational Linguistics: EMNLP 2023}}. \bibinfo{publisher}{Association for Computational Linguistics}, \bibinfo{pages}{12395--12412}.
\newblock


\bibitem[Su et~al\mbox{.}(2023a)]%
        {su2023hc3}
\bibfield{author}{\bibinfo{person}{Zhenpeng Su}, \bibinfo{person}{Xing Wu}, \bibinfo{person}{Wei Zhou}, \bibinfo{person}{Guangyuan Ma}, {and} \bibinfo{person}{Songlin Hu}.} \bibinfo{year}{2023}\natexlab{a}.
\newblock \showarticletitle{Hc3 plus: A semantic-invariant human chatgpt comparison corpus}.
\newblock \bibinfo{journal}{\emph{arXiv preprint arXiv:2309.02731}} (\bibinfo{year}{2023}).
\newblock


\bibitem[Team(2024)]%
        {llama3}
\bibfield{author}{\bibinfo{person}{Meta~AI Team}.} \bibinfo{year}{2024}\natexlab{}.
\newblock \showarticletitle{Meta Llama 3: The most capable openly available LLM to date}.
\newblock \bibinfo{journal}{\emph{Hugging Face}} (\bibinfo{year}{2024}).
\newblock
\urldef\tempurl%
\url{https://huggingface.co/meta-llama/Meta-Llama-3}
\showURL{%
\tempurl}


\bibitem[Tian et~al\mbox{.}(2023)]%
        {tian2023multiscale}
\bibfield{author}{\bibinfo{person}{Yuchuan Tian}, \bibinfo{person}{Hanting Chen}, \bibinfo{person}{Xutao Wang}, \bibinfo{person}{Zheyuan Bai}, \bibinfo{person}{Qinghua Zhang}, \bibinfo{person}{Ruifeng Li}, \bibinfo{person}{Chao Xu}, {and} \bibinfo{person}{Yunhe Wang}.} \bibinfo{year}{2023}\natexlab{}.
\newblock \showarticletitle{Multiscale positive-unlabeled detection of ai-generated texts}.
\newblock \bibinfo{journal}{\emph{arXiv preprint arXiv:2305.18149}} (\bibinfo{year}{2023}).
\newblock


\bibitem[Touvron et~al\mbox{.}(2023)]%
        {touvron2023llama}
\bibfield{author}{\bibinfo{person}{Hugo Touvron}, \bibinfo{person}{Thibaut Lavril}, \bibinfo{person}{Gautier Izacard}, \bibinfo{person}{Xavier Martinet}, \bibinfo{person}{Marie-Anne Lachaux}, \bibinfo{person}{Timoth{\'e}e Lacroix}, \bibinfo{person}{Baptiste Rozi{\`e}re}, \bibinfo{person}{Naman Goyal}, \bibinfo{person}{Eric Hambro}, \bibinfo{person}{Faisal Azhar}, {et~al\mbox{.}}} \bibinfo{year}{2023}\natexlab{}.
\newblock \showarticletitle{Llama: Open and efficient foundation language models}.
\newblock \bibinfo{journal}{\emph{arXiv preprint arXiv:2302.13971}} (\bibinfo{year}{2023}).
\newblock


\bibitem[Wang et~al\mbox{.}(2023a)]%
        {seqxgpt}
\bibfield{author}{\bibinfo{person}{Pengyu Wang}, \bibinfo{person}{Linyang Li}, \bibinfo{person}{Ke Ren}, \bibinfo{person}{Botian Jiang}, \bibinfo{person}{Dong Zhang}, {and} \bibinfo{person}{Xipeng Qiu}.} \bibinfo{year}{2023}\natexlab{a}.
\newblock \showarticletitle{{S}eq{XGPT}: Sentence-Level {AI}-Generated Text Detection}. \bibinfo{publisher}{Proceedings of the 2023 Conference on Empirical Methods in Natural Language Processing}, \bibinfo{pages}{1144--1156}.
\newblock


\bibitem[Wang(2023)]%
        {wang2023large}
\bibfield{author}{\bibinfo{person}{Yiheng Wang}.} \bibinfo{year}{2023}\natexlab{}.
\newblock \showarticletitle{Large Language Models Evaluate Machine Translation via Polishing}. \bibinfo{publisher}{2023 6th International Conference on Algorithms, Computing and Artificial Intelligence}, \bibinfo{pages}{158--163}.
\newblock


\bibitem[Wang et~al\mbox{.}(2024)]%
        {wang2024m4gt}
\bibfield{author}{\bibinfo{person}{Yuxia Wang}, \bibinfo{person}{Jonibek Mansurov}, \bibinfo{person}{Petar Ivanov}, \bibinfo{person}{Jinyan Su}, \bibinfo{person}{Artem Shelmanov}, \bibinfo{person}{Akim Tsvigun}, \bibinfo{person}{Osama Mohammed~Afzal}, \bibinfo{person}{Tarek Mahmoud}, \bibinfo{person}{Giovanni Puccetti}, \bibinfo{person}{Thomas Arnold}, \bibinfo{person}{Alham Aji}, \bibinfo{person}{Nizar Habash}, \bibinfo{person}{Iryna Gurevych}, {and} \bibinfo{person}{Preslav Nakov}.} \bibinfo{year}{2024}\natexlab{}.
\newblock \showarticletitle{{M}4{GT}-Bench: Evaluation Benchmark for Black-Box Machine-Generated Text Detection}. In \bibinfo{booktitle}{\emph{Proceedings of the 62nd Annual Meeting of the Association for Computational Linguistics (Volume 1: Long Papers)}}. \bibinfo{publisher}{Association for Computational Linguistics}, \bibinfo{pages}{3964--3992}.
\newblock


\bibitem[Wang et~al\mbox{.}(2023b)]%
        {wang2023m4}
\bibfield{author}{\bibinfo{person}{Yuxia Wang}, \bibinfo{person}{Jonibek Mansurov}, \bibinfo{person}{Petar Ivanov}, \bibinfo{person}{Jinyan Su}, \bibinfo{person}{Artem Shelmanov}, \bibinfo{person}{Akim Tsvigun}, \bibinfo{person}{Chenxi Whitehouse}, \bibinfo{person}{Osama~Mohammed Afzal}, \bibinfo{person}{Tarek Mahmoud}, \bibinfo{person}{Alham~Fikri Aji}, {et~al\mbox{.}}} \bibinfo{year}{2023}\natexlab{b}.
\newblock \showarticletitle{M4: Multi-generator, multi-domain, and multi-lingual black-box machine-generated text detection}.
\newblock \bibinfo{journal}{\emph{arXiv preprint arXiv:2305.14902}} (\bibinfo{year}{2023}).
\newblock


\bibitem[Xie et~al\mbox{.}(2023)]%
        {xienext}
\bibfield{author}{\bibinfo{person}{Zhuohan Xie}, \bibinfo{person}{Trevor Cohn}, {and} \bibinfo{person}{Jey~Han Lau}.} \bibinfo{year}{2023}\natexlab{}.
\newblock \showarticletitle{The Next Chapter: A Study of Large Language Models in Storytelling}. \bibinfo{publisher}{Proceedings of the 16th International Natural Language Generation Conference}, \bibinfo{pages}{323--351}.
\newblock


\bibitem[Xu et~al\mbox{.}(2023)]%
        {xu2023generalization}
\bibfield{author}{\bibinfo{person}{Han Xu}, \bibinfo{person}{Jie Ren}, \bibinfo{person}{Pengfei He}, \bibinfo{person}{Shenglai Zeng}, \bibinfo{person}{Yingqian Cui}, \bibinfo{person}{Amy Liu}, \bibinfo{person}{Hui Liu}, {and} \bibinfo{person}{Jiliang Tang}.} \bibinfo{year}{2023}\natexlab{}.
\newblock \showarticletitle{On the generalization of training-based chatgpt detection methods}.
\newblock \bibinfo{journal}{\emph{arXiv preprint arXiv:2310.01307}} (\bibinfo{year}{2023}).
\newblock


\bibitem[Yang et~al\mbox{.}(2023b)]%
        {yang2023baichuan}
\bibfield{author}{\bibinfo{person}{Aiyuan Yang}, \bibinfo{person}{Bin Xiao}, \bibinfo{person}{Bingning Wang}, \bibinfo{person}{Borong Zhang}, \bibinfo{person}{Ce Bian}, \bibinfo{person}{Chao Yin}, \bibinfo{person}{Chenxu Lv}, \bibinfo{person}{Da Pan}, \bibinfo{person}{Dian Wang}, \bibinfo{person}{Dong Yan}, {et~al\mbox{.}}} \bibinfo{year}{2023}\natexlab{b}.
\newblock \showarticletitle{Baichuan 2: Open large-scale language models}.
\newblock \bibinfo{journal}{\emph{arXiv preprint arXiv:2309.10305}} (\bibinfo{year}{2023}).
\newblock


\bibitem[Yang et~al\mbox{.}(2024)]%
        {yang2024harnessing}
\bibfield{author}{\bibinfo{person}{Jingfeng Yang}, \bibinfo{person}{Hongye Jin}, \bibinfo{person}{Ruixiang Tang}, \bibinfo{person}{Xiaotian Han}, \bibinfo{person}{Qizhang Feng}, \bibinfo{person}{Haoming Jiang}, \bibinfo{person}{Shaochen Zhong}, \bibinfo{person}{Bing Yin}, {and} \bibinfo{person}{Xia Hu}.} \bibinfo{year}{2024}\natexlab{}.
\newblock \showarticletitle{Harnessing the power of llms in practice: A survey on chatgpt and beyond}.
\newblock \bibinfo{journal}{\emph{ACM Transactions on Knowledge Discovery from Data}} \bibinfo{volume}{18}, \bibinfo{number}{6} (\bibinfo{year}{2024}), \bibinfo{pages}{1--32}.
\newblock


\bibitem[Yang et~al\mbox{.}(2023a)]%
        {yang2023chatgpt}
\bibfield{author}{\bibinfo{person}{Lingyi Yang}, \bibinfo{person}{Feng Jiang}, {and} \bibinfo{person}{Haizhou Li}.} \bibinfo{year}{2023}\natexlab{a}.
\newblock \showarticletitle{Is chatgpt involved in texts? measure the polish ratio to detect chatgpt-generated text}.
\newblock \bibinfo{journal}{\emph{APSIPA Transactions on Signal and Information Processing}} \bibinfo{volume}{13}, \bibinfo{number}{2} (\bibinfo{year}{2023}).
\newblock


\bibitem[Yang et~al\mbox{.}(2019)]%
        {yang2019xlnet}
\bibfield{author}{\bibinfo{person}{Zhilin Yang}, \bibinfo{person}{Zihang Dai}, \bibinfo{person}{Yiming Yang}, \bibinfo{person}{Jaime Carbonell}, \bibinfo{person}{Russ~R Salakhutdinov}, {and} \bibinfo{person}{Quoc~V Le}.} \bibinfo{year}{2019}\natexlab{}.
\newblock \showarticletitle{Xlnet: Generalized autoregressive pretraining for language understanding}.
\newblock \bibinfo{journal}{\emph{Advances in neural information processing systems}}  \bibinfo{volume}{32} (\bibinfo{year}{2019}).
\newblock


\bibitem[Zeng et~al\mbox{.}(2022)]%
        {zeng2022glm}
\bibfield{author}{\bibinfo{person}{Aohan Zeng}, \bibinfo{person}{Xiao Liu}, \bibinfo{person}{Zhengxiao Du}, \bibinfo{person}{Zihan Wang}, \bibinfo{person}{Hanyu Lai}, \bibinfo{person}{Ming Ding}, \bibinfo{person}{Zhuoyi Yang}, \bibinfo{person}{Yifan Xu}, \bibinfo{person}{Wendi Zheng}, \bibinfo{person}{Xiao Xia}, {et~al\mbox{.}}} \bibinfo{year}{2022}\natexlab{}.
\newblock \showarticletitle{Glm-130b: An open bilingual pre-trained model}.
\newblock \bibinfo{journal}{\emph{arXiv preprint arXiv:2210.02414}} (\bibinfo{year}{2022}).
\newblock


\bibitem[Zhang et~al\mbox{.}(2024)]%
        {gao2024llm}
\bibfield{author}{\bibinfo{person}{Qihui Zhang}, \bibinfo{person}{Chujie Gao}, \bibinfo{person}{Dongping Chen}, \bibinfo{person}{Yue Huang}, \bibinfo{person}{Yixin Huang}, \bibinfo{person}{Zhenyang Sun}, \bibinfo{person}{Shilin Zhang}, \bibinfo{person}{Weiye Li}, \bibinfo{person}{Zhengyan Fu}, \bibinfo{person}{Yao Wan}, {and} \bibinfo{person}{Lichao Sun}.} \bibinfo{year}{2024}\natexlab{}.
\newblock \showarticletitle{{LLM}-as-a-Coauthor: Can Mixed Human-Written and Machine-Generated Text Be Detected?}. In \bibinfo{booktitle}{\emph{Findings of the Association for Computational Linguistics: NAACL 2024}}. \bibinfo{publisher}{Association for Computational Linguistics}, \bibinfo{pages}{409--436}.
\newblock


\bibitem[Zhang et~al\mbox{.}(2023)]%
        {zhang2023visar}
\bibfield{author}{\bibinfo{person}{Zheng Zhang}, \bibinfo{person}{Jie Gao}, \bibinfo{person}{Ranjodh~Singh Dhaliwal}, {and} \bibinfo{person}{Toby Jia-Jun Li}.} \bibinfo{year}{2023}\natexlab{}.
\newblock \showarticletitle{Visar: A human-ai argumentative writing assistant with visual programming and rapid draft prototyping}. \bibinfo{publisher}{Proceedings of the 36th Annual ACM Symposium on User Interface Software and Technology}, \bibinfo{pages}{1--30}.
\newblock


\bibitem[Zhao et~al\mbox{.}(2024)]%
        {zhao2024explainability}
\bibfield{author}{\bibinfo{person}{Haiyan Zhao}, \bibinfo{person}{Hanjie Chen}, \bibinfo{person}{Fan Yang}, \bibinfo{person}{Ninghao Liu}, \bibinfo{person}{Huiqi Deng}, \bibinfo{person}{Hengyi Cai}, \bibinfo{person}{Shuaiqiang Wang}, \bibinfo{person}{Dawei Yin}, {and} \bibinfo{person}{Mengnan Du}.} \bibinfo{year}{2024}\natexlab{}.
\newblock \showarticletitle{Explainability for large language models: A survey}.
\newblock \bibinfo{journal}{\emph{ACM Transactions on Intelligent Systems and Technology}} \bibinfo{volume}{15}, \bibinfo{number}{2} (\bibinfo{year}{2024}), \bibinfo{pages}{1--38}.
\newblock


\bibitem[Zhao et~al\mbox{.}(2023)]%
        {zhao2023survey}
\bibfield{author}{\bibinfo{person}{Wayne~Xin Zhao}, \bibinfo{person}{Kun Zhou}, \bibinfo{person}{Junyi Li}, \bibinfo{person}{Tianyi Tang}, \bibinfo{person}{Xiaolei Wang}, \bibinfo{person}{Yupeng Hou}, \bibinfo{person}{Yingqian Min}, \bibinfo{person}{Beichen Zhang}, \bibinfo{person}{Junjie Zhang}, \bibinfo{person}{Zican Dong}, {et~al\mbox{.}}} \bibinfo{year}{2023}\natexlab{}.
\newblock \showarticletitle{A survey of large language models}.
\newblock \bibinfo{journal}{\emph{arXiv preprint arXiv:2303.18223}} (\bibinfo{year}{2023}).
\newblock


\bibitem[Zhou et~al\mbox{.}(2024)]%
        {zhou2024vision}
\bibfield{author}{\bibinfo{person}{Xingcheng Zhou}, \bibinfo{person}{Mingyu Liu}, \bibinfo{person}{Ekim Yurtsever}, \bibinfo{person}{Bare~Luka Zagar}, \bibinfo{person}{Walter Zimmer}, \bibinfo{person}{Hu Cao}, {and} \bibinfo{person}{Alois~C Knoll}.} \bibinfo{year}{2024}\natexlab{}.
\newblock \showarticletitle{Vision Language Models in Autonomous Driving: A Survey and Outlook}.
\newblock \bibinfo{journal}{\emph{IEEE Transactions on Intelligent Vehicles}} (\bibinfo{year}{2024}).
\newblock


\end{thebibliography}


\appendix
\clearpage
\section{Appendix}

\renewcommand{\thetable}{A\arabic{table}}
\renewcommand{\thefigure}{A\arabic{figure}}

% 重置计数器
\setcounter{table}{0}
\setcounter{figure}{0}

% Table generated by Excel2LaTeX from sheet 'Human Text'
\begin{table}[!h]
  \centering
  \caption{Detailed information on original human-generated texts.}
  \resizebox{11cm}{!}{
    \begin{tabular}{llll}
    \toprule
    \textbf{Language} & \multicolumn{1}{l}{\textbf{LLMs Operation}} & \multicolumn{1}{l}{\textbf{Data Composition and Source}} & \textbf{Scale} \\
    \midrule
    \multirow{3}[6]{*}{Chinese} & \multicolumn{1}{p{8em}}{Complete, Polish, Expand, Summary, Refine, Rewrite} & \multicolumn{1}{p{22.165em}}{News: Sina\tablefootnote{https://www.sina.com.cn/}, Guangming\tablefootnote{https://www.gmw.cn/
}; Thesis: Baidu Academic\tablefootnote{https://xueshu.baidu.com/}} & 65,324 \\
\cmidrule{2-4}          & QA    & \multicolumn{1}{p{22.165em}}{Community: Baidu Tieba\tablefootnote{https://tieba.baidu.com/}; Encyclopedia: Baidu Baike\tablefootnote{https://baike.baidu.com/}; Medicine: Meddialog\tablefootnote{https://github.com/Toyhom/Chinese-medical-dialogue-data}; Finance: ChineseNlpCorpus\tablefootnote{https://github.com/SophonPlus/ChineseNlpCorpus}} & 16,389 \\
\cmidrule{2-4}          & Translate & Translation: translation2019zh\tablefootnote{https://github.com/genjinshuaji/translation2019zh} & 5,200,000 \\
    \midrule
    \multirow{3}[6]{*}{English} & \multicolumn{1}{p{8em}}{Complete, Polish, Expand, Summary, Refine, Rewrite} & \multicolumn{1}{p{22.165em}}{News: BBC\tablefootnote{https://www.bbc.com/news}; Thesis: Arxiv\tablefootnote{https://arxiv.org/}} & 171,302 \\
\cmidrule{2-4}          & QA    & \multicolumn{1}{p{22.165em}}{Community: Reddit\tablefootnote{https://www.reddit.com/}; Encyclopedia: Wikipedia\tablefootnote{https://www.wikipedia.org/}; Medicine: Meddialog; Finance: FiQA\tablefootnote{https://dl.acm.org/doi/pdf/10.1145/3184558.3192301}} & 25,861 \\
\cmidrule{2-4}          & Translate & Translation: translation2019zh & 5,200,000 \\
    \bottomrule
    \end{tabular}}
  \label{tabA1}
\end{table}%

\begin{figure}[h]
    \centering
    \subfigure[English Complete]{
        \includegraphics[width=0.23\textwidth]{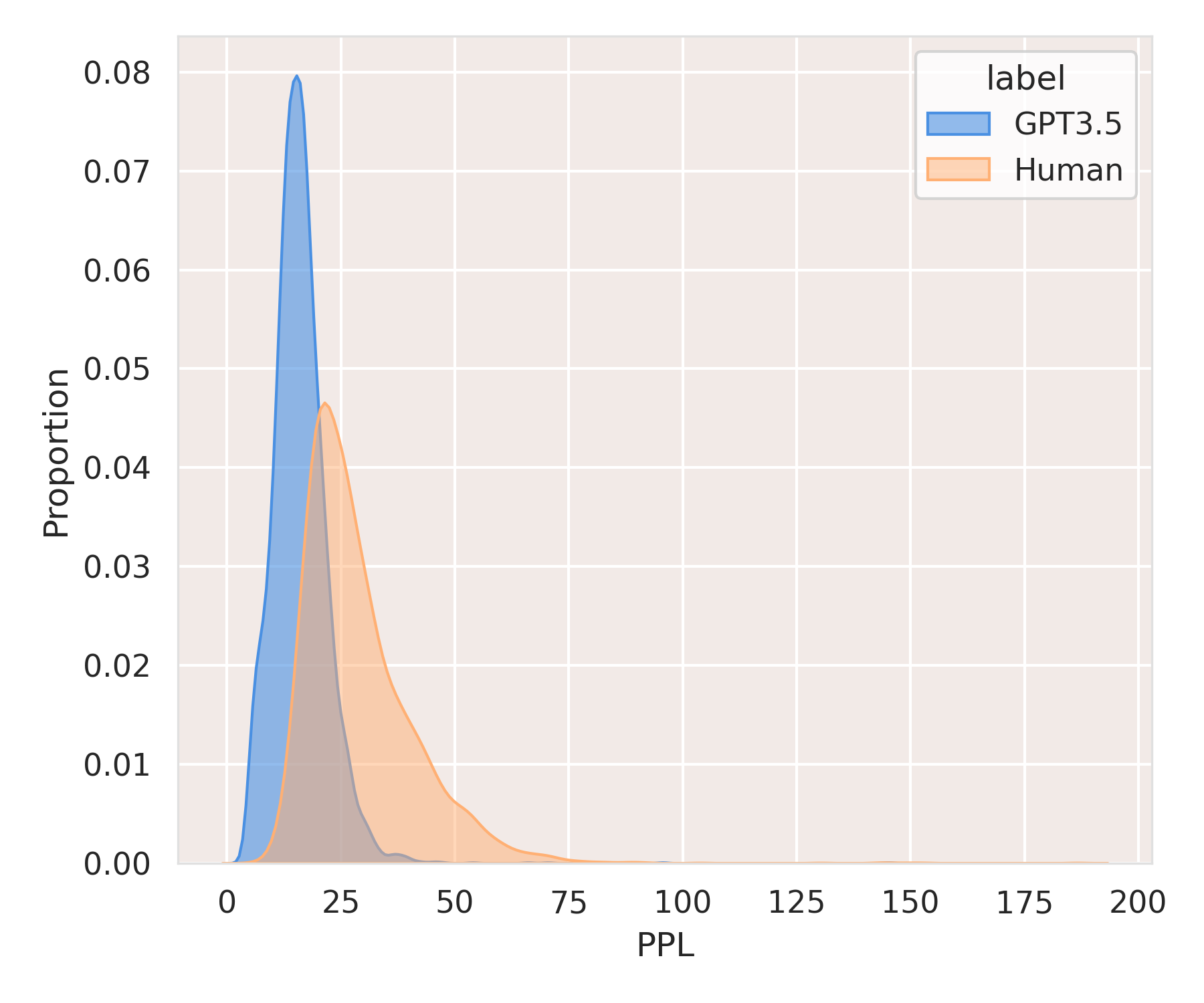}
    }
    \subfigure[English QA]{
        \includegraphics[width=0.23\textwidth]{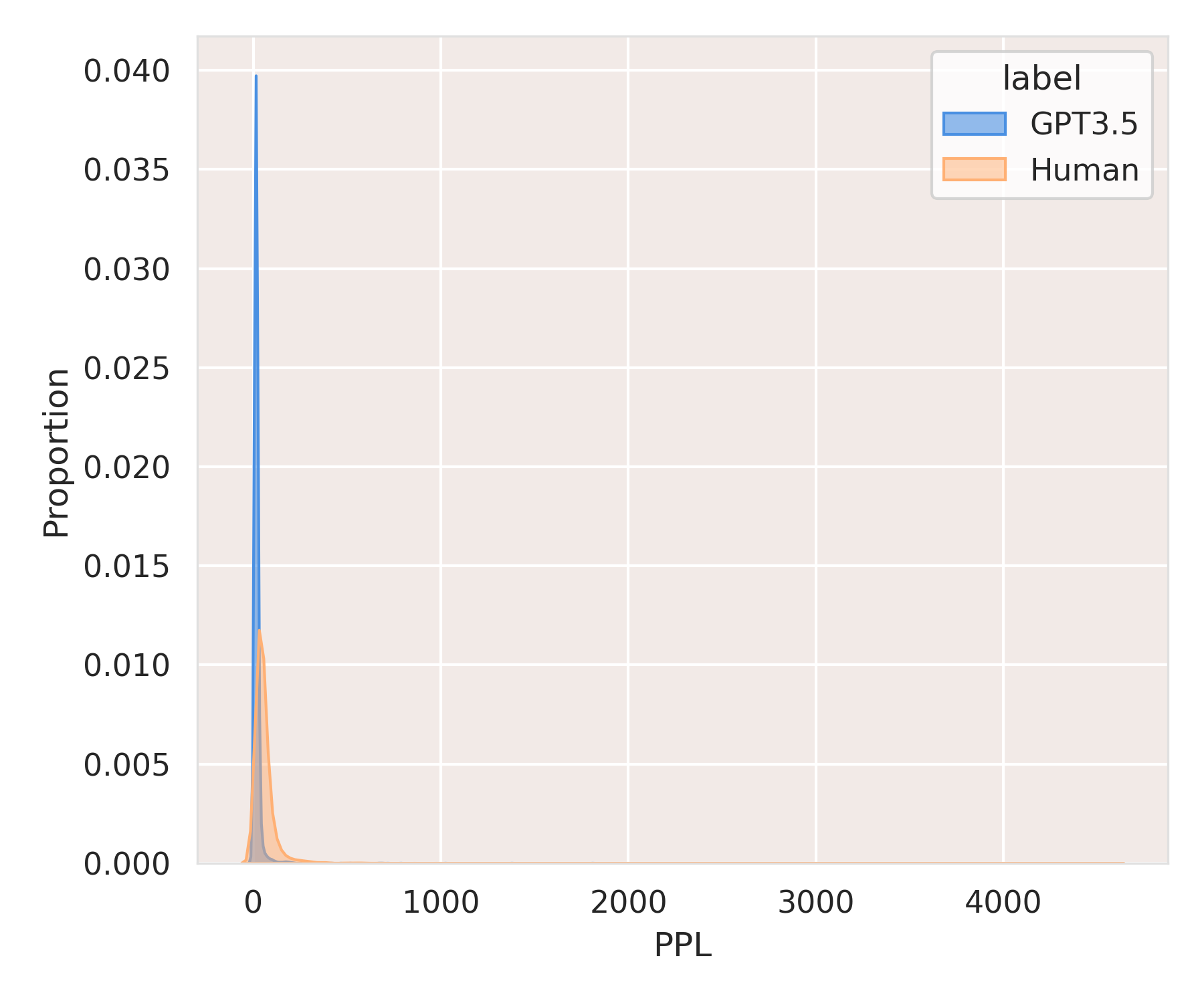}
    }
    \subfigure[English Polish]{
        \includegraphics[width=0.23\textwidth]{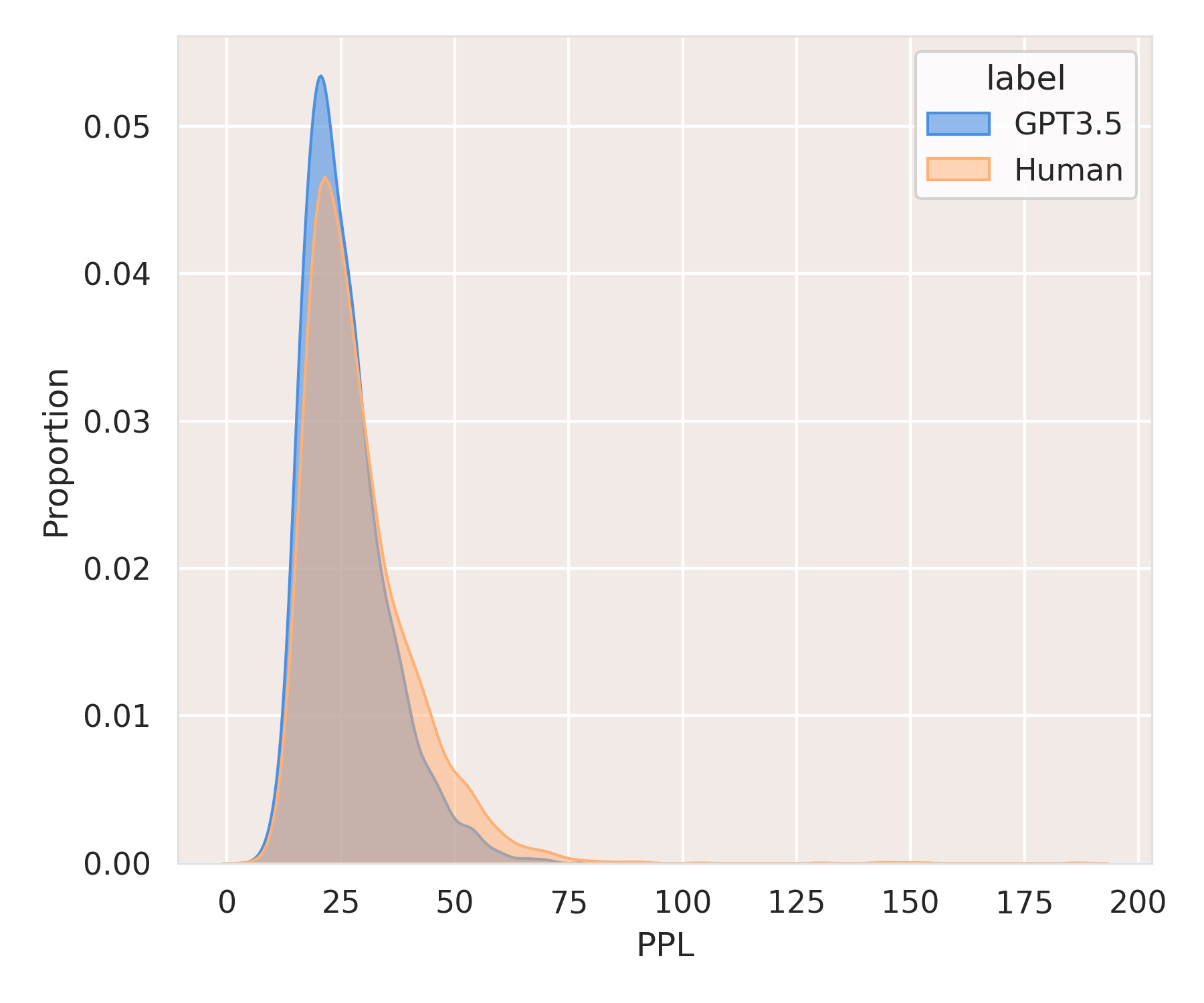}
    }
    \subfigure[English Expand]{
        \includegraphics[width=0.23\textwidth]{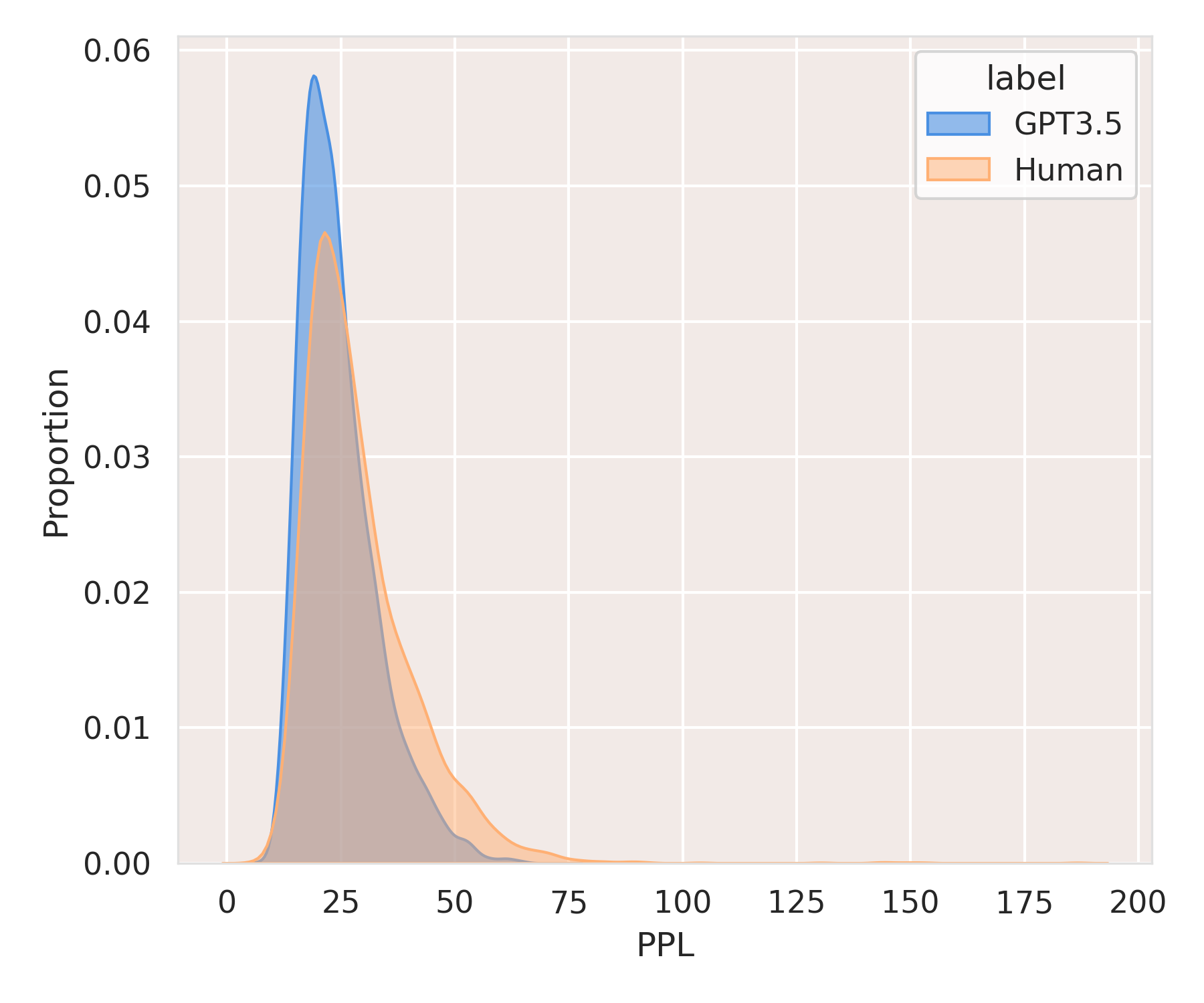}
    }

    \subfigure[English Summary]{
        \includegraphics[width=0.23\textwidth]{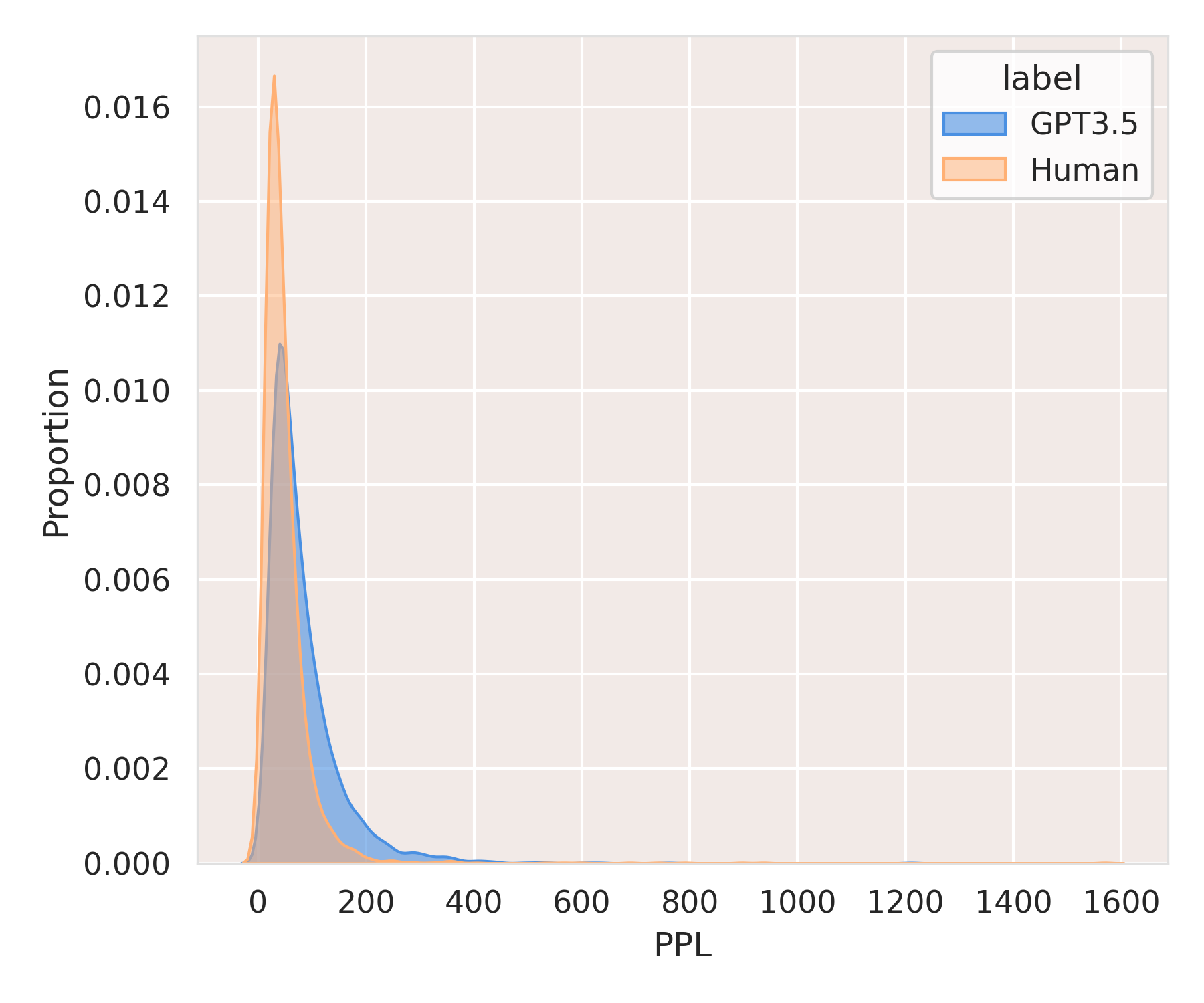}
    }
    \subfigure[English Refine]{
        \includegraphics[width=0.23\textwidth]{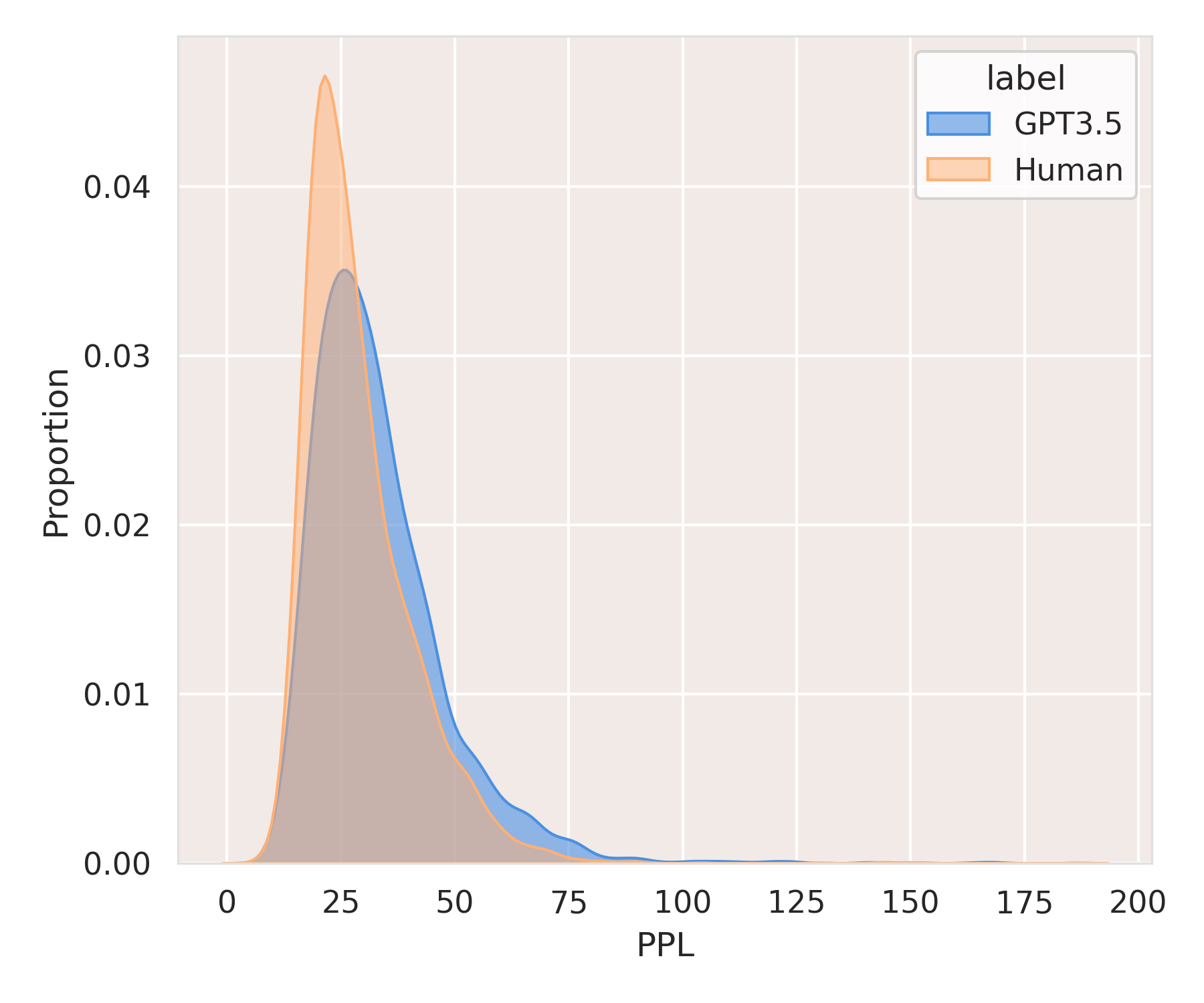}
    }
    \subfigure[English Rewrite]{
        \includegraphics[width=0.23\textwidth]{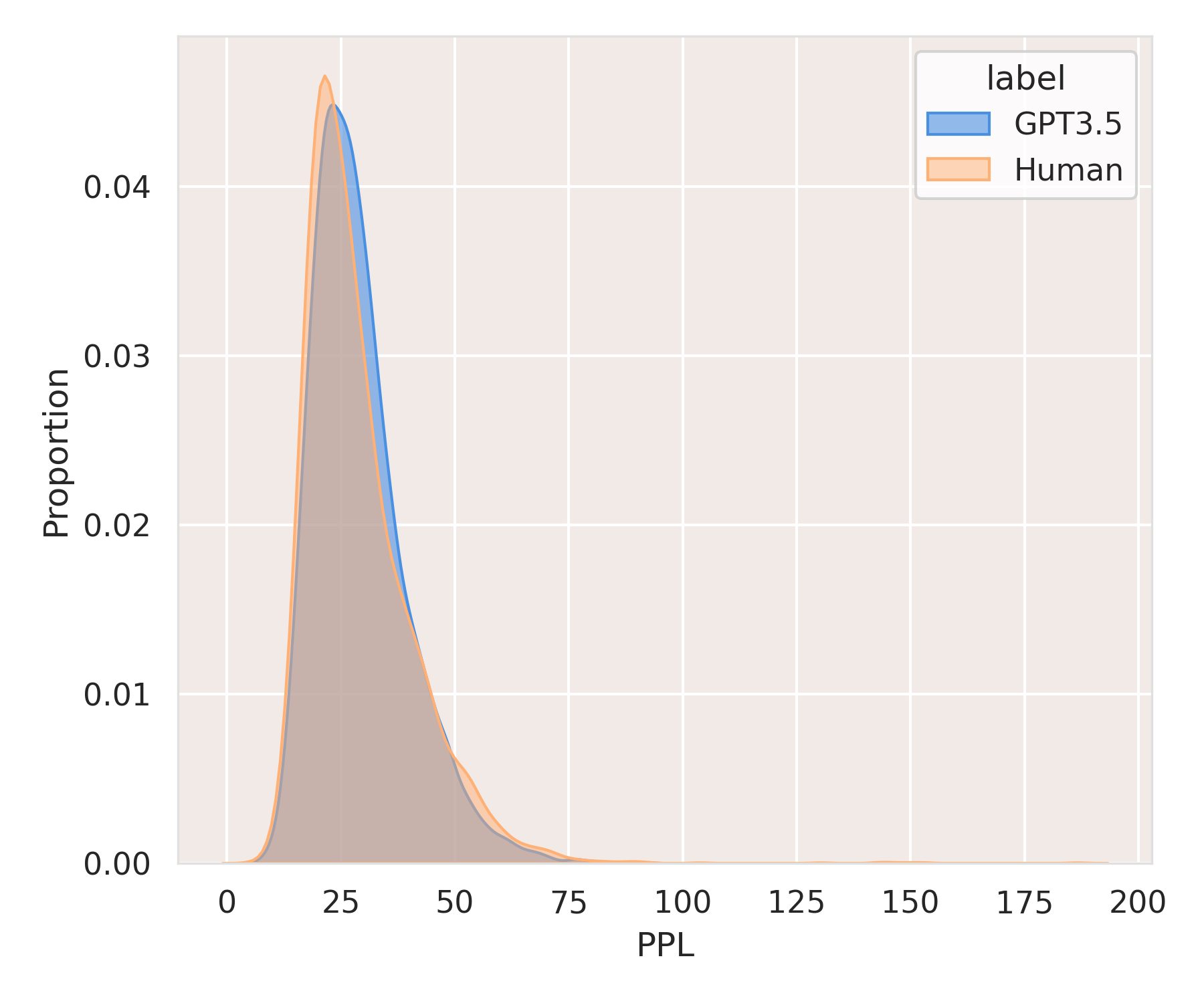}
    }
    \subfigure[English Translate]{
        \includegraphics[width=0.23\textwidth]{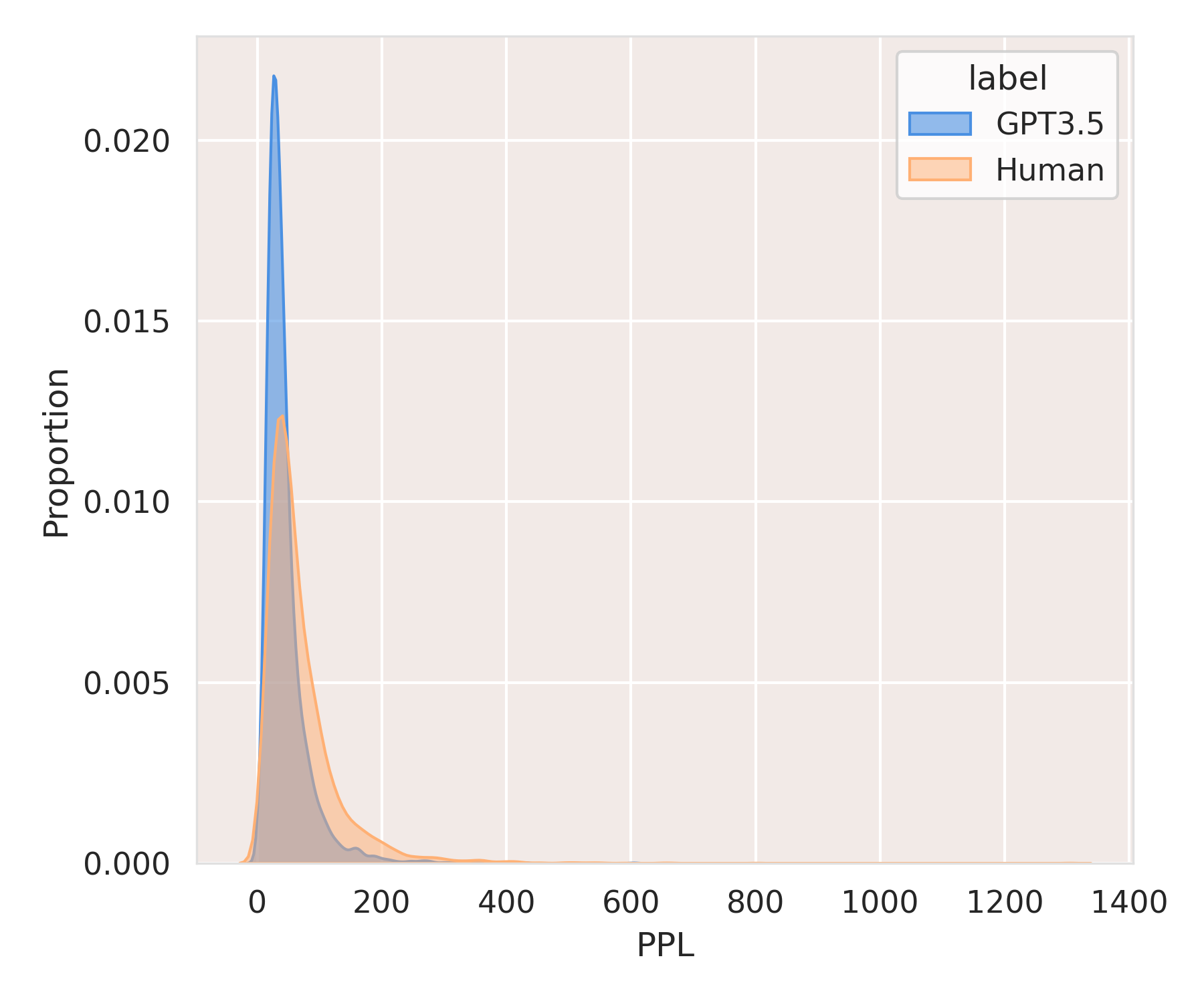}
    }
    \caption{PPL distributions on the proposed English dataset.}
    \label{figA1}
\end{figure}

% Table generated by Excel2LaTeX from sheet 'Cross_Operation_LLM_roberta'
\begin{table}[h]
  \centering
  \caption{Results of RoBERTa model trained on single operation data and tested on each LLM data.}
  \renewcommand{\arraystretch}{1.2}
  \resizebox{\textwidth}{!}{
    \begin{tabular}{l|lllll|lllll}
    \toprule
    \multirow{2}[2]{*}{\textbf{Train Data}} & \multirow{2}[2]{*}{\textbf{Test Data}} & \multicolumn{4}{c|}{\textbf{Chinese}} & \multirow{2}[2]{*}{\textbf{Test Data}} & \multicolumn{4}{c}{\textbf{English}} \\
          &       & Accuracy & Precision & Recall & F1-score &       & Accuracy & Precision & Recall & F1-score \\
    \midrule
    \midrule
    \multirow{6}[2]{*}{\textbf{Create}} & Baichuan2-13B & 0.6368  & 0.6124  & 0.5403  & 0.5686  & Baichuan2-13B & 0.8056  & 0.7796  & 0.7889  & 0.7781  \\
          & ChatGLM3-6B-32K & 0.6411  & 0.6123  & 0.5434  & 0.5697  & ChatGLM3-6B-32K & 0.8245  & 0.7855  & 0.8196  & 0.7948  \\
          & GPT-3.5-Turbo & 0.6506  & 0.6184  & 0.5585  & 0.5810  & GPT-3.5-Turbo & 0.8215  & 0.7841  & 0.8215  & 0.7947  \\
          & GPT-4-1106 & \cellcolor{lightpink}\textbf{0.6578} & \cellcolor{lightpink}\textbf{0.6219} & \cellcolor{lightpink}\textbf{0.5671} & \cellcolor{lightpink}\textbf{0.5866} & Llama3-8B & 0.8256  & 0.7856  & 0.8242  & 0.7971  \\
          & Qwen1.5-32B & 0.6532  & 0.6189  & 0.5614  & 0.5819  & Llama2-13B & \cellcolor{lightpink}\textbf{0.8295} & \cellcolor{lightpink}\textbf{0.7882} & \cellcolor{lightpink}\textbf{0.8288} & \cellcolor{lightpink}\textbf{0.8009} \\
          \rowcolor{Ocean}
          & \textbf{All Average} & 0.6479  & 0.6168  & 0.5542  & 0.5776  & \textbf{All Average} & 0.8213  & 0.7846  & 0.8166  & 0.7931  \\
    \midrule
    \midrule
    \multirow{6}[2]{*}{\textbf{Update}} & Baichuan2-13B & 0.8112  & 0.8131  & 0.7935  & 0.7996  & Baichuan2-13B & 0.8609  & 0.8569  & 0.8497  & 0.8512  \\
          & ChatGLM3-6B-32K & 0.8169  & 0.8181  & 0.7998  & 0.8055  & ChatGLM3-6B-32K & 0.8770  & 0.8612  & 0.8780  & 0.8671  \\
          & GPT-3.5-Turbo & 0.8400  & 0.8211  & 0.8406  & 0.8289  & GPT-3.5-Turbo & 0.8835  & 0.8578  & 0.8950  & 0.8750  \\
          & GPT-4-1106 & 0.8617  & \cellcolor{lightpink}\textbf{0.8281} & 0.8763  & 0.8503  & Llama3-8B & 0.8802  & 0.8587  & 0.8854  & 0.8706  \\
          & Qwen1.5-32B & \cellcolor{lightpink}\textbf{0.8619} & 0.8264  & \cellcolor{lightpink}\textbf{0.8795} & \cellcolor{lightpink}\textbf{0.8506} & Llama2-13B & \cellcolor{lightpink}\textbf{0.8915} & \cellcolor{lightpink}\textbf{0.8645} & \cellcolor{lightpink}\textbf{0.9048} & \cellcolor{lightpink}\textbf{0.8829} \\
          \rowcolor{Ocean}
          & \textbf{All Average} & 0.8383  & 0.8214  & 0.8379  & 0.8270  & \textbf{All Average} & 0.8786  & 0.8598  & 0.8826  & 0.8694  \\
    \midrule
    \midrule
    \multirow{6}[2]{*}{\textbf{Delete}} & Baichuan2-13B & 0.7824  & 0.8725  & 0.6997  & 0.7670  & Baichuan2-13B & 0.8933  & 0.8915  & 0.8952  & 0.8918  \\
          & ChatGLM3-6B-32K & 0.8124  & 0.8885  & 0.7430  & 0.8036  & ChatGLM3-6B-32K & \cellcolor{lightpink}\textbf{0.9186} & 0.8978  & \cellcolor{lightpink}\textbf{0.9397} & \cellcolor{lightpink}\textbf{0.9167} \\
          & GPT-3.5-Turbo & 0.7983  & 0.8536  & 0.7286  & 0.7810  & GPT-3.5-Turbo & 0.9173  & 0.8959  & 0.9391  & 0.9158  \\
          & GPT-4-1106 & 0.8587  & 0.8859  & \cellcolor{lightpink}\textbf{0.8199} & 0.8499  & Llama3-8B & 0.9130  & 0.8937  & 0.9320  & 0.9116  \\
          & Qwen1.5-32B & \cellcolor{lightpink}\textbf{0.8589} & \cellcolor{lightpink}\textbf{0.8914} & 0.8176  & \cellcolor{lightpink}\textbf{0.8512} & Llama2-13B & 0.9164  & \cellcolor{lightpink}\textbf{0.9002} & 0.9336  & 0.9148  \\
          \rowcolor{Ocean}
          & \textbf{All Average} & 0.8221  & 0.8784  & 0.7617  & 0.8105  & \textbf{All Average} & 0.9117  & 0.8958  & 0.9279  & 0.9101  \\
    \midrule
    \midrule
    \multirow{6}[2]{*}{\textbf{Rewrite}} & Baichuan2-13B & 0.7983  & 0.8109  & 0.7688  & 0.7839  & Baichuan2-13B & 0.8389  & 0.8487  & 0.7934  & 0.8186  \\
          & ChatGLM3-6B-32K & 0.8001  & 0.8186  & 0.7660  & 0.7872  & ChatGLM3-6B-32K & 0.8623  & 0.8535  & 0.8363  & 0.8418  \\
          & GPT-3.5-Turbo & 0.8227  & 0.8181  & 0.8071  & 0.8088  & GPT-3.5-Turbo & 0.8698  & 0.8512  & \cellcolor{lightpink}\textbf{0.8543} & \cellcolor{lightpink}\textbf{0.8506} \\
          & GPT-4-1106 & 0.8637  & \cellcolor{lightpink}\textbf{0.8346} & 0.8739  & 0.8526  & Llama3-8B & 0.8675  & 0.8519  & 0.8474  & 0.8475  \\
          & Qwen1.5-32B & \cellcolor{lightpink}\textbf{0.8671} & 0.8340  & \cellcolor{lightpink}\textbf{0.8814} & \cellcolor{lightpink}\textbf{0.8559} & Llama2-13B & \cellcolor{lightpink}\textbf{0.8701} & \cellcolor{lightpink}\textbf{0.8547} & 0.8504  & 0.8502  \\
          \rowcolor{Ocean}
          & \textbf{All Average} & 0.8304  & 0.8233  & 0.8195  & 0.8177  & \textbf{All Average} & 0.8617  & 0.8520  & 0.8364  & 0.8417  \\
    \midrule
    \midrule
    \multirow{6}[2]{*}{\textbf{Translate}} & Baichuan2-13B & 0.6627  & 0.6945  & 0.6190  & 0.6524  & Baichuan2-13B & 0.6328  & 0.6221  & 0.4783  & 0.5259  \\
          & ChatGLM3-6B-32K & 0.7076  & 0.7277  & 0.6834  & 0.7032  & ChatGLM3-6B-32K & \cellcolor{lightpink}\textbf{0.6370} & 0.6239  & \cellcolor{lightpink}\textbf{0.4841} & \cellcolor{lightpink}\textbf{0.5297} \\
          & GPT-3.5-Turbo & 0.6419  & 0.6679  & 0.5971  & 0.6289  & GPT-3.5-Turbo & 0.6247  & 0.6206  & 0.4618  & 0.5169  \\
          & GPT-4-1106 & \cellcolor{lightpink}\textbf{0.7372} & \cellcolor{lightpink}\textbf{0.7466} & \cellcolor{lightpink}\textbf{0.7248} & \cellcolor{lightpink}\textbf{0.7343} & Llama3-8B & 0.6345  & 0.6239  & 0.4774  & 0.5261  \\
          & Qwen1.5-32B & 0.6747  & 0.7115  & 0.6319  & 0.6681  & Llama2-13B & 0.6369  & \cellcolor{lightpink}\textbf{0.6244} & 0.4823  & 0.5287  \\
          \rowcolor{Ocean}
          & \textbf{All Average} & 0.6848  & 0.7096  & 0.6513  & 0.6774  & \textbf{All Average} & 0.6332  & 0.6230  & 0.4768  & 0.5255  \\
    \bottomrule
    \end{tabular}}
  \label{tabA2}%
\end{table}%

\newpage

% Table generated by Excel2LaTeX from sheet 'Cross_Operation_LLM_xlnet'
\begin{table}[h]
  \centering
  \caption{Results of XLNet model trained on single operation data and tested on each LLM data.}
  \renewcommand{\arraystretch}{1.2}
  \resizebox{\textwidth}{!}{
    \begin{tabular}{l|lllll|lllll}
    \toprule
    \multicolumn{1}{l|}{\multirow{2}[2]{*}{\textbf{Train Data}}} & \multirow{2}[2]{*}{\textbf{Test Data}} & \multicolumn{4}{c|}{\textbf{Chinese}} & \multirow{2}[2]{*}{\textbf{Test Data}} & \multicolumn{4}{c}{\textbf{English}} \\
          &       & Accuracy & Precision & Recall & F1-score &       & Accuracy & Precision & Recall & F1-score \\
    \midrule
    \midrule
    \multicolumn{1}{l|}{\multirow{6}[2]{*}{\textbf{Create}}} & Baichuan2-13B & 0.6473  & 0.6392  & 0.5423  & 0.5774  & Baichuan2-13B & 0.8283  & 0.8347  & 0.7818  & 0.8036  \\
          & ChatGLM3-6B-32K & 0.6599  & 0.6454  & 0.5512  & 0.5844  & ChatGLM3-6B-32K & 0.8547  & \cellcolor{lightpink}\textbf{0.8410} & 0.8290  & 0.8303  \\
          & GPT-3.5-Turbo & 0.6555  & 0.6417  & 0.5495  & 0.5824  & GPT-3.5-Turbo & 0.8471  & 0.8306  & 0.8342  & 0.8273  \\
          & GPT-4-1106 & \cellcolor{lightpink}\textbf{0.6686} & \cellcolor{lightpink}\textbf{0.6497} & \cellcolor{lightpink}\textbf{0.5636} & \cellcolor{lightpink}\textbf{0.5931} & Llama3-8B & \cellcolor{lightpink}\textbf{0.8621} & 0.8388  & \cellcolor{lightpink}\textbf{0.8457} & \cellcolor{lightpink}\textbf{0.8374} \\
          & Qwen1.5-32B & 0.6615  & 0.6452  & 0.5510  & 0.5846  & Llama2-13B & 0.8619  & 0.8407  & 0.8424  & 0.8368  \\
          \rowcolor{Ocean}
          & \textbf{All Average} & 0.6585  & 0.6443  & 0.5515  & 0.5844  & \textbf{All Average} & 0.8508  & 0.8372  & 0.8266  & 0.8271  \\
    \midrule
    \midrule
    \multicolumn{1}{l|}{\multirow{6}[2]{*}{\textbf{Update}}} & Baichuan2-13B & 0.7261  & 0.7622  & 0.6038  & 0.6635  & Baichuan2-13B & 0.8445  & 0.8388  & 0.8213  & 0.8296  \\
          & ChatGLM3-6B-32K & 0.7501  & 0.7698  & 0.6413  & 0.6883  & ChatGLM3-6B-32K & 0.8587  & 0.8436  & 0.8466  & 0.8434  \\
          & GPT-3.5-Turbo & 0.7666  & 0.7698  & 0.6707  & 0.7040  & GPT-3.5-Turbo & 0.8609  & 0.8381  & 0.8585  & 0.8472  \\
          & GPT-4-1106 & 0.7841  & \cellcolor{lightpink}\textbf{0.7708} & 0.7041  & 0.7209  & Llama3-8B & 0.8670  & 0.8439  & 0.8624  & 0.8523  \\
          & Qwen1.5-32B & \cellcolor{lightpink}\textbf{0.7842} & 0.7706  & \cellcolor{lightpink}\textbf{0.7042} & \cellcolor{lightpink}\textbf{0.7210} & Llama2-13B & \cellcolor{lightpink}\textbf{0.8706} & \cellcolor{lightpink}\textbf{0.8449} & \cellcolor{lightpink}\textbf{0.8686} & \cellcolor{lightpink}\textbf{0.8556} \\
          \rowcolor{Ocean}
          & \textbf{All Average} & 0.7622  & 0.7686  & 0.6648  & 0.6995  & \textbf{All Average} & 0.8603  & 0.8419  & 0.8515  & 0.8456  \\
    \midrule
    \midrule
    \multicolumn{1}{l|}{\multirow{6}[2]{*}{\textbf{Delete}}} & Baichuan2-13B & 0.7502  & 0.8493  & 0.6598  & 0.7297  & Baichuan2-13B & 0.8736  & 0.8730  & 0.8647  & 0.8689  \\
          & ChatGLM3-6B-32K & \cellcolor{lightpink}\textbf{0.8136} & \cellcolor{lightpink}\textbf{0.8901} & 0.7487  & \cellcolor{lightpink}\textbf{0.8070} & ChatGLM3-6B-32K & \cellcolor{lightpink}\textbf{0.9078} & \cellcolor{lightpink}\textbf{0.8846} & \cellcolor{lightpink}\textbf{0.9246} & \cellcolor{lightpink}\textbf{0.9029} \\
          & GPT-3.5-Turbo & 0.7715  & 0.8353  & 0.7005  & 0.7487  & GPT-3.5-Turbo & 0.8908  & 0.8763  & 0.8967  & 0.8854  \\
          & GPT-4-1106 & 0.8073  & 0.8498  & \cellcolor{lightpink}\textbf{0.7537} & 0.7904  & Llama3-8B & 0.8896  & 0.8762  & 0.8947  & 0.8847  \\
          & Qwen1.5-32B & 0.7711  & 0.8473  & 0.6978  & 0.7488  & Llama2-13B & 0.8993  & 0.8839  & 0.9072  & 0.8942  \\
          \rowcolor{Ocean}
          & \textbf{All Average} & 0.7827  & 0.8544  & 0.7121  & 0.7649  & \textbf{All Average} & 0.8922  & 0.8788  & 0.8976  & 0.8872  \\
    \midrule
    \midrule
    \multicolumn{1}{l|}{\multirow{6}[2]{*}{\textbf{Rewrite}}} & Baichuan2-13B & 0.7116  & 0.7698  & 0.5709  & 0.6473  & Baichuan2-13B & 0.8114  & 0.8268  & 0.7413  & 0.7741  \\
          & ChatGLM3-6B-32K & 0.7483  & 0.7845  & 0.6256  & 0.6869  & ChatGLM3-6B-32K & 0.8335  & 0.8289  & 0.7814  & 0.7964  \\
          & GPT-3.5-Turbo & 0.7515  & 0.7774  & 0.6359  & 0.6888  & GPT-3.5-Turbo & \cellcolor{lightpink}\textbf{0.8424} & 0.8272  & \cellcolor{lightpink}\textbf{0.8033} & \cellcolor{lightpink}\textbf{0.8069} \\
          & GPT-4-1106 & 0.7926  & 0.7871  & 0.7049  & 0.7303  & Llama3-8B & 0.8398  & 0.8289  & 0.7925  & 0.8022  \\
          & Qwen1.5-32B & \cellcolor{lightpink}\textbf{0.7954} & \cellcolor{lightpink}\textbf{0.7853} & \cellcolor{lightpink}\textbf{0.7131} & \cellcolor{lightpink}\textbf{0.7333} & Llama2-13B & 0.8409  & \cellcolor{lightpink}\textbf{0.8294} & 0.7956  & 0.8042  \\
          \rowcolor{Ocean}
          & \textbf{All Average} & 0.7599  & 0.7808  & 0.6501  & 0.6973  & \textbf{All Average} & 0.8336  & 0.8282  & 0.7828  & 0.7968  \\
    \midrule
    \midrule
    \multicolumn{1}{l|}{\multirow{6}[2]{*}{\textbf{Translate}}} & Baichuan2-13B & 0.5841  & 0.7553  & 0.4341  & 0.5085  & Baichuan2-13B & 0.6242  & 0.6086  & 0.4731  & 0.5141  \\
          & ChatGLM3-6B-32K & \cellcolor{lightpink}\textbf{0.6016} & \cellcolor{lightpink}\textbf{0.7961} & 0.4684  & \cellcolor{lightpink}\textbf{0.5338} & ChatGLM3-6B-32K & \cellcolor{lightpink}\textbf{0.6292} & 0.6122  & \cellcolor{lightpink}\textbf{0.4791} & \cellcolor{lightpink}\textbf{0.5189} \\
          & GPT-3.5-Turbo & 0.5417  & 0.6204  & 0.3735  & 0.4431  & GPT-3.5-Turbo & 0.6139  & 0.6065  & 0.4528  & 0.5036  \\
          & GPT-4-1106 & 0.5530  & 0.5984  & 0.3939  & 0.4545  & Llama3-8B & 0.6248  & 0.6096  & 0.4713  & 0.5138  \\
          & Qwen1.5-32B & 0.5451  & 0.4214  & \cellcolor{lightpink}\textbf{0.5905} & 0.4404  & Llama2-13B & 0.6291  & \cellcolor{lightpink}\textbf{0.6129} & 0.4775  & 0.5185  \\
          \rowcolor{Ocean}
          & \textbf{All Average} & 0.5651  & 0.6383  & 0.4521  & 0.4761  & \textbf{All Average} & 0.6242  & 0.6100  & 0.4708  & 0.5138  \\
    \bottomrule
    \end{tabular}}
  \label{tabA3}%
\end{table}%

\begin{table}[h]
  \centering
  \caption{Results of RoBERTa model trained on single LLM data and tested on each operation data.}
  \renewcommand{\arraystretch}{1.2}
  \resizebox{\textwidth}{!}{
    \begin{tabular}{rlllll|rlllll}
    \toprule
    \multicolumn{1}{l}{\multirow{2}[2]{*}{\textbf{Train Data}}} & \multirow{2}[2]{*}{\textbf{Test Data}} & \multicolumn{4}{c|}{\textbf{Chinese}} & \multicolumn{1}{l}{\multirow{2}[2]{*}{\textbf{Train Data}}} & \multirow{2}[2]{*}{\textbf{Test Data}} & \multicolumn{4}{c}{\textbf{English}} \\
          &       & Accuracy & Precision & Recall & F1-score &       &       & Accuracy & Precision & Recall & F1-score \\
    \midrule
    \midrule
    \multicolumn{1}{l}{\multirow{6}[2]{*}{\textbf{Baichuan2}}} & Create & 0.8821  & 0.8495  & 0.9162  & 0.8806  & \multicolumn{1}{l}{\multirow{6}[2]{*}{\textbf{Baichuan2}}} & Create & 0.9624  & 0.9683  & 0.9580  & 0.9623  \\
          & Update & 0.5368  & 0.5192  & 0.3429  & 0.4130  &       & Update & 0.9642  & 0.9484  & 0.9808  & 0.9642  \\
          & Delete & 0.7489  & 0.7460  & 0.6486  & 0.6881  &       & Delete & 0.9670  & \cellcolor{lightpink}\textbf{0.9713} & 0.9643  & 0.9669  \\
          & Rewrite & 0.5338  & 0.5176  & 0.3345  & 0.4063  &       & Rewrite & \cellcolor{lightpink}\textbf{0.9687} & 0.9481  & \cellcolor{lightpink}\textbf{0.9902} & \cellcolor{lightpink}\textbf{0.9687} \\
          & Translate & \cellcolor{lightpink}\textbf{0.9472} & \cellcolor{lightpink}\textbf{0.9439} & \cellcolor{lightpink}\textbf{0.9509} & \cellcolor{lightpink}\textbf{0.9472} &       & Translate & 0.9330  & 0.9213  & 0.9457  & 0.9329  \\
          \rowcolor{Ocean}
          & \textbf{All Average} & 0.7298  & 0.7152  & 0.6386  & 0.6670  &       & \textbf{All Average} & 0.9591  & 0.9515  & 0.9678  & 0.9590  \\
    \midrule
    \midrule
    \multicolumn{1}{l}{\multirow{6}[2]{*}{\textbf{ChatGLM3}}} & Create & 0.8536  & 0.8175  & \cellcolor{lightpink}\textbf{0.8852} & 0.8487  & \multicolumn{1}{l}{\multirow{6}[2]{*}{\textbf{ChatGLM3}}} & Create & \cellcolor{lightpink}\textbf{0.9389} & 0.9353  & 0.9444  & \cellcolor{lightpink}\textbf{0.9388} \\
          & Update & 0.5232  & 0.5119  & 0.3082  & 0.3847  &       & Update & 0.8494  & 0.7735  & 0.9348  & 0.8465  \\
          & Delete & 0.7410  & 0.7373  & 0.6332  & 0.6721  &       & Delete & 0.9190  & 0.8874  & \cellcolor{lightpink}\textbf{0.9549} & 0.9176  \\
          & Rewrite & 0.5237  & 0.5121  & 0.3086  & 0.3851  &       & Rewrite & 0.8550  & 0.7768  & 0.9432  & 0.8520  \\
          & Translate & \cellcolor{lightpink}\textbf{0.8930} & \cellcolor{lightpink}\textbf{0.9791} & 0.8228  & \cellcolor{lightpink}\textbf{0.8916} &       & Translate & 0.8447  & \cellcolor{lightpink}\textbf{0.9833} & 0.7410  & 0.8391  \\
          \rowcolor{Ocean}
          & \textbf{All Average} & 0.7069  & 0.7116  & 0.5916  & 0.6365  &       & \textbf{All Average} & 0.8814  & 0.8713  & 0.9036  & 0.8788  \\
    \midrule
    \midrule
    \multicolumn{1}{l}{\multirow{6}[2]{*}{\textbf{GPT-3.5}}} & Create & 0.8818  & 0.8527  & 0.9133  & 0.8807  & \multicolumn{1}{l}{\multirow{6}[2]{*}{\textbf{GPT-3.5}}} & Create & 0.9255  & 0.9849  & 0.8782  & 0.9240  \\
          & Update & 0.5419  & 0.5222  & 0.3640  & 0.4289  &       & Update & 0.9468  & 0.9951  & 0.9051  & 0.9463  \\
          & Delete & 0.7526  & 0.7465  & 0.6597  & 0.6957  &       & Delete & \cellcolor{lightpink}\textbf{0.9663} & 0.9937  & \cellcolor{lightpink}\textbf{0.9427} & \cellcolor{lightpink}\textbf{0.9660} \\
          & Rewrite & 0.5454  & 0.5239  & 0.3643  & 0.4297  &       & Rewrite & 0.9590  & \cellcolor{lightpink}\textbf{0.9953} & 0.9277  & 0.9588  \\
          & Translate & \cellcolor{lightpink}\textbf{0.9450} & \cellcolor{lightpink}\textbf{0.9571} & \cellcolor{lightpink}\textbf{0.9337} & \cellcolor{lightpink}\textbf{0.9450} &       & Translate & 0.9216  & 0.9273  & 0.9170  & 0.9215  \\
          \rowcolor{Ocean}
          & \textbf{All Average} & 0.7334  & 0.7205  & 0.6470  & 0.6760  &       & \textbf{All Average} & 0.9438  & 0.9793  & 0.9141  & 0.9433  \\
    \midrule
    \midrule
    \multicolumn{1}{l}{\multirow{6}[2]{*}{\textbf{GPT-4}}} & Create & 0.8966  & \cellcolor{lightpink}\textbf{0.9365} & 0.8633  & 0.8927  & \multicolumn{1}{l}{\multirow{6}[2]{*}{\textbf{Llama3}}} & Create & 0.9527  & 0.9692  & 0.9399  & 0.9524  \\
          & Update & 0.8203  & 0.7986  & 0.8466  & 0.8193  &       & Update & 0.9715  & 0.9848  & 0.9602  & 0.9714  \\
          & Delete & 0.9096  & 0.9017  & 0.9230  & 0.9090  &       & Delete & 0.9722  & \cellcolor{lightpink}\textbf{0.9893} & 0.9572  & 0.9720  \\
          & Rewrite & 0.8626  & 0.8149  & 0.9164  & 0.8618  &       & Rewrite & \cellcolor{lightpink}\textbf{0.9778} & 0.9813  & \cellcolor{lightpink}\textbf{0.9746} & \cellcolor{lightpink}\textbf{0.9778} \\
          & Translate & \cellcolor{lightpink}\textbf{0.9266} & 0.9233  & \cellcolor{lightpink}\textbf{0.9309} & \cellcolor{lightpink}\textbf{0.9266} &       & Translate & 0.9162  & 0.9309  & 0.9022  & 0.9161  \\
          \rowcolor{Ocean}
          & \textbf{All Average} & 0.8832  & 0.8750  & 0.8960  & 0.8819  &       & \textbf{All Average} & 0.9581  & 0.9711  & 0.9468  & 0.9579  \\
    \midrule
    \midrule
    \multicolumn{1}{l}{\multirow{6}[2]{*}{\textbf{Qwen1.5}}} & Create & 0.8597  & 0.9739  & 0.7745  & 0.8442  & \multicolumn{1}{l}{\multirow{6}[2]{*}{\textbf{Llama2}}} & Create & 0.9490  & 0.9899  & 0.9135  & 0.9486  \\
          & Update & 0.7896  & 0.8873  & 0.7085  & 0.7715  &       & Update & 0.9698  & 0.9918  & 0.9508  & 0.9696  \\
          & Delete & 0.8899  & 0.9471  & 0.8447  & 0.8868  &       & Delete & 0.9691  & 0.9938  & 0.9472  & 0.9690  \\
          & Rewrite & 0.8468  & 0.9135  & 0.7952  & 0.8403  &       & Rewrite & \cellcolor{lightpink}\textbf{0.9754} & \cellcolor{lightpink}\textbf{0.9944} & \cellcolor{lightpink}\textbf{0.9579} & \cellcolor{lightpink}\textbf{0.9753} \\
          & Translate & \cellcolor{lightpink}\textbf{0.9344} & \cellcolor{lightpink}\textbf{0.9485} & \cellcolor{lightpink}\textbf{0.9215} & \cellcolor{lightpink}\textbf{0.9343} &       & Translate & 0.9253  & 0.9469  & 0.9053  & 0.9252  \\
          \rowcolor{Ocean}
          & \textbf{All Average} & 0.8641  & 0.9341  & 0.8089  & 0.8554  &       & \textbf{All Average} & 0.9577  & 0.9834  & 0.9349  & 0.9575  \\
    \bottomrule
    \end{tabular}}
  \label{tabA4}%
\end{table}%

% Table generated by Excel2LaTeX from sheet 'Cross_LLM_opera_xlnet'
\begin{table}[hh]
  \centering
  \caption{Results of XLNet model trained on single LLM data and tested on each operation data.}
  \renewcommand{\arraystretch}{1.2}
  \resizebox{\textwidth}{!}{
    \begin{tabular}{llllll|llllll}
    \toprule
    \multirow{2}[2]{*}{\textbf{Train Data}} & \multirow{2}[2]{*}{\textbf{Test Data}} & \multicolumn{4}{c|}{\textbf{Chinese}} & \multirow{2}[2]{*}{\textbf{Train Data}} & \multirow{2}[2]{*}{\textbf{Test Data}} & \multicolumn{4}{c}{\textbf{English}} \\
          &       & Accuracy & Precision & Recall & F1-score &       &       & Accuracy & Precision & Recall & F1-score \\
    \midrule
    \midrule
    \multirow{6}[2]{*}{\textbf{Baichuan2}} & Create & 0.9037  & 0.8938  & 0.9145  & 0.9036  & \multirow{6}[2]{*}{\textbf{Baichuan2}} & Create & 0.9472  & 0.9416  & 0.9547  & 0.9471  \\
          & Update & 0.5490  & 0.5263  & 0.3856  & 0.4450  &       & Update & 0.9629  & 0.9610  & 0.9655  & 0.9629  \\
          & Delete & 0.7381  & 0.7412  & 0.6422  & 0.6828  &       & Delete & 0.9598  & \cellcolor{lightpink}\textbf{0.9722} & 0.9500  & 0.9596  \\
          & Rewrite & 0.5528  & 0.5281  & 0.3860  & 0.4460  &       & Rewrite & \cellcolor{lightpink}\textbf{0.9679} & 0.9616  & \cellcolor{lightpink}\textbf{0.9747} & \cellcolor{lightpink}\textbf{0.9679} \\
          & Translate & \cellcolor{lightpink}\textbf{0.9618} & \cellcolor{lightpink}\textbf{0.9643} & \cellcolor{lightpink}\textbf{0.9596} & \cellcolor{lightpink}\textbf{0.9617} &       & Translate & 0.8995  & 0.8730  & 0.9277  & 0.8994  \\
          \rowcolor{Ocean}
          & \textbf{All Average} & 0.7411  & 0.7307  & 0.6576  & 0.6878  &       & \textbf{All Average} & 0.9475  & 0.9419  & 0.9545  & 0.9474  \\
    \midrule
    \midrule
    \multirow{6}[2]{*}{\textbf{ChatGLM3}} & Create & \cellcolor{lightpink}\textbf{0.8958} & 0.8810  & \cellcolor{lightpink}\textbf{0.9116} & \cellcolor{lightpink}\textbf{0.8956} & \multirow{6}[2]{*}{\textbf{ChatGLM3}} & Create & \cellcolor{lightpink}\textbf{0.9443} & 0.9475  & 0.9426  & \cellcolor{lightpink}\textbf{0.9442} \\
          & Update & 0.5691  & 0.5378  & 0.4342  & 0.4804  &       & Update & 0.8774  & 0.8159  & 0.9468  & 0.8763  \\
          & Delete & 0.7524  & 0.7593  & 0.6684  & 0.7074  &       & Delete & 0.9307  & 0.9040  & 0.9623  & 0.9300  \\
          & Rewrite & 0.5738  & 0.5401  & 0.4347  & 0.4817  &       & Rewrite & 0.8867  & 0.8184  & \cellcolor{lightpink}\textbf{0.9644} & 0.8854  \\
          & Translate & 0.8310  & \cellcolor{lightpink}\textbf{0.9874} & 0.7162  & 0.8240  &       & Translate & 0.8475  & \cellcolor{lightpink}\textbf{0.9572} & 0.7621  & 0.8434  \\
          \rowcolor{Ocean}
          & \textbf{All Average} & 0.7244  & 0.7411  & 0.6330  & 0.6778  &       & \textbf{All Average} & 0.8973  & 0.8886  & 0.9156  & 0.8959  \\
    \midrule
    \midrule
    \multirow{6}[2]{*}{\textbf{GPT-3.5}} & Create & 0.8874  & 0.8616  & \cellcolor{lightpink}\textbf{0.9157} & 0.8864  & \multirow{6}[2]{*}{\textbf{GPT-3.5}} & Create & 0.9439  & 0.9729  & 0.9210  & 0.9433  \\
          & Update & 0.6445  & 0.5896  & 0.6296  & 0.6088  &       & Update & 0.9520  & 0.9874  & 0.9208  & 0.9517  \\
          & Delete & 0.7953  & 0.7814  & 0.7772  & 0.7754  &       & Delete & 0.9668  & 0.9814  & \cellcolor{lightpink}\textbf{0.9546} & 0.9667  \\
          & Rewrite & 0.6620  & 0.5990  & 0.6511  & 0.6240  &       & Rewrite & \cellcolor{lightpink}\textbf{0.9706} & \cellcolor{lightpink}\textbf{0.9899} & 0.9529  & \cellcolor{lightpink}\textbf{0.9705} \\
          & Translate & \cellcolor{lightpink}\textbf{0.9259} & \cellcolor{lightpink}\textbf{0.9910} & 0.8726  & \cellcolor{lightpink}\textbf{0.9243} &       & Translate & 0.8658  & 0.8136  & 0.9231  & 0.8645  \\
          \rowcolor{Ocean}
          & \textbf{All Average} & 0.7830  & 0.7645  & 0.7692  & 0.7638  &       & \textbf{All Average} & 0.9398  & 0.9490  & 0.9345  & 0.9393  \\
    \midrule
    \midrule
    \multirow{6}[2]{*}{\textbf{GPT-4}} & Create & 0.9030  & 0.9448  & 0.8688  & 0.8965  & \multirow{6}[2]{*}{\textbf{Llama3}} & Create & 0.9421  & 0.9326  & 0.9536  & 0.9419  \\
          & Update & 0.8308  & 0.8990  & 0.7776  & 0.8214  &       & Update & 0.9668  & \cellcolor{lightpink}\textbf{0.9696} & 0.9650  & 0.9668  \\
          & Delete & 0.9225  & 0.9345  & \cellcolor{lightpink}\textbf{0.9148} & 0.9219  &       & Delete & 0.9602  & 0.9694  & 0.9532  & 0.9601  \\
          & Rewrite & 0.9002  & 0.9218  & 0.8856  & 0.8990  &       & Rewrite & \cellcolor{lightpink}\textbf{0.9718} & 0.9663  & \cellcolor{lightpink}\textbf{0.9775} & \cellcolor{lightpink}\textbf{0.9718} \\
          & Translate & \cellcolor{lightpink}\textbf{0.9266} & \cellcolor{lightpink}\textbf{0.9901} & 0.8750  & \cellcolor{lightpink}\textbf{0.9248} &       & Translate & 0.8431  & 0.7831  & 0.9092  & 0.8412  \\
          \rowcolor{Ocean}
          & \textbf{All Average} & 0.8966  & 0.9380  & 0.8644  & 0.8927  &       & \textbf{All Average} & 0.9368  & 0.9242  & 0.9517  & 0.9364  \\
    \midrule
    \midrule
    \multirow{6}[2]{*}{\textbf{Qwen1.5}} & Create & 0.8026  & 0.9363  & 0.6833  & 0.7647  & \multirow{6}[2]{*}{\textbf{Llama2}} & Create & 0.9435  & 0.9612  & 0.9285  & 0.9433  \\
          & Update & 0.7078  & 0.9766  & 0.5284  & 0.6414  &       & Update & 0.9627  & 0.9744  & 0.9527  & 0.9626  \\
          & Delete & 0.8284  & 0.9197  & 0.7627  & 0.8193  &       & Delete & 0.9589  & \cellcolor{lightpink}\textbf{0.9835} & 0.9386  & 0.9585  \\
          & Rewrite & 0.7611  & \cellcolor{lightpink}\textbf{0.9881} & 0.6107  & 0.7217  &       & Rewrite & \cellcolor{lightpink}\textbf{0.9699} & 0.9738  & \cellcolor{lightpink}\textbf{0.9668} & \cellcolor{lightpink}\textbf{0.9699} \\
          & Translate & \cellcolor{lightpink}\textbf{0.9237} & 0.9622  & \cellcolor{lightpink}\textbf{0.8931} & \cellcolor{lightpink}\textbf{0.9223} &       & Translate & 0.8890  & 0.8622  & 0.9181  & 0.8888  \\
          \rowcolor{Ocean}
          & \textbf{All Average} & 0.8047  & 0.9566  & 0.6956  & 0.7739  &       & \textbf{All Average} & 0.9448  & 0.9510  & 0.9409  & 0.9446  \\
    \bottomrule
    \end{tabular}}
  \label{tabA5}
\end{table}%

\end{document}